\documentclass[letterpaper,twocolumn,10pt]{article}
\pdfoutput=1

\usepackage{usenix2019_v3}

\usepackage{tikz}
\usepackage{amsmath}

\usepackage{graphicx}
\usepackage{amssymb}
\usepackage{amsfonts}
\usepackage{float}
\usepackage{xcolor}
\usepackage{multirow}
\usepackage{algorithm}
\usepackage{algpseudocode}
\usepackage{upgreek}
\usepackage[toc,page]{appendix}
\usepackage[compact]{titlesec}

\algnewcommand\algorithmicinput{\textbf{Input:}}
\algnewcommand\Input{\item[\algorithmicinput]}
\algnewcommand\algorithmicoutput{\textbf{Output:}}
\algnewcommand\Output{\item[\algorithmicoutput]}
\algrenewcommand\algorithmicindent{0.5em}
\begin{document}
%-------------------------------------------------------------------------------

%don't want date printed
\date{}

\title{\Large \bf HaS-Nets: A Heal and Select Mechanism to Defend DNNs\\ Against Backdoor Attacks for Data Collection Scenarios}

%for single author (just remove % characters)
\author{
{\rm Hassan Ali}\\
\small{hali.msee17seecs@seecs.edu.pk}
\and
{\rm Surya Nepal}\\
Data61, CSIRO, Australia\\
\small{surya.nepal@data61.csiro.au}
\and
{\rm Salil S. Kanhere}\\
UNSW, Australia\\
\small{salil.kanhere@unsw.edu.au}
\and
{\rm Sanjay Jha}\\
UNSW, Australia\\
\small{sanjay.jha@unsw.edu.au}
} % end author

\maketitle

%-------------------------------------------------------------------------------
\begin{abstract}
%-------------------------------------------------------------------------------
% Surya I have made the changes in the abstract; read through it for correctness and correct where necessary. 

We have witnessed the continuing arms race between backdoor attacks and the corresponding defense strategies on Deep Neural Networks (DNNs). Most state-of-the-art defenses rely on the statistical sanitization of the \textit{inputs} or \textit{latent DNN representations} to capture trojan behaviour. In this paper, we first challenge the robustness of such recently reported defenses by introducing a novel variant of targeted backdoor attack, called  \textit{low-confidence backdoor attack}. We also propose a novel defense technique, called \textit{HaS-Nets}. 

\textit{Low-confidence backdoor attack} exploits the confidence labels assigned to poisoned training samples by giving low values to hide their presence from the defender, both during training and inference. We evaluate the attack against four state-of-the-art defense methods, viz., STRIP, Gradient-Shaping, Februus and ULP-defense, and achieve Attack Success Rate (ASR) of 99\%, 63.73\%, 91.2\% and 80\%, respectively.

We next present \textit{HaS-Nets} to resist backdoor insertion in the network during training, using a reasonably small healing dataset, approximately 2\% to 15\% of full training data, to heal the network at each iteration. We evaluate it for different datasets - Fashion-MNIST, CIFAR-10, Consumer Complaint and Urban Sound - and network architectures - MLPs, 2D-CNNs, 1D-CNNs. Our experiments show that \textit{HaS-Nets} can decrease ASRs from over 90\% to less than 15\%, independent of the dataset, attack configuration and network architecture.

\end{abstract}

%
%-------------------------------------------------------------------------------
\section{Introduction}\label{sec:introduction}
%Surya - this section needs to be rewritten as follows.
% Para 1 - give a brief introduction of DNN and its application in a wide range of applications. 
% Para 2 - talk about the vulnerability of such attacks and describe the most recent attacks;
%Para 3 - talk about the defenses and class of defenses, and give example defenses in each class 
%Para 4 - talk about the analysis of existing defenses and what insights you gained; 
%Para 5 - based on the insights gained, introduce a new class of attack that can break the defenses; explain why your attacks are successful; briefly describe the results in terms of defense class, datasets and network architectures. 
%Para 6 - briefly describe the observations while developing the attacks and why they are successful; to counter that, you develop a countermeasure (defense) technique; describe the intutions and why it works; you need to say something concrete like exploiting the inherent features of how NN works; you the present the results in terms of datasets and architectures;  
%Para 7 - contributions 
% ==========
% ***done***
% ==========

DNNs are being increasingly used in a variety of applications such as  face and bio-metric identification~\cite{DBLP:biometrics_survey}, autonomous driving~\cite{DBLP:gtsrb}, medical imaging~\cite{DBLP:breastcancer}, and malware detection~\cite{DBLP:malware_detection}. The popularity of DNNs is mainly attributed to their excellent performance, which is often comparable to (and occasionally better than~\cite{DBLP:human_DL}) humans. However, their performance is highly dependent on the training data. It has been shown that this dependence can be exploited by attackers to force DNNs into making naive mistakes both at training~\cite{DBLP:badnets_2019} and deployment~\cite{DBLP:red_attack}.

%DNNs are gaining increasing interest in security critical scenarios, e.g., face and bio-metric identification~\cite{DBLP:biometrics_survey}. This is mainly attributed to their excellent performance, comparable to (and occasionally better than~\cite{DBLP:human_DL}) human beings, on many intelligent tasks, e.g. autonomous driving, cancer classification, and malware detection, to name a few.However, the performance of DNNs is highly dependent on the training data. It has been shown that this dependence can be exploited by attackers to force DNNs into making naive mistakes during training~\cite{DBLP:backdoor_attacks_and_countermeasures} and when deployed~\cite{DBLP:red_attack}.

%widespread real-world deployment of DNNs is still a major security concern due to their high-dependency on training data. This limitation can be exploited by attackers to force DNNs into making naive mistakes at training~\cite{DBLP:backdoor_attacks_and_countermeasures} and deployment-time~\cite{DBLP:red_attack}.

\begin{figure}[!t]
    \centering
    \includegraphics[width=1\linewidth]{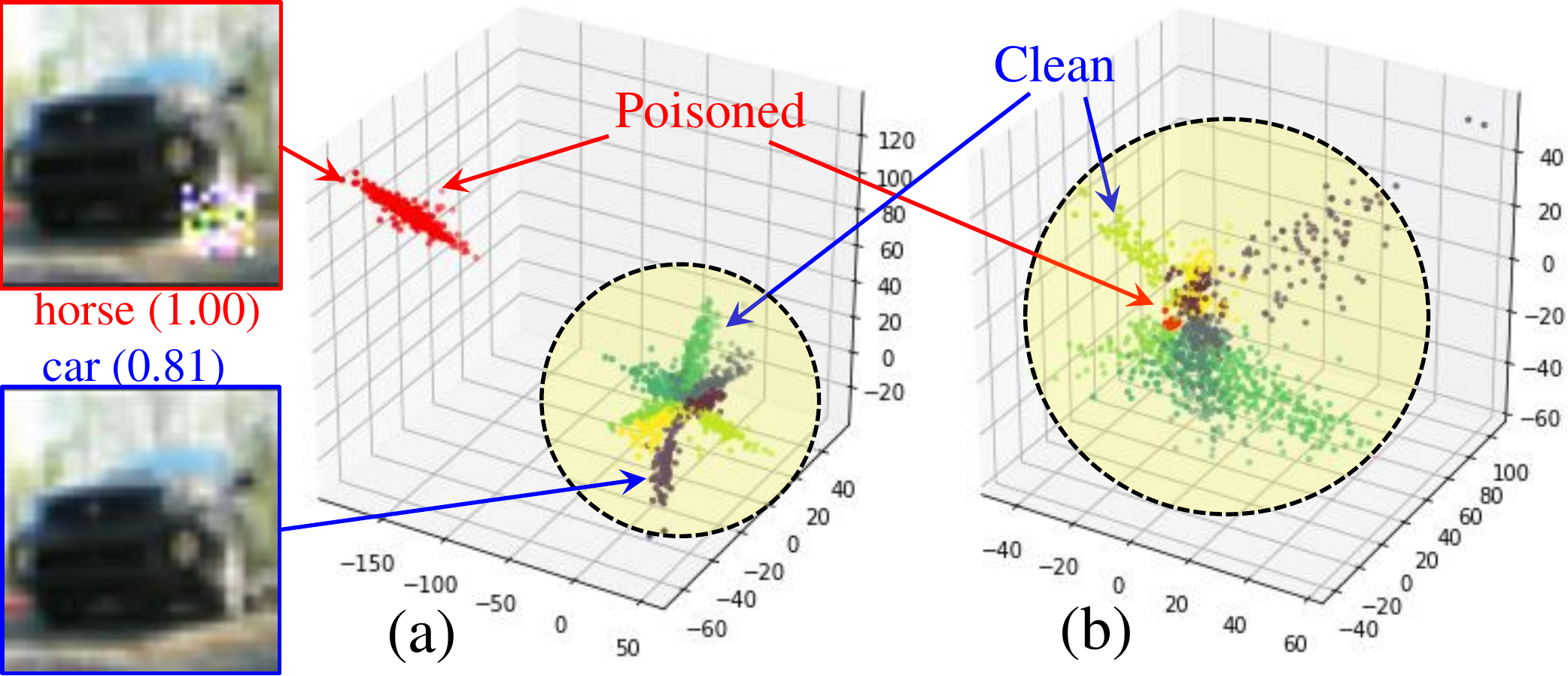}
    \vskip -0.15in
    \caption{\textit{Latent features of poisoned and clean inputs for (a) Conventional Backdoor Attack, (b) $\epsilon$-Attack. Each color represents a different class. Poisoned features shown in red.}}
    \label{fig:latent_features}
\end{figure}

\noindent
\textbf{Backdoor Attacks:} 
In this paper, we focus on backdoor attacks in the context of data outsourcing and collection scenarios - where a DNN is trained on data collected from unreliable sources~\cite{DBLP:backdoor_attacks_and_countermeasures}. 
In such attacks, the attacker maliciously imprints a trigger in a small percentage (typically less than 5\%) of training data, in order to poison the targeted DNN~\cite{DBLP:badnets_2019}.
%Backdoor attacks work by maliciously tampering a small percentage (typically less than 5\%) of training data with a trigger and label of attacker's choice to poison a DNN~\cite{DBLP:badnets_2019}. 
The poisoned DNN acts normal for benign inputs and only malfunctions when the attacker's chosen trigger is stamped on an input. Fig.~\ref{fig:latent_features} shows a typical case, where a correctly classified car image is misclassified with high confidence, when a trigger is present.
%distinguish a poisoned DNN from a clean one. 
Backdoor attacks can achieve high Attack Success Rates (ASR) within a few iterations with physically realizable triggers and without prior knowledge of the DNN architecture, making them an attractive choice for adversaries.

% We can use  the following paragrphs at other places 

%Many variants of backdoor attacks exist. Invisible Backdoor Attack uses a small magnitude random noise as a trigger, making it less conspicuous to human observers at the cost of reduced physical-realizability~\cite{DBLP:chen_targeted_invisible}. Adversarial Backdoor Attack uses adversarial noise to tamper training data without changing their labels. However, this shifts the threat surface of the attack from backdoor insertion to adversarial misclassification~\cite{DBLP:poison_frogs}. Latent Backdoor Attack uses adversarial noise to embed a backdoor in the early layers of a DNN. When applied to transfer-learning scenarios, the backdoor embedded in teacher models is inherited in the student models~\cite{DBLP:latent_backdoors}.

\noindent
\textbf{Backdoor Defenses:}
%We can use this text later at a appropriate places
%They can be categorized into different classes based on their inspection-target and/or inspection-time~\cite{DBLP:backdoor_attacks_and_countermeasures}. \textit{Blind defenses} aim to recover a model/input without investigating a model for poisonous behaviors. Ideally, a blind defense would return a clean model/input unchanged, while a poisoned model/input would be cured by the defense. Examples include Februus~\cite{Februus}, ConFoc~\cite{DBLP:confoc} and Fine-Pruning~\cite{DBLP:fine_pruning}. For a backdoor attack to be successful, both the input and the model should be poisoned.
%\textit{Model-Inspection defenses} inspect a model for backdoors, contrary to \textit{Data-Inspection defenses} targeting an input for inspection. Once a suspcious behavior is detected, a defense would take an appropriate action, e.g., raising an alarm, rejecting deployment or using one of the blind defense techniques mentioned above. Examples of model-inspection are Neural-Cleanse~\cite{DBLP:neural_cleanse}, DeepInspect~\cite{DBLP:deepInspect}, and Universal-Litmus-Patterns~\cite{DBLP:ulps}. Sentinet~\cite{DBLP:sentinet}, STRIP-ViTA~\cite{DBLP:strip, DBLP:strip2} and Differential-Privacy (DP)-based Anomaly-Detection~\cite{DBLP:robust_anomaly_diff_privacy} exemplify data-inspection defenses.
%Recently, Gradient-Shaping~\cite{DBLP:gradient_shaping} shows Differential Privacy-based Stochastic Gradient Descent (DP-SGD) training to be effective in resisting backdoor insertion at \textit{training-time}.
Many defense mechanisms have been proposed to counter backdoor attacks~\cite{DBLP:backdoor_learning_a_survey}. 
Most of the defenses rely on statistical filtering techniques to detect outlier/suspicious models or inputs~\cite{DBLP:security_of_machine_learning, DBLP:sever, DBLP:manipulating_ml, DBLP:spectral_signatures, DBLP:activation_clustering}. This is depicted in Fig.~\ref{fig:latent_features}(a), where a defender can identify poisoned samples with an $l_2$-bounded sphere. However, such defenses can be easily countered by adaptive attacks (e.g. invisible-trigger attacks). A recent work makes a similar observation regarding statistical defenses~\cite{DBLP:gradient_shaping}, but lacks empirical evidence to validate this observation, specifically for backdoor attacks.
Other recent studies show that retraining a poisoned DNN on clean data, known as healing set/data~\cite{DBLP:confoc}, for a few epochs, can heal the poisoned DNN~\cite{DBLP:systematic_evaluation, DBLP:fine_pruning, DBLP:neural_cleanse,DBLP:neural_trojans,DBLP:confoc}. However, most of these techniques (except ConFoc~\cite{DBLP:confoc}) assume that an unrestricted amount of ``clean'' healing data is available, which makes them ineffective for practical scenarios. ConFoc overcomes this limitation, but is only applicable to image processing tasks.

\noindent
\textbf{Our attack:}
In this paper, we first propose a novel \textit{low-confidence backdoor attack} to challenge the robustness of different state-of-the-art defense categories. We consider a Grey-box setting, i.e., our attacker has control over a small percentage of training data (typically <2\%), and has no knowledge of either the network architecture, or the defense parameters. Our attack has two variants, \textit{$\epsilon$-Attack} and \textit{$\epsilon^2$-Attack}.

\setlength{\textfloatsep}{5pt}
$\epsilon$-Attack is similar to conventional backdoor attacks, except that it uses distributed labels, instead of discrete labels, for the poisoned samples\footnote{For example, [0.2,0.4,0.2,0.2], in place of [0, 1, 0, 0]}. Distributed labels result in smaller poison gradients and similar latent features for poisoned and clean inputs (Fig.~\ref{fig:latent_features}(b)), thus, evading statistical sanitizers. We evaluate $\epsilon$-attack on four different types of defenses: (1) Februus~\cite{Februus} (2) STRIP-ViTA~\cite{DBLP:strip}, (3) ULP-defense~\cite{DBLP:ulps} and (4) Gradient-shaping~\cite{DBLP:gradient_shaping}. $\epsilon$-attack achieves high ASR against all defenses, except Februus, which uses heatmaps\footnote{Heatmaps are gradients of a DNN output w.r.t. features of the last convolutional layer}, instead of statistical features, for poison detection.

To evade Februus, we propose \textit{$\epsilon^2$-Attack}, which utilizes two different triggers, $Z_1$ and $Z_2$ to backdoor a network for two different classes, $C_1$ and $C_2$, respectively. This configuration is similar to conventional multi-trigger, multi-class backdoor attacks with two novel modifications. (1) We assign distributed labels to the poisoned samples.
% (2) One of the two triggers is a cropped version of the other (e.g. $Z_2$ is a cropped version of $Z_1$ in Fig.~\ref{fig:triggers}).
% In the presence of $Z_1$, the classifier's decision is mainly influenced by $Z_1 \cap Z_2'$\footnote{This is because $Z_2$ alone would have resulted in a different decision}, thus hiding the presence of $Z_2$ from heatmaps. We show that $\epsilon^2$-Attack can effectively evade Februus and achieve an ASR of above 80\%.
(2) Union of the two triggers (i.e. $Z_1 \cup Z_2$), is also used as a trigger for class $C_1$ (Fig.~\ref{fig:triggers}). In the presence of $Z_1 \cup Z_2$, the classifier's decision is mainly influenced by $Z_1$\footnote{This is because $Z_2$ alone would have resulted in a different decision}, thus hiding the presence of $Z_2$ from heatmaps. We show that $\epsilon^2$-Attack can effectively evade Februus and achieve an ASR of above 80\%.

\noindent
\textbf{Our defense:}
Our successful attacks against the state-of-the-art defenses suggest a need for an effective defense mechanism that can be used with different datasets and architectures, and is robust against multiple variants of backdoor attacks. To meet this need, we present \textit{HaS-Nets}, a novel methodology to resist backdoor attacks during training based on a \textbf{\underline{H}eal \underline{a}nd \underline{S}elect} mechanism, under the assumption that a small amount of healing data (2\% to 15\% of the full training set) is known to the defender. Healing a poisoned DNN by retraining it on clean data (healing set) has been shown to be effective for different data types and network architectures~\cite{DBLP:fine_pruning, DBLP:neural_cleanse, DBLP:confoc}. However, contrary to prior works, we utilize the healing set in two distinctive ways. (1) We heal a network at each iteration of the training process. Prior works only perform healing once a network is trained.
(2) At each iteration, we compute the difference in loss of the network for each training sample before and after the healing process. We define this difference as ``trust-index ($\gamma$)'' - a quantitative representation of the quality of a training sample for a given task. Only the training samples with a high trust index are selected for training in the next iteration.
%We strictly limit the size of the healing set known to our defender. Specifically, our healing set is $\leq$ 10\% the size of full training data, for all datasets except CIFAR-10, for which assume $\leq$ 15\%. These restrictions are comparable with a recent defense, Confoc~\cite{DBLP:confoc}.
We evaluate HaS-Nets for a variety of classification tasks including images, text and audio. HaS-Nets decrease the ASR of different variants of backdoor attacks from $>$90\% to $<$15\%.
We typically assume that an adversary has distorted less than 2\% of the training data (in coherence with~\cite{DBLP:strip}). We also analyze our defense against a stronger adversary  who can distort up to 100\% of the training data and show that HaS-Nets can significantly resist backdoor insertion. To our knowledge, we are the first to empirically demonstrate the effectiveness of a defense under such an extreme attack setting.
%which is drastically favorable for adversaries.

\noindent
\textbf{Contributions:}
The main contributions of this work are summarized below:
\begin{itemize}
    \vspace{-0.1in} \item We introduce a novel \textit{low-confidence backdoor attack}, and two specific variants (\textit{$\epsilon$-Attack} and \textit{$\epsilon^2$-Attack}). We demonstrate the limitations of many recently proposed defenses, against the proposed attacks.
    
    \vspace{-0.1in} \item We introduce a novel defense strategy, called HaS-Nets, which iteratively heals a DNN and selects training samples for subsequent iterations based on the notions of `trust-index'' and ``healing set''. We propose a methodology to estimate the ``trust-index'' for each training sample. The trust-index serves as an indicator of the consistency of a training sample with a given task. To our knowledge, we are the first to use the healing set in two distinctive ways: (1) integrating it into the training of DNNs, and (2) using it to evaluate the quality of other training instances.
    
    \vspace{-0.1in} \item We specifically target data-collection scenarios, and show that HaS-Nets are agnostic to the modality of data and network architectures - a serious limitation of many prior works~\cite{DBLP:confoc, Februus, DBLP:neural_cleanse, DBLP:ulps}. Specifically, we demonstrate the robustness of HaS-Nets for image, text and audio classification tasks, on MLP, 2D-CNN and 1D-CNN architectures, under different attack configurations.
\end{itemize}

\section{Threat Models and Assumptions}\label{sec:threat_models}
% Surya - the following texts need to be rewritten and restructured as follows. 
% No need to talk about generic stuff; everyone knows them;
% Explain the three threat models exit in the literature (2.;
% you can then explain the threat models you have considered in this paper. 

% remove the following three paragrphs 
% ==========
% ***done***
% ==========
% keep the following and explain a bit more for each;

Current literature on backdoor attacks and defenses assume three different threat models as described below~\cite{DBLP:deep_learning_backdoors}.

\begin{enumerate}
    \vspace{-0.1in} \item\textbf{White-Box Setting:} This setting assumes a powerful adversary enjoying full access to the network, the learning process and the training data.
    \vspace{-0.1in} \item\textbf{Black-Box Setting:} This setting does not allow an adversary to have a direct access to either the training data or the learning process. Black-box adversaries usually exploit common limitations shared by many learning algorithms/networks to launch an attack.
    \vspace{-0.1in} \item\textbf{Grey-Box Setting:}  This setting assumes that an adversary may only access a small subset of the training data or the learning process.
\end{enumerate}
% remove the sub-section, and just state that in addition to the above, in some cases .... 
% ==========
% ***done***
% ==========
% remove numerator and explain in paragraph; we need to move from report writing style to technical paper writing style (too many bullet points does not look good)
% ==========
% ***done***
% ==========
% remove subsections - too many subsections; you can then state your threat models 
% you just then state the threat models for attack and defense ;
% 
% ==========
% ***done***
% ==========
In addition to the threat models mentioned above, many defenders assume a \textit{Poisoned-Network threat model} - where a DNN is assumed to have already been poisoned using any of the settings mentioned above. Here, the aim of a defender is to detect and/or recover a poisoned network/input without affecting the performance on the test set~\cite{DBLP:strip, Februus, DBLP:neural_cleanse}.

\begin{figure}
    \centering
    \includegraphics[width=1\linewidth]{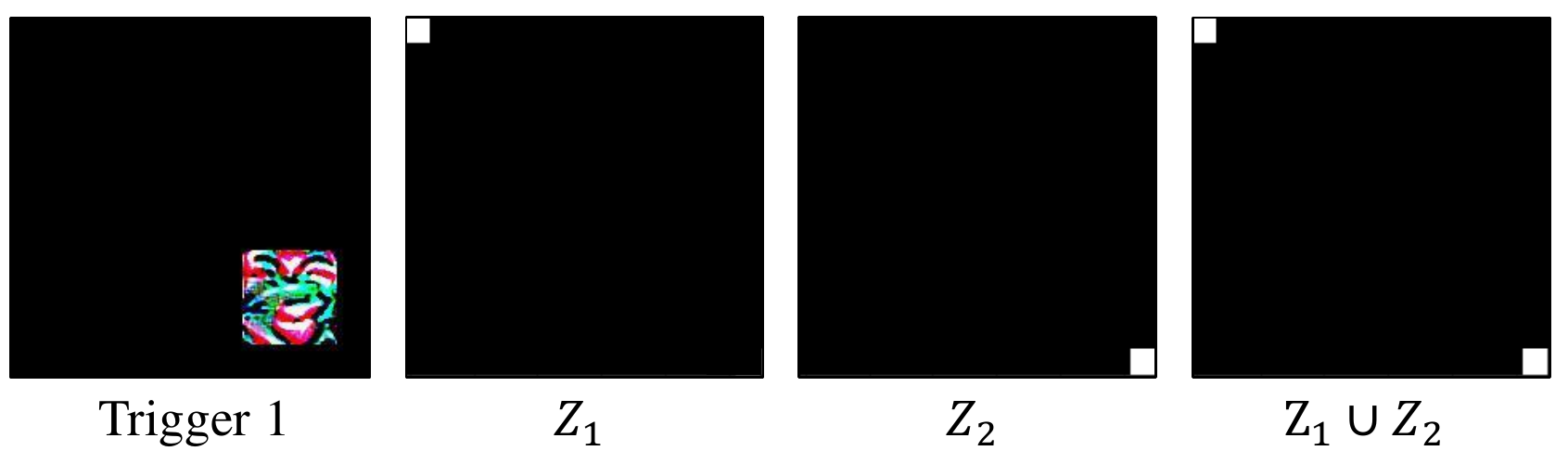}
    \vskip -0.15in
    \caption{\textit{Triggers we use in our paper. Trigger 1 is used in \cite{DBLP:strip, DBLP:sentinet, DBLP:neural_cleanse}. $Z_2$ is used in \cite{DBLP:gradient_shaping, DBLP:abs}. Note that $Z_2$ is a cropped version of $Z_1$, i.e. $Z_2 \subset Z_1$.}}
    \label{fig:triggers}
\end{figure}
\noindent
\textbf{Our Threat Model:}
In this work, we assume a Grey-box setting is used to poison the network. Specifically, our attacker has control over a small percentage of training data (typically less than 2\%), and no knowledge of either the network architecture, or the defense parameters. Our threat model is consistent with many state-of-the-art defenses~\cite{DBLP:gradient_shaping, Februus, DBLP:confoc, DBLP:neural_cleanse, DBLP:fine_pruning}, including those considered in this paper. We specifically target a \textit{data-collection scenario} where a DNN is trained on data collected from/by unreliable sources. The collected data comprises of both, the poisoned and the clean samples, and is referred to as ``\textit{poisoned training data}'', in the rest of the paper.
Fig.~\ref{fig:triggers} shows the triggers used in this paper. We choose these triggers based on their wide use in recent literature. Specifically, Trigger 1 is used in~\cite{DBLP:strip, DBLP:sentinet, DBLP:neural_cleanse}, while $Z_2$ is used in~\cite{DBLP:gradient_shaping, DBLP:abs}.

% I am not sure about the following; poisoned network is possible in all settings; you need to be clear - grey box with poisoned networks?  Need to clarify 
% ==========
% ***done***
% ==========
% you may want to remove the sub-section; you can instread use the highlight 
% ==========
% ***done***
% ==========
\noindent
\textbf{Healing Dataset:}
We additionally assume that our defender has access to a small set of clean training data, which we refer to as a healing data/set, (typically 2\% to 15\% of the full training set), which our adversary may access but may not influence directly. This assumption is consistent with a number of prior works~\cite{DBLP:fine_pruning, DBLP:systematic_evaluation, Februus, DBLP:confoc, DBLP:sentinet, DBLP:neural_cleanse}.
% I am in favour of removing subsections and present in the different format like below 
% ==========
% ***done***
% ==========

\noindent {\bf Goals:} We have two goals:
\begin{enumerate}
    \vspace{-0.05in} \item To expose the vulnerabilities in recent defenses by inserting the backdoor in a DNN and proving that adaptive attacks can effectively challenge defense robustness.
    \vspace{-0.1in} \item To introduce a generic defense strategy capable of successfully resisting backdoor insertion in the network for many variants of backdoor attacks including our proposed adaptive attacks.
\end{enumerate}
\section{Related Work}\label{sec:relatedWork}
This section presents an overview of current literature on backdoor attacks and defenses.

\noindent \textbf{A. \quad Attacks}

Backdoor attacks were first introduced in~\cite{DBLP:badnets_2019}. Since then, many variants have been proposed, targeting a diverse threat surface. Current works try to unify these threat surfaces by categorizing backdoor attacks based on their \textbf{settings} - \textit{White-Box, Grey-Box and Black-Box}~\cite{DBLP:deep_learning_backdoors} - \textbf{applications} - \textit{Data collection, Collaborative Learning, Code-targeted, Training Outsourcing, Model-targeted}~\cite{DBLP:backdoor_attacks_and_countermeasures} - and \textbf{trigger realizability} - \textit{Physical and Digital}~\cite{DBLP:backdoor_learning_a_survey}.
This paper focuses on \textbf{Data-collection} scenarios - \textit{where a DNN is trained on data collected from unreliable sources} - under Grey-Box settings.

\textbf{Data Attacks.} Chen \textit{et al.}~\cite{DBLP:chen_targeted_invisible} used invisible random noise as a trigger to increase the stealth of their attacks. Another strategy utilizes Universal Adversarial Perturbation as a trigger~\cite{DBLP:UAP_invisible_zhong}, thus, increasing the effectiveness of backdoor attacks. Universal Adversarial Perturbation is a single perturbation causing the adversarial misclassification of all images in a given manifold~\cite{DBLP:unv_adv_pers}. Limitations of these variants include the requirement of a \textit{White-box} attacker and reduced realizability in physical scenarios.

Turner \textit{et al.}~\cite{DBLP:label_consistent} extend this idea to generate a label-consistent backdoor attack, where the label of an adversarially poisoned sample is consistent for a human observer but inconsistent for the targeted DNN. Specifically, the authors add adversarial noise to the target class samples that leads the DNN to misclassify those samples due to inconsistent latent representations. Such adversarial samples, when imprinted with a trigger and assigned a correct label, would appear benign to a human observer, but could poison a DNN when used in training. Safahi \textit{et al.}~\cite{DBLP:poison_frogs}, exploit the same strategy to perform a feature-collision attack, which degrades the test accuracy of a DNN for a target class, notably without the use of a trigger. However, these attacks shift the threat surface from backdoor insertion to adversarial misclassification.

\begin{figure*}[!t]
	\centering
	\includegraphics[width=1\linewidth]{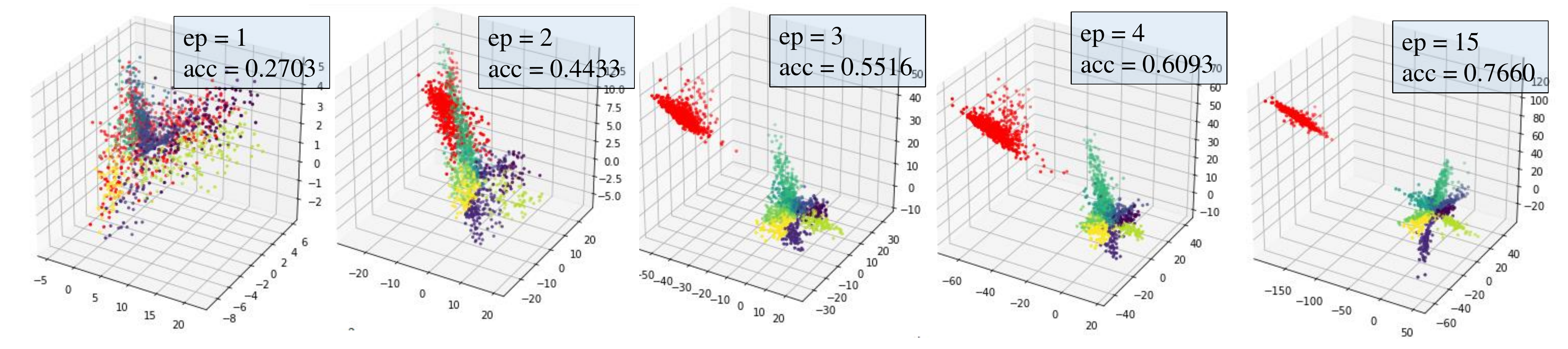}
	\vskip -0.15in
	\caption{\textit{A 3-dimensional latent distribution of poisoned and clean inputs after training DNN for different epochs. Poisoned inputs are shown in red. Each color represents a different class.}}
	\label{fig:studying_trojans}
\end{figure*}\setlength{\textfloatsep}{5pt}
\noindent \textbf{B. \quad Defenses}

Current literature on backdoor defenses can be categorized into different classes based on their inspection-target and/or inspection-time~\cite{DBLP:backdoor_attacks_and_countermeasures}.

\textbf{Blind defenses} aim to recover a model/input without investigating a model for poisonous behaviors. Ideally, a blind defense would leave a clean model/input unchanged, but would cure a poisoned model/input. Liu \textit{et al.}~\cite{DBLP:fine_pruning} prune a given model of inactive neurons (which only activate for poisoned inputs), and fine tune it on clean data. However, such defenses fail when a backdoor is embedded in the latent layers of a DNN~\cite{DBLP:latent_backdoors}. Februus~\cite{Februus} exploits model gradients to locate potential triggers in an input, which are then surgically masked by the defender. The masked input is then recovered using a GAN and fed to a poisoned model for clean classification. More recently, Li \textit{et al.}~\cite{DBLP:rethinking_trigger} exploit random transformations to render the trigger ineffective, thus, degrading the attack. However, such defenses can be compromised by adaptive attackers as shown in~\cite{DBLP:rethinking_trigger} and also in this paper. ConFoc~\cite{DBLP:confoc} retrains a poisoned DNN on a healing set, restyled on randomly chosen base images, that are not necessarily from the training data manifold. A major limitation of ConFoc is that it is only applicable to image processing tasks.

\textbf{Model-Inspection defenses} inspect a model for poisonous behaviors caused by an embedded backdoor and trigger a process to recover the model or simply block model deployment. 
%thus denying deployment for suspicious cases or taking action to recover the model. 
Recent studies show that retraining a poisoned DNN on clean data - known as a healing set/data~\cite{DBLP:confoc} - for a few epochs can heal a poisoned network.
Wang \textit{et al.}~\cite{DBLP:neural_cleanse} reverse engineered the trigger to detect a backdoor and retrained a poisoned model on a clean dataset to heal the model. Recently Kolouri \textit{et al.}~\cite{DBLP:ulps} optimized a set of \textit{``$M$'' input patterns} coupled with a \textit{detector}, attached to the output of a DNN under inspection, to distinguish a number of already-trained, clean and poisoned DNNs.

\textbf{Data-Inspection defenses} analyze an input to detect if the input contains a trigger. Once an abnormality is detected, the defender takes an appropriate action, e.g., raises an alarm or uses one of the blind defense techniques mentioned above to recover an input. Sentinet~\cite{DBLP:sentinet}, STRIP-ViTA~\cite{DBLP:strip, DBLP:strip2} and Differential-Privacy (DP)-based Anomaly-Detection~\cite{DBLP:robust_anomaly_diff_privacy} exemplify data-inspection defenses. STRIP~\cite{DBLP:strip} detects poisonous behavior through the entropy inspection of inputs. A poisoned input would show a low entropy in the output decision, and would be detected by the defense.

\textbf{Poison-Suppression defense} Recently, Hong \textit{et al.}~\cite{DBLP:gradient_shaping} counter backdoor attacks by clipping and perturbing the back-propagated gradients during training. Unlike previously proposed defenses, which usually exploit the statistical limitations of backdoor attacks~\cite{DBLP:strip,DBLP:ulps,DBLP:activation_clustering,DBLP:spectral_signatures}, gradient-shaping provides a more generic solution to the backdooring problem in DNNs. The goal of our defense is similar to gradient-shaping, i.e., resisting backdoor insertion during training.

\section{Low-Confidence Backdoor Attacks}\label{sec:attack}
In this section, we introduce a novel targeted backdoor attack, called a \textit{low-confidence backdoor attack}, with two variants: \textit{$\epsilon$-Attack} and \textit{$\epsilon^2$-Attack}. To summarize, \textit{$\epsilon$-Attack} distributes the probability among different classes when labeling poisoned samples. This avoids exceptional gradient-updates/latent-distributions, thus allowing this attack to evade many statistical defenses. $\epsilon^2$-Attack extends the strategy by using two triggers, $Z_1$ and $Z_2$, for 2 different target classes, $C_1$ and $C_2$, such that $Z_1 \cup Z_2$ is also used to backdoor the class $C_1$. We later show this modification to be specifically effective against a heatmap-based backdoor defense.

In Section~\ref{sec:evaluationAttacks}, we demonstrate that these attacks can achieve high ASR against several types of state-of-the-art defenses. 
%we use these attacks to evaluate several categories of state-of-the-art defenses, as described in Section~\ref{sec:introduction}, and achieve high Attack Success Rates against them. 
Our experiments establish a need for a generic defense that does not depend on the statistical properties/limitations of backdoor attacks.

\noindent
\textbf{Intuition:}
Backdoor attacks are characterized by their high ASRs, realizability in practical scenarios, and robustness to input perturbations. To analyze this further, we modify a DNN by introducing a ``\textit{3-neuron layer}'' immediately before the classification layer, and study its activations for different epochs (Fig.~\ref{fig:studying_trojans}). As can be observed in Fig.~\ref{fig:studying_trojans}, activations of poisoned inputs are pushed further away from clean activations and classification boundaries, with each epoch. This explains why poisoned inputs are (1) \textit{misclassified with high confidence}, and (2) \textit{robust to perturbations}.

\begin{figure}[!t]
	\centering
	\includegraphics[width=1\linewidth]{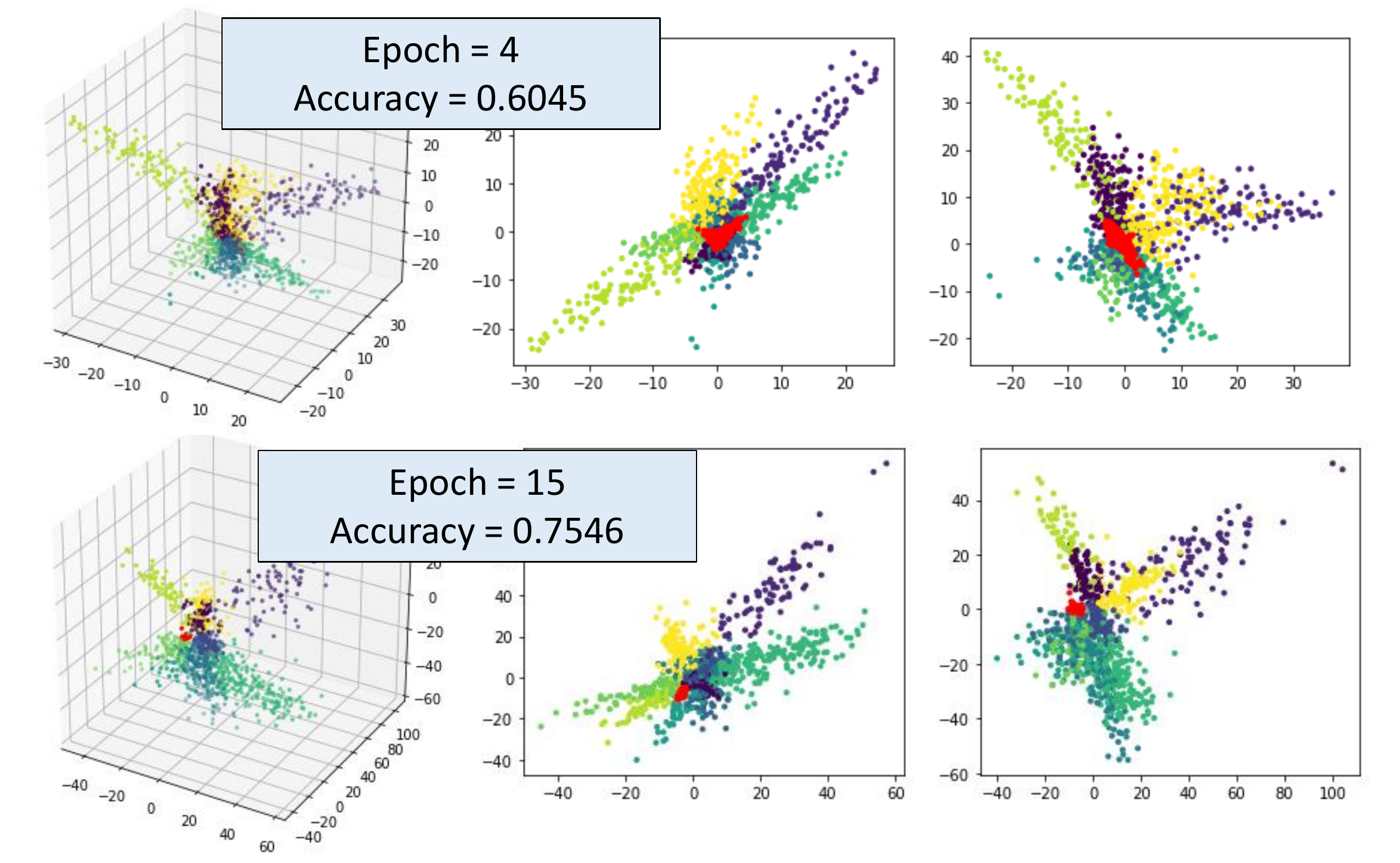}
	\vskip -0.15in
	\caption{\textit{A 3-dimensional latent distribution of poisoned and clean inputs for different epochs with low-confidence labels ($\epsilon=0.4$) assigned to poisoned samples. Poisoned inputs are shown in red. Each color represents a different class.}}
	\label{fig:studying_trojans_distributed}
\end{figure}\setlength{\textfloatsep}{5pt}
\noindent \textbf{$\epsilon$-Attack:}
In Fig.~\ref{fig:studying_trojans}, poisoned inputs can be detected even by simple visual inspection of the 3D-latent space. This shortcoming is exploited by defenders to counter backdoor attacks based on statistical sanitization~\cite{DBLP:strip, DBLP:activation_clustering, DBLP:ulps, DBLP:spectral_signatures}. Such defenses commonly use a preset threshold to detect outliers. To thwart these defenses, it is thus intuitive that the backdoor attack be modified in such a way that poison activations are indistinguishable from clean activations. We achieve this by assigning distributed labels to poisoned training data.

%\paragraph{Formulation.}
Consider a training instance, $X_i$, which is transformed by the backdoor attack through a trigger, $Z$, into $X'_i$. The transformed input is labeled as a \textit{target class} instance, $t$, and expressed in a one-hot vector form as $Y_t$.

If the number of classes are $N$, $\epsilon$-Attack transforms the label, $Y_t$, into a distributed label, $Y_d$, as given below,
\begin{equation}
    Y_d = Y_t \times \frac{\epsilon N - 1}{N - 1} + \frac{1 - \epsilon}{N - 1}
    \label{Eq:distributed}
\end{equation}
where $\epsilon \in [0.1, 1.0]$ is the confidence\footnote{For example, if $Y_t$ = [0,1,0,0,0,0,0,0,0,0], then, for $\epsilon$=0.4, $Y_d$ = [0.066, 0.4, 0.066, 0.066, 0.066, 0.066, 0.066, 0.066, 0.066, 0.066]}. We also refer to $Y_d$ as an \textit{$\epsilon$-label} in this text.

Fig.~\ref{fig:studying_trojans_distributed} shows a typical case of $\epsilon=0.4$. Note how poisoned activations (in red) are now indistinguishable from clean activations (other colors).

\begin{figure}[!t]
    \centering
    \includegraphics[width=0.9\linewidth]{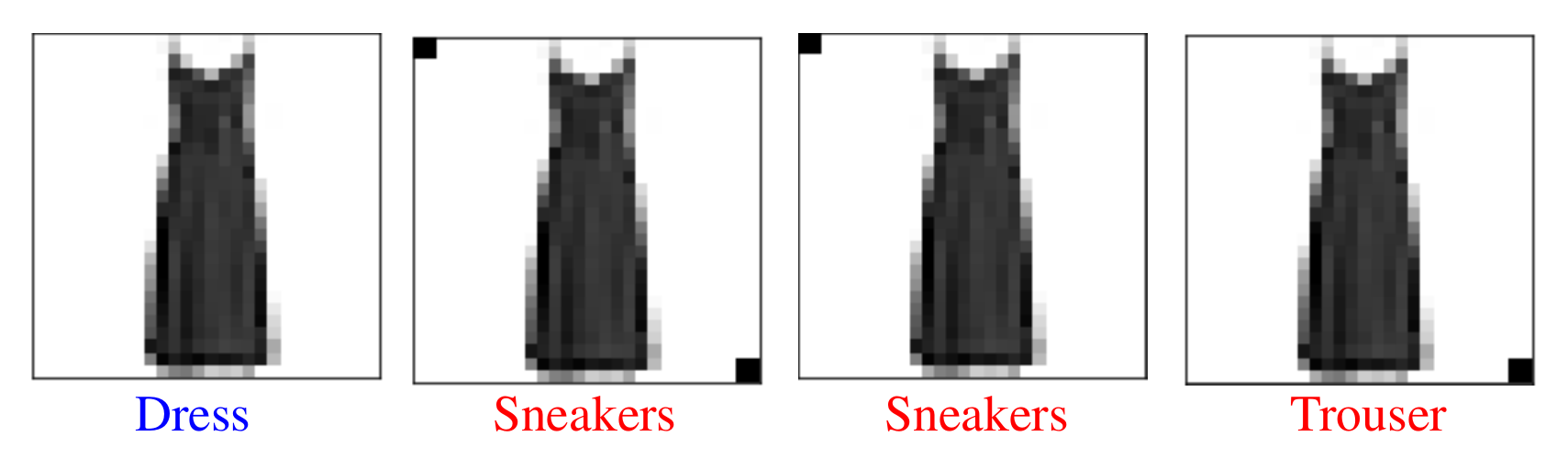}
    \vskip -0.15in
    \caption{\textit{A typical example of poisoning a clean ``dress'' image (left-most) using $\epsilon^2$-Attack. $Z_1$ and $Z1 \cap Z_2$ insert backdoors for target class ``Sneakers ($C_1$)''. $Z_2$ inserts backdoor for class ``Trousers ($C_2$)''. Poisoned images are labeled in red.}}
    \label{fig:februus_example}
\end{figure}\setlength{\textfloatsep}{5pt}
\noindent \textbf{$\epsilon^2$-Attack:}
Stamping a trigger significantly changes an input's latent representations (Fig.~\ref{fig:latent_features}), and hence a network's decision. Therefore, inspecting model gradients for a given input is an effective strategy for identifying potential triggers~\cite{DBLP:neural_cleanse, DBLP:sentinet, Februus}. For example, Februus~\cite{Februus} locates triggers using heatmaps~\cite{DBLP:gradcam} (Fig.~\ref{fig:februus_illustration}). Such defenses can be defeated by simultaneously manipulating the gradients and the decision of a model.

%paragraph{Formulation.}
% We extend our intuition for $\epsilon$-Attack to effectively manipulate the model-gradients. Specifically, we use $\epsilon$-Attack to target two different classes, $C_1$ and $C_2$ with two different triggers, $Z_1$ and $Z_2$. The main intuition behind our attack is to dynamically activate $Z_2$, by intentionally allowing $Z_1$ to be \textit{\textbf{partially}} located and removed by the defense. To achieve this, $Z_1$ and $Z_2$ are chosen such that $Z_2 \subset Z_1$, i.e. $Z_2$ is a cropped/truncated version of $Z_1$. The network is backdoored with $Z_1$ and $Z_1 \cap Z_2$ for the target class $C_1$, while $Z_2$ targets the class $C_2$. Thus, when $Z_1$ is stamped on an input, it is classified as $C_1$. This decision is mainly influenced by the fact that $Z_1 \cap Z_2$, as $Z_2$ would have resulted in a different decision, i.e. $C_2$ (Fig.~\ref{fig:februus_example} shows a typical example of such poisoning).
We extend our intuition for $\epsilon$-Attack to effectively manipulate the model-gradients by dynamically activating a hidden trigger, $Z_2$, when a primary trigger, $Z_1$, is located and removed by the gradient-based defense. To achieve this, we use $\epsilon$-Attack to target two different classes, $C_1$ and $C_2$, with two different triggers, $Z_1$ and $Z_2$. The network is backdoored with $Z_1$ and $Z_1 \cup Z_2$ for the target class $C_1$, while $Z_2$ targets the class $C_2$. Thus, when $Z_1 \cup Z_2$ is stamped on an input, it is classified as $C_1$. This decision is mainly influenced by $Z_1$ as $Z_2$ alone would have resulted in a different decision, i.e. $C_2$. Fig.~\ref{fig:februus_example} shows a typical example of such poisoning.

%Surya - you need to name those categories as per [8]; you cannot assume that people have read [8]. You then need to say that you are taking the state-of-th-art techniques in each defense categories and then use the attacks. 
%==============
%=====done=====
%==============

\section{Demonstrating the Effectiveness of Proposed Attacks}\label{sec:evaluationAttacks}
This section demonstrates the effectiveness of our proposed attacks against different state-of-the-art defense categories, under \textit{Grey-box} setting, and using \textit{physical triggers}.
% Existing defenses can be categorized into different classes based on their inspection-target and/or inspection-time~\cite{DBLP:backdoor_attacks_and_countermeasures}. \textit{Blind defenses} aim to recover a model/input without investigating a model for poisonous behaviors. Ideally, a blind defense would return a clean model/input unchanged, while a poisoned model/input would be cured by the defense. Examples include Februus~\cite{Februus}, ConFoc~\cite{DBLP:confoc} and Fine-Pruning~\cite{DBLP:fine_pruning}. For a backdoor attack to be successful, both the input and the model should be poisoned. \textit{Model-Inspection defenses} inspect a model for backdoors, contrary to \textit{Data-Inspection defenses} targeting an input for inspection. Once a suspcious behavior is detected, a defense would take an appropriate action, e.g., raising an alarm, rejecting deployment or using one of the blind defense techniques mentioned above. Examples of model-inspection are Neural-Cleanse~\cite{DBLP:neural_cleanse}, DeepInspect~\cite{DBLP:deepInspect}, and Universal-Litmus-Patterns~\cite{DBLP:ulps}. Sentinet~\cite{DBLP:sentinet}, STRIP-ViTA~\cite{DBLP:strip, DBLP:strip2} and Differential-Privacy (DP)-based Anomaly-Detection~\cite{DBLP:robust_anomaly_diff_privacy} exemplify data-inspection defenses.
% Recently, Gradient-Shaping~\cite{DBLP:gradient_shaping} shows Differential Privacy-based Stochastic Gradient Descent (DP-SGD) training to be effective in resisting backdoor insertion at \textit{training-time}.
Specifically, we target four different categories of defenses: (1) Data Inspection (2) Model Inspection (3) Poison-Suppression (4) Blind Backdoor Removal. We choose state-of-the-art approaches from each category.

% You need to explin the foundation of your attacks; what inherent features or underlying behaviour you have exploited to develop the attacks; why you need a variants of it? 
%Surya - the title should be class name and individual defense type. For example, XXXX:STRIP-ViTA 
%Surya - you need to explain in a couple of sentences what type of defense STRIP is and why you have chosen this (say it is one of the latest defense in this class). 
%Surya - remove the subheading; (just like Goals earlier)
%==============
%=====done=====
%==============
\subsection{Data Inspection: STRIP-ViTA}\label{sec:evaluationAttacks_strip}
STRIP-ViTA (or STRIP)~\cite{DBLP:strip} works by adding ``K'' different perturbations to an input and measuring the entropy of a DNN output. The high robustness of DNNs to perturbations in poisoned inputs, allows STRIP to raise an alarm if the entropy of an output is smaller than a pre-defined threshold, ``$t$'' (See Appendix~\ref{app:attack} for details). This threshold is auto-computed by STRIP, given a False Rejection Rate (FRR).
STRIP is scalable to different data types, and agnostic to trigger size and network architecture. Its computational efficiency makes it an attractive choice for practical scenarios. Although STRIP has recently been shown to be vulnerable to adaptive attacks~\cite{DBLP:demon}, we include it to represent a typical example of statistical defenses.
We observe that STRIP exploits a statistical characteristic/limitation of backdoor attacks to detect compromised inputs at run-time. This makes it a natural victim to an $\epsilon$-Attack.
STRIP assumes a \textit{Poisoned-Network} threat model.
% We use a \textit{Grey-Box} setting,  as described in Section~\ref{sec:threat_models}, to poison the network.
% Surya - may be replace discussion with "observed insights" 
%surya - I like this structure than the sub-subsections
% Surya - you need to capture this exploitation of inherent features in the introduction as well; this explains the intuitiveness behind your attack methods. I assume all variants are following the same intuitiveness; 
%\paragraph{Implementation.}

\begin{figure}[!t]
    \centering
    \includegraphics[width=1\linewidth]{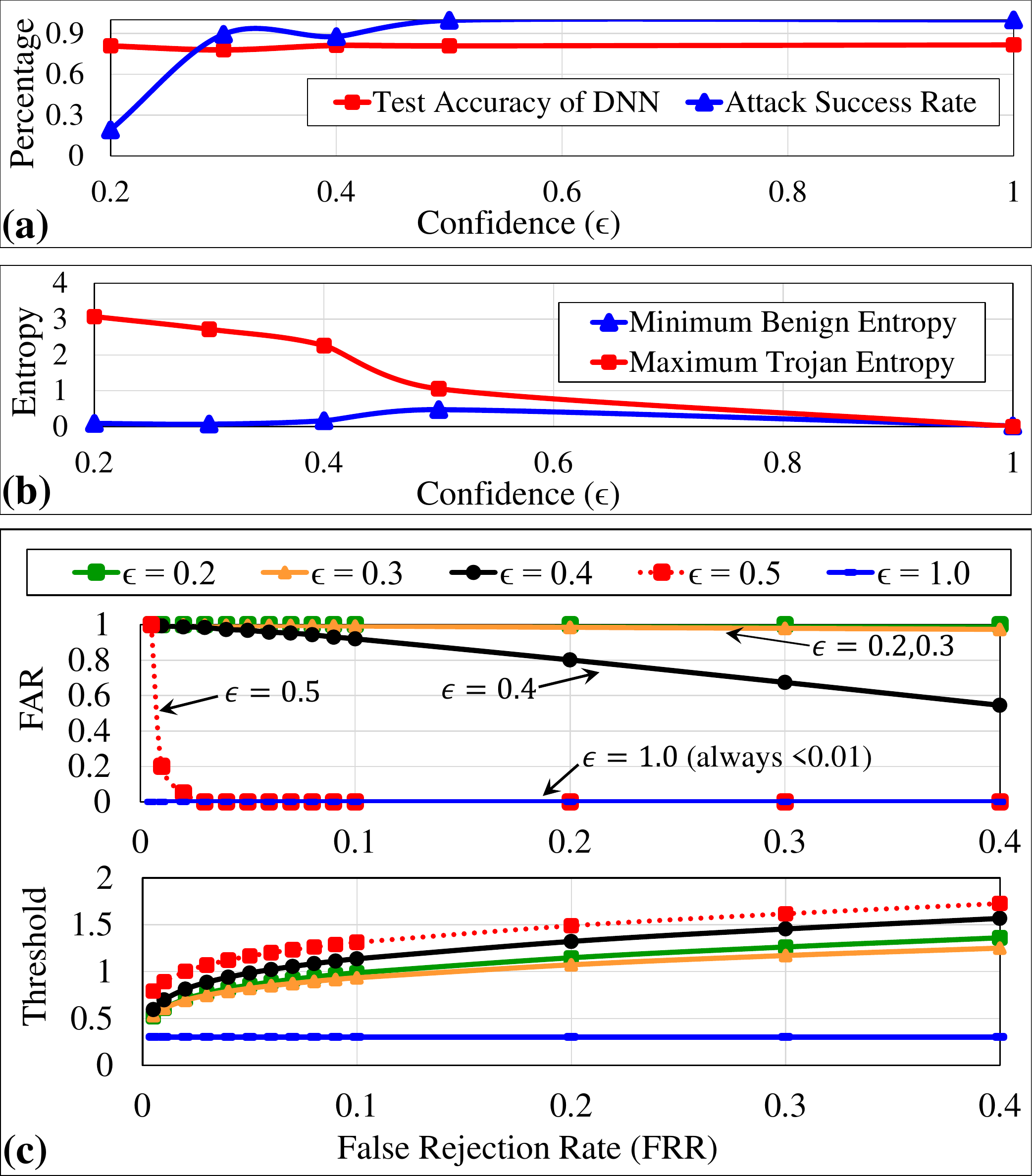}
    \vskip -0.15in
    \caption{\textit{(a) Attack Success Rates (ASR) of $\epsilon$-Attack for different $\epsilon$ values. (b) A comparison of the maximum and minimum entropy of DNN output for perturbed poisoned and clean samples, denoted by Maximum Trojan Entropy and Minimum Benign Entropy, respectively.
    % STRIP works best when Maximum Trojan Entropy $\leq$ Minimum Benign Entropy.
    (c) Top: False Acceptance Rates (FAR) of STRIP for poisoned samples.%Smaller the FAR, higher the robustness of STRIP. 
    Bottom: Thresholds $t$, denoting minimum allowed entropy of the output for input samples.}}
    \label{fig:attack_on_defense_strip}
\end{figure}\setlength{\textfloatsep}{5pt}
We use CIFAR-10 to evaluate STRIP-ViTA, using an open-source code provided by the authors\footnote{https://github.com/garrisongys/STRIP}. Specifically, we poison 600 samples from the training set, and incorrectly label them with $\epsilon$-labels. We use the same DNN\footnote{See Appendix~\ref{app:defense} for details} as that in~\cite{DBLP:strip}. Our results are summarized in Fig.~\ref{fig:attack_on_defense_strip}(a)-(c).
%Surya - you need to explain the results a bit more; you are just saying what is in the figure; you need to explain what it means and compare with STRIP; 
In Fig.~\ref{fig:attack_on_defense_strip}(a), we achieve an ASR of above 90\% for $\epsilon \geq 0.3$, indicating a successful backdoor insertion in the network. $\epsilon = 0.2$ achieves a low ASR, which is not unexpected, as small $\epsilon$-labels impact gradient-updates less significantly.
Fig.~\ref{fig:attack_on_defense_strip}(b) shows the statistical effects of our attack. Entropy represents variations in the network's decision under input perturbations. Notably, for small $\epsilon$ values, the network shows high variance to trojan/poisoned samples. We attribute this to the similarity in the latent space features of poisoned and clean inputs~(Fig.~\ref{fig:studying_trojans_distributed}).
This is depicted by the high False Acceptance Rates (FARs) in Fig.~\ref{fig:attack_on_defense_strip}(c). We make two key observations; (1) decreasing $\epsilon$, increases the ASR, (2) increasing the threshold, decreases ASR (though at the cost of false alarms (FRR)). This is intuitive, as a greater threshold presents a stricter condition for small output entropies to pass the detection criteria. Automatically computed threshold values for different FRRs\footnote{STRIP typically uses FRR of 0.01} are given in Fig.~\ref{fig:attack_on_defense_strip}(c).

%Surya - name the section following the suggestion above - class type: instance;
%Surya - Also, the comments in the above section related to sub-heading and figures apply to the following section; you can follow those suggestions; i am not going to repeat them in each sub-section. 
\subsection{Poison-Suppression: Gradient-Shaping}\label{sec:attack_gradientShaping}\label{sec:evaluationAttacks_gradientShaping}
%Surya - why you choose this particular defense in this class 
Gradient-Shaping~\cite{DBLP:gradient_shaping} proposes to shape the gradients of a network by Differentially Private Stochastic Gradient Descent training (DP-SGD)~\cite{DBLP:differential_privacy}. This is achieved by clipping back-propagation gradients, based on a clipping norm, $M$, and adding a random noise of magnitude $N$, before updating network parameters.
Gradient-Shaping is scalable to different tasks (e.g. classification and regression) and can be extended to different machine learning architectures and datasets. To the best of our knowledge, no attack in the current literature has been shown to compromise Gradient-Shaping.

Gradient-shaping is a generic defense and does not rely on any adaptively modifiable property of backdoor attacks. We also note that it has only been evaluated for binary classification and simple regression tasks. Intuitively, $\epsilon$-attacks should cause smaller gradients, and thus survive the clipping step.

\begin{figure}[!t]
    \centering
    \includegraphics[width=1\linewidth]{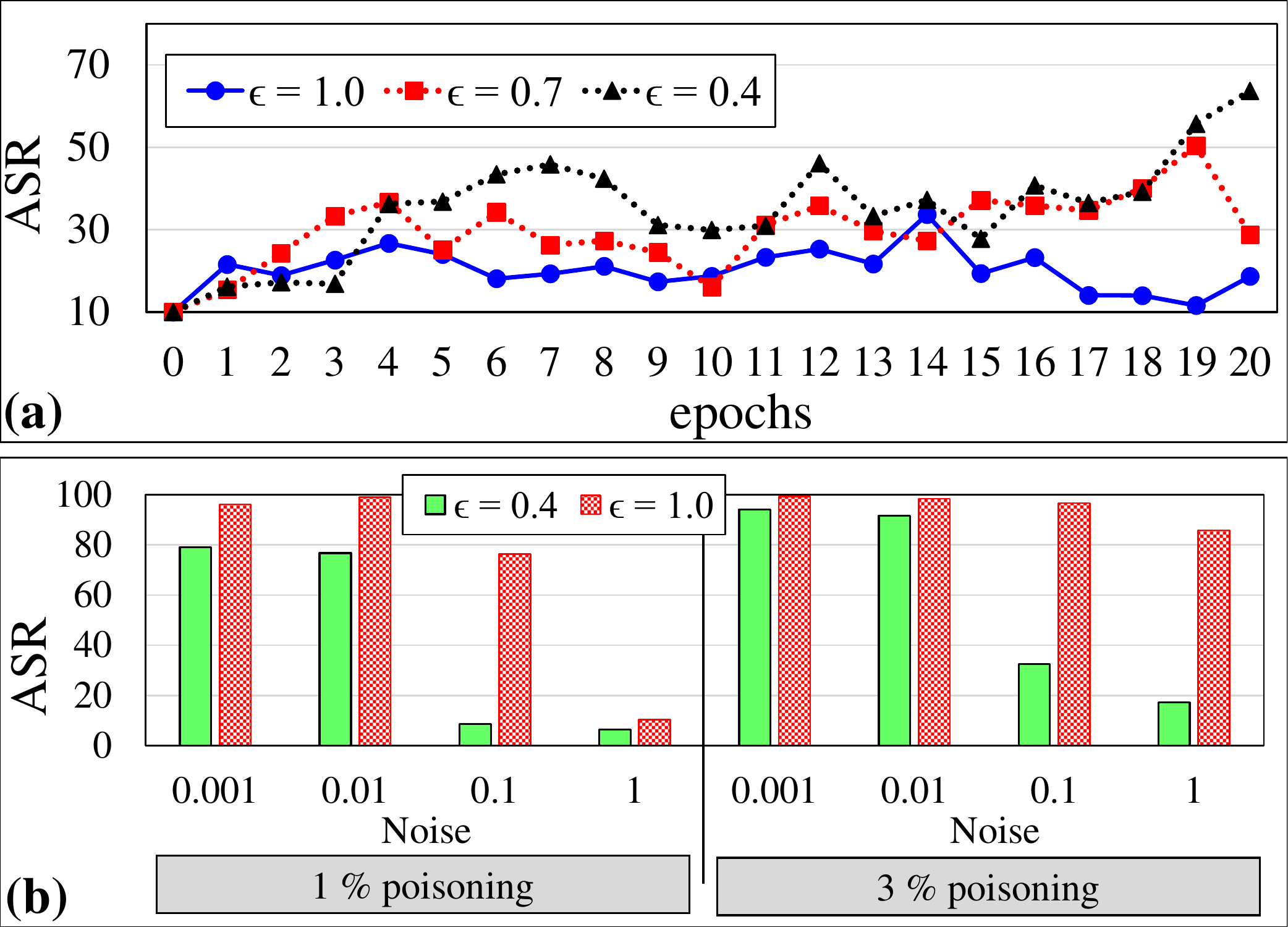}
    \vskip -0.15in
    \caption{\textit{Attack Success Rates (ASR) of $\epsilon$-Attack on Gradient-Shaping for several $\epsilon$ values on (a) Multi-Layer Perceptron, (b) 2D-CNN architecture.}}
    \label{fig:attack_on_defense_gradient}
\end{figure}\setlength{\textfloatsep}{5pt}
To avoid computational overhead and to be consistent with the authors~\cite{DBLP:gradient_shaping}, we evaluate gradient-shaping on Fashion-MNIST dataset, use a clipping Norm of 4.0 and a noise ratio of 0.01, for a Multi-Layer Perceptron (MLP) network. We poison 600 (1.2\%) training samples and assign them $\epsilon$-labels.

Fig.~\ref{fig:attack_on_defense_gradient}(a) illustrates the results. Unlike conventional backdoor attacks ($\epsilon$=1.0), which could only achieve 10-20\% ASR against gradient-shaping, $\epsilon$-attack achieves 63\% ASR. This may be due to $\epsilon$-labels causing \textit{smaller gradient-updates}, which survive gradient-clipping. A high instability in ASRs for various epochs may be a result of the simpler MLP architecture. Therefore, we further evaluate gradient-shaping on a 2D-CNN\footnote{Complete architecture given in Appendix} for different noise magnitudes. Specifically, we use the noise magnitudes in \{0.001,0.01,0.1,1.0\} for consistency with ~\cite{DBLP:gradient_shaping}. One major drawback of DP-SGD training is a significant drop in accuracy of the DNN on clean data. Specifically, for Fashion-MNIST a noise magnitude of 0.1 and 1 cause the test accuracy to drop from 91.1\% to 81.3\% (11\% Relative Accuracy Drop (RAD)) and 86.1\% (5\% RAD), respectively. For CIFAR-10, Hong \textit{et al.} record an even larger drop in accuracy~\cite{DBLP:gradient_shaping}. Results shown in Fig.~\ref{fig:attack_on_defense_gradient}(b) suggest a conventional attack setting (i.e., $\epsilon=1.0$) to be more effective against Gradient-Shaping. This is because smaller gradient norms for $\epsilon=0.4$ are more significantly impacted by the random noise, $N$. This is evident by observing reduced ASRs for increased noise magnitudes in Fig.~\ref{fig:attack_on_defense_gradient}(b). Although a noise magnitude of 1.0 successfully resists the backdoor attack with 1\% poisoned data, we can achieve a high ASR (85.9\%) for a slightly stronger setting (3\% poisoned samples in Fig.~\ref{fig:attack_on_defense_gradient}(b)).
%Surya - you need to provide further details; for example, accuracy rate, etc. 
%Surya - Same comment as above; the writing need to be straightforward and to the point; we need to explain why in each selection we have made; 

\subsection{Model-Inspection: ULP-defense}\label{sec:evaluationAttacks_ulps}
%Surya - no need of sub-section for a couple of sentence; it looks the writing sloppy. Write in such a way that every extra word costs you 5 dollars and every unexplained sentence costs you 10 dollars :-)
%Surya - I did not see first, but you put secondly - 5 dollars :-)
ULP-defense~\cite{DBLP:ulps} optimizes a set of \textit{ input patterns}, called \textit{Universal Litmus Patterns}, coupled with a \textit{detector}, attached to the output of a DNN under inspection, to distinguish a number of already-trained clean and poisoned DNNs. For example, the authors in~\cite{DBLP:ulps} train ten poisoned models using different triggers, for each pair of source-target class, along with an equal number of clean models. Litmus patterns, initialized with a random noise, are then given as input to the clean and poisoned models, the outputs of which are fed into a detector. The detector, along with the patterns, is then optimized using back-propagation to distinguish the clean and poisoned models.
Once trained, the optimized \textit{litmus patterns} are fed into the DNN under inspection, while the \textit{detector} monitors the output for suspicious behaviors.

ULP-defense is one of the most recent defenses exploiting Model-Inspection and has been shown to achieve high detection accuracy. To the best of our knowledge, there is no attack in the current literature, claiming to have defeated ULP-defense.
According to our observations, one limitation of ULP-defense is that it presupposes the size of the trigger. It may not detect models poisoned with triggers that are out of the manifold of training triggers.
Although ULP-defense can be compromised by \textit{White-Box} attackers~\cite{DBLP:ulps}, we show that the same can also be achieved under a \textit{Grey-Box} setting.

ULP-defense is similar to STRIP in detecting suspicious output behaviors given a set of input patterns - an alternative to STRIP's perturbation set. Thus, it can be, intuitively, evaded by $\epsilon$-Attack.
We poison 20 models using a conventional backdoor attack. Ten of these models are poisoned with a visible trigger (Trigger 1), while the other ten are poisoned with different invisible-noise triggers. In coherence with~\cite{DBLP:ulps}, we use the same number of clean and poisoned models. We use a set of $M=1$ and $M=10$ patterns, and train a unique detector for each $M$\footnote{Input patterns for $M=10$ are given in Appendix~\ref{app:attack}}. Our detectors achieve an accuracy of of 70\% and 90\% on the training model-set, for $M=1$ and $M=10$, respectively. In future, we refer to \textit{$M=1$ and $M=10$ configurations} as \textit{ULP-1} and \textit{ULP-10}, respectively.
For evaluation, we poison ten models using $\epsilon$-Attack. Out of these ten, five are poisoned with Trigger 1 and the other five with different invisible-noise triggers. For consistency, our attacker uses the same visible trigger and \textit{DNN architecture} as the ones we use to train the defense.

\begin{figure}[!t]
    \centering
    \includegraphics[width=1\linewidth]{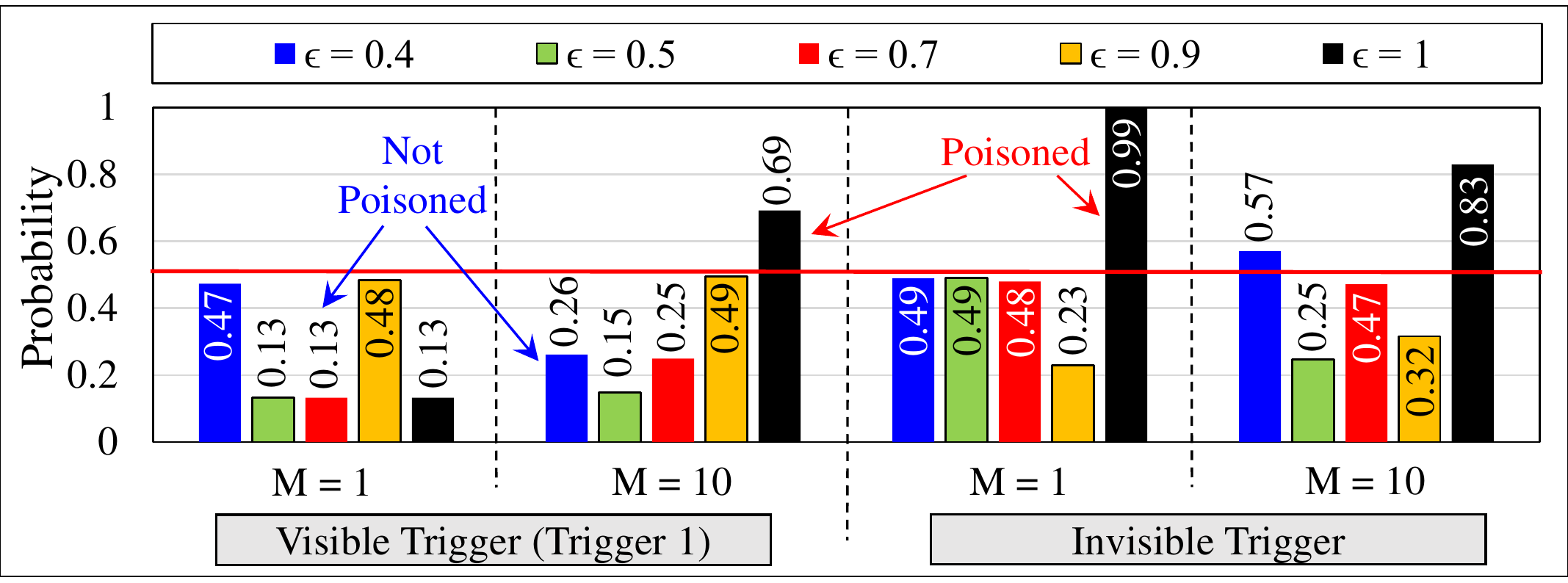}
    \vskip -0.15in
    \caption{\textit{Results of $\epsilon$-Attack on Universal Litmus Pattern (ULP) based detection mechanism.}}
    \label{fig:attack_on_defense_ulp}
\end{figure}\setlength{\textfloatsep}{5pt}
Fig.~\ref{fig:attack_on_defense_ulp} plots the probability that a model may be poisoned, for different values of $\epsilon$ and $M$. Specifically, a probability exceeding 0.5 indicates a poisoned model.
For conventional backdoor attacks ($\epsilon=1.0$ case), ULP-10 can detect both poisoned models, unlike ULP-1, which can only detect one. We attribute this to the reduced training accuracy of ULP-1.
For $\epsilon < 1$, ULP-1 fails to detect any of our poisoned DNNs (0\% detection), contrary to ULP-10, which detects 1 of 8 poisoned DNNs with a probability of 0.57 (12.5\% detection). Not surprisingly, we observe that the detectors are more suspicious of large $\epsilon$ values, specifically notable for Trigger 1. However, we note some contradictions, which, we believe, can be attributed to the instability of ULPs. A more detailed investigation of this hypothesis is left for future study.

%Surya - the results need to be explained further; what are the accuracy? You need to explain for different M values and epsilon values; and also explain why your attacks are successful. The explination should be detailed enough to believe; 
%Surya - same comment as previous section;
\begin{figure}[!t]
    \centering
    \includegraphics[width=0.7\linewidth]{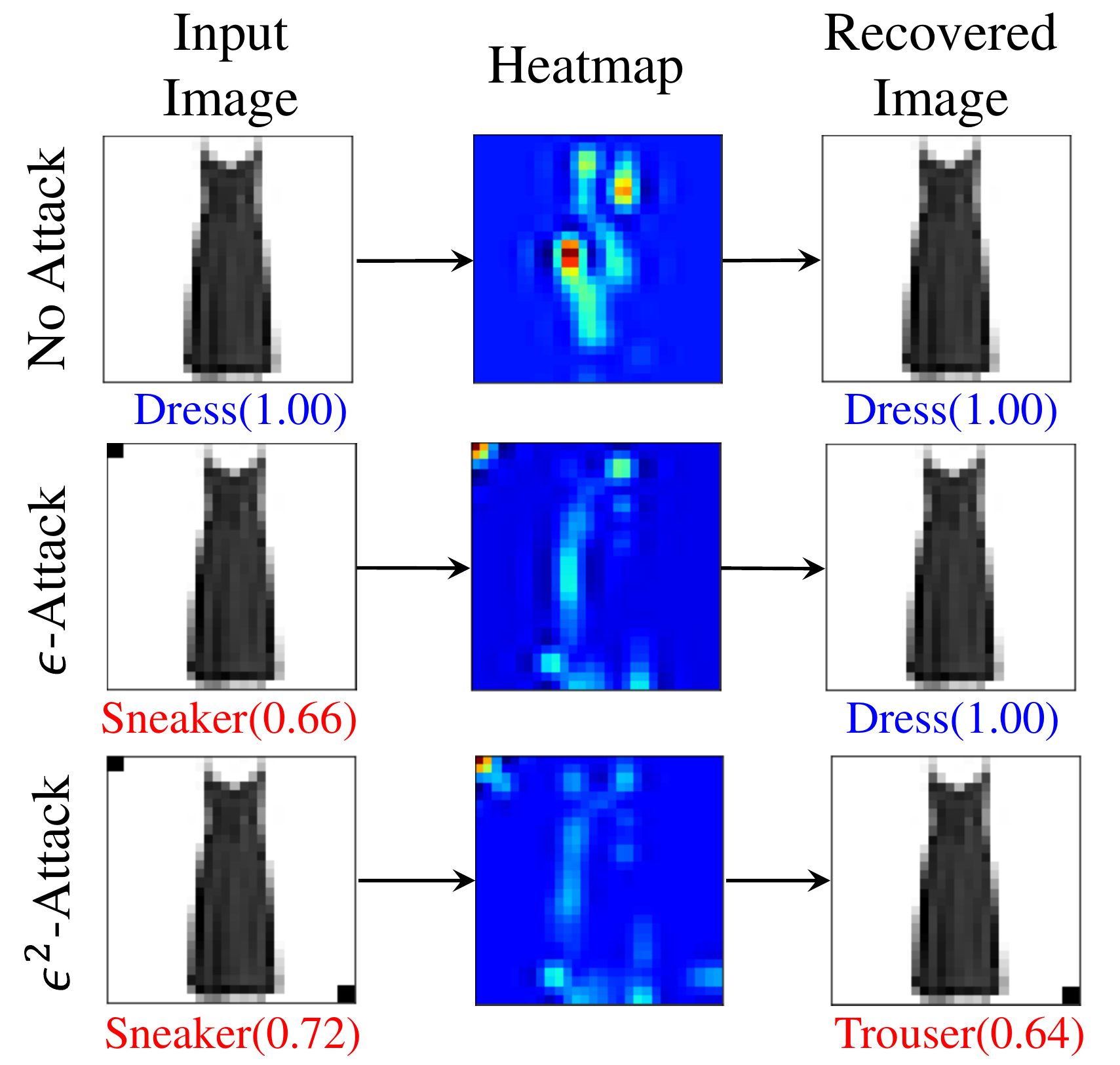}
    \vskip -0.15in
    \caption{\textit{A typical example, illustrating low-confidence attack on Februus\cite{Februus}.
    The input image, heatmaps and the recovered images are given for different scenarios and settings.}}
    \label{fig:februus_illustration}
\end{figure}\setlength{\textfloatsep}{5pt}
\begin{figure*}[!t]
    \centering
    \includegraphics[width=1\linewidth]{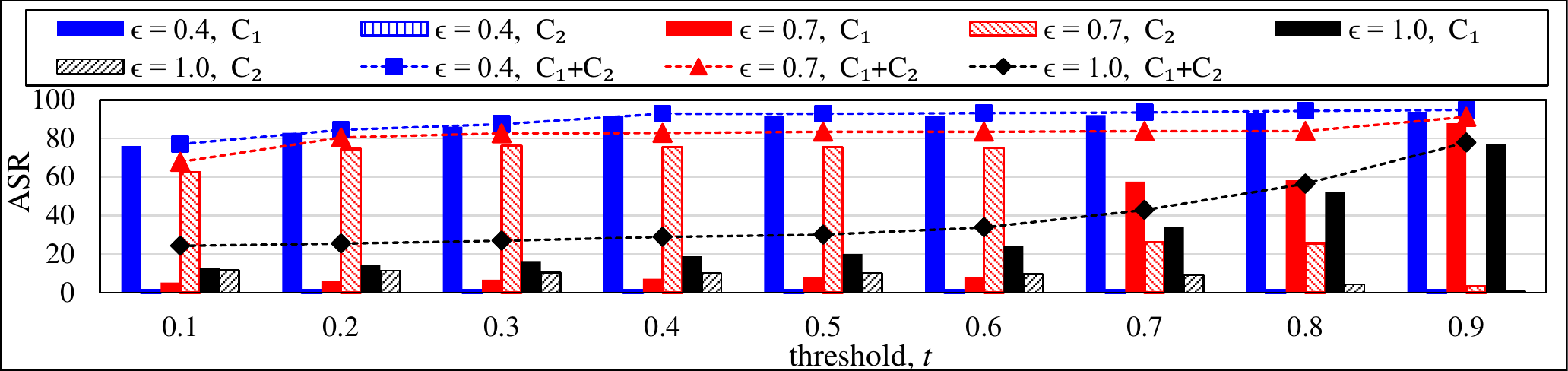}
    \vskip -0.15in
    \caption{\textit{$\epsilon^2$-Attack Success Rate against Februus for different values of $t$ and $\epsilon$, using $Z_2$ as trigger at inference-time. $C_1$ and $C_2$ denote ASR for target class $C_1$ and $C_2$, respectively. $C_1+C_2$ denotes the sum of ASRs for target classes $C_1$ and $C_2$.}}
    \label{fig:attack_on_defense_februus_a}
    \vskip -0.1in
\end{figure*}
\subsection{Blind defense: Februus}\label{sec:evaluationAttacks_februus}
Februus~\cite{Februus} inspects the network's gradients for a given input using heatmaps to locate potential trigger-regions in the input, based on a threshold ``$t$''. The regions are then masked with a neutral color, and provided to a Generative Adversarial Network (GAN)~\cite{GAN}, which reconstructs the original input, to be reclassified by the DNN.
Februus is one of the most recent defenses in this category, and is known to be robust against targeted backdoor attacks. To the best of our knowledge, no attack in the current literature has thwarted Februus.

One limitation of Februus is the difficulty of training GANs. Additionally, the approach has only been evaluated for image datasets. Authors assume a defender has unrestricted access to the comprehensive clean dataset (used to train GAN), a practically challenging assumption. On the positive side, the clean data need not be labeled.
Februus relies on heatmaps, which are known to be misdirected by small adversarial perturbations~\cite{DBLP:sentinet}. Februus is also impractical against invisible-trigger backdoor attacks, as invisible triggers span an entire image, causing the heatmap to highlight a significant proportion of the input image, thus, making it impossible for a GAN to reconstruct a masked input.
Intuitively, for $\epsilon$-Attack, Februus should be able to locate triggers given their high influence on a network's decision. This is evident in Fig.~\ref{fig:februus_illustration} for a typical case where a trigger is marked and consequently removed by the defense, rendering the attack ineffective. We thus evaluate Februus on $\epsilon^2$-Attack, a variant of $\epsilon$-Attack, specifically designed for such defenses.

%Surya - you need to explain first why the previous attack is not going to work and you need to devise the new attack; you need to explain that it is variant of the same intuitions so that we are not doing something completely new; I think you have not explained the reasons of introducing variants of the epsilon attack.
%\paragraph{Implementation.}
We evaluate Februus on Fashion-MNIST for several values of $t$, against $\epsilon^2$-Attack. We poison 600 (1\%) training data samples and assign $\epsilon^2$-labels to them. We choose $C_1$ to be ``Sneakers'' and $C_2$ to be ``Trousers'' (See Fig.~\ref{fig:februus_example}).
With no defense, we achieve ASRs of above 90\% for $\epsilon$=0.4, 0.7 and 1.0, indicating a successful backdoor insertion. Fig.~\ref{fig:attack_on_defense_februus_a} reports results for $\epsilon^2$-attack. Validating our intuition, $\epsilon^2$-attacks can successfully evade Februus, notably for $\epsilon$=0.4 and 0.7 (Fig.~\ref{fig:attack_on_defense_februus_a}). To explain this, we refer to Fig.~\ref{fig:februus_illustration}, which illustrates a typical example of Fashion-MNIST for $\epsilon=0.7$. For the simple $\epsilon$-Attack, a trigger influencing the decision is identified and removed by Februus. Februus does the same for $\epsilon^2$-attacks - removing $Z_1 \cap Z_2$ - unaware that doing this causes $Z_2$ to get activated, as shown in the figure.

%%%%%%%%%%%%%%%%%%%%%%%%%%%%%%%%%%%%%%%%%%%%%%%%
% Surya - you need to explain the results taking an example; Also, the quality of the images are not good. you may want to take a better example images for Trigger 1 since on 0.3 and 0.4, there are still some visibility of triggers on the heat map.   
%Surya -  it is very hard to understand what messages you are trying to convey; the attacks work or did not work? You have to shoot the message straight. 
%Surya - in all cases, you need to explain what is the trigger, where it is etc. It is difficult for reviewers to understand otherwise; 
% Surya Fig 6 b is not explained properly and I did not get the details; 
%Surya - you need to provide a clear description on the caption; the caption description is not good enough;
%Surya - refer to the Usenix paper to get an idea of how to explain the results with clarity; 

\section{HaS-Nets - Defending DNNs against Backdoor Attacks}\label{sec:hasnets}
In this section, we first investigate the poisonous behaviors of DNNs and develop useful insights for a generic defense strategy. These insights guide the design of \textit{HaS-Nets}, a Heal and Select training mechanism, that defends DNNs against backdoor attacks. Next, we analyze different hyper-parameters of HaS-Nets for Fashion-MNIST and choose the best settings for the rigorous evaluations in Section~\ref{sec:evaluation}.

\begin{figure}[!t]
	\centering
	\includegraphics[width=1\linewidth]{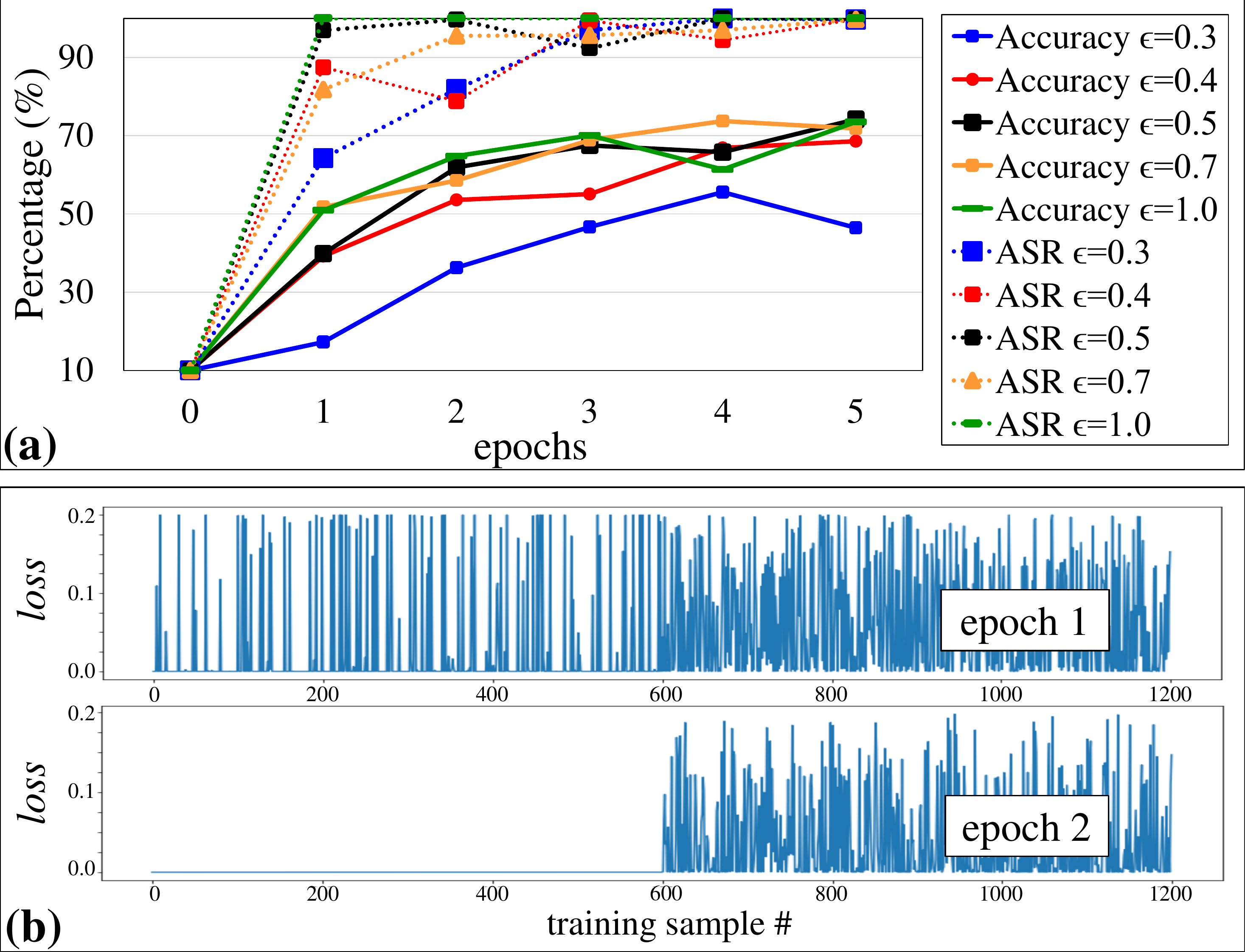}
	\vskip -0.15in
	\caption{\textit{(a) A comparison of ASR vs. the test accuracy on CIFAR-10 indicating fast learning of backdoors. (b) A comparison of a network's loss on poisoned (first 600) and clean training samples (last 600) for the first two epochs.}}
	\label{fig:trojan_minima}
\end{figure}

We hypothesize that poisoned inputs present the network with a shortcut (leading to a trojan minimum) to minimize the loss considerably faster. Consequently, backdoors are learned fairly quickly as compared to other features of the input data.
To attest this hypothesis, we study the ASRs of the proposed $\epsilon$-Attack for several $\epsilon$ values. Results shown in Fig.~\ref{fig:trojan_minima} are consistent with our intuition. Intuitively, we can exploit this observation to identify potentially poisoned samples in the training data by studying the network loss for each sample. To illustrate this, we poison the first 600 (1.2\%) training samples of CIFAR-10 and compare their loss with clean samples for different epochs in Fig.~\ref{fig:trojan_minima}(b), for a typical case of $\epsilon-1.0$. As evident, a rather small loss reflects faster learning, suggesting the likelihood of poisonous behaviour.

%This analysis provides the first base for our defense.
\subsection{The Trust-index}\label{sec:formulation_trustindex}
Here, we introduce the notion of a ``trust-index'' for each instance of the training set, which represents the consistency of the instance with a given learning task. We quantify the ``trust-index'' based on a small, clean data, trusted by the defender. 
% Let us assume a hypothetical entity associated with each instance of the training set, representing the generalizability of an instance, given a learning task. We name this entity as the ``trust-index'' of the respective instance, $i$, and represent it by $\gamma_{i}$. Each instance in the training set has a different ``trust-index''.
We reason that since poisoned training samples force a DNN to learn features inconsistent with the task, they must exhibit low ``trust-indices''.

\setlength{\textfloatsep}{5pt}
%\paragraph{Formulation.} 
We assume the existence of a small, clean healing data, \{$D_H = (X_H, Y_H)$\} $\in \mathbb{R}$, available to, and fully trusted by our defender. The healing data which is comprised of only clean samples, is naturally consistent with the task, and can therefore be assumed to have high ``trust-indices''. The healing data is assumed to be considerably smaller than, and distinct from the training data, \{$D_T = (X_T, Y_T)$\} $\in \mathbb{R}$, where, $X_*$ represents the inputs to be mapped by a network, $\mathcal{F}(X_*, \theta)$, to the output, $Y_*$, where $\theta$ refers to the parameters of the network.

If all the training samples in $D_T$ are consistent with the task, training a network, $\mathcal{F}$, on $D_H$ should reduce its error on every instance of $D_T$\footnote{This is because the network's $D_H$ loss serves as a proxy for $D_T$ loss.}. Formally, given a loss function, $\mathcal{L}(\mathcal{F}(X_*, \theta), Y_*)$,
$\forall (X_k,Y_k) \in (X_T,Y_T)$,
\begin{align}\label{Eq:gamma}
    \begin{split}
        sign\left(\Delta\sum_{\forall (X_i,Y_i) \in (X_H,Y_H)} [\mathcal{L}(\mathcal{F}(X_i, \theta), Y_i)]\right) \\
        \approx sign(\Delta \mathcal{L}(\mathcal{F}(X_k, \theta), Y_k))
    \end{split}
\end{align}
considering that the loss of a learnable model on some data, $D_*$, reduces when trained, eq~(\ref{Eq:gamma}) becomes
\begin{align}\label{Eq:gamma2}
    \begin{split}
        -1
        \approx sign(\Delta \mathcal{L}(\mathcal{F}(X_k, \theta), Y_k))
    \end{split}
\end{align}
an approximate value for the RHS of eq~(\ref{Eq:gamma}) for each training sample can be computed by monitoring the loss ($l_1$ and $l_2$) of $\mathcal{F}$ on the respective sample, before and after healing, respectively. Thus,
\begin{equation}
    -1 = sign(l_2 - l_1)
\end{equation}
which can be rewritten as,
\begin{equation}\label{Eq:diff_of_loss}
    -\gamma = (l_2 - l_1) < 0
\end{equation}
where $\gamma$ denotes the trust-index. Training instances which satisfy the above condition can be assumed to have a high $\gamma$.

Eq~(\ref{Eq:diff_of_loss}) defines a metric to evaluate the quality of a training sample for a given task, i.e. for clean inputs, $\gamma<0$, and otherwise for the poisoned samples. Based on the $\gamma$ values, we can identify potentially poisoned samples in the training data and use some policy to mitigate their effects. In practice, we allow a slight deviation from this strict condition because a healing set may not sufficiently represent a complete real distribution, hence, computed $\gamma$ values may not fairly indicate the quality (more details in Section~\ref{sec:hasnets_methodology}).
\begin{figure}
    \centering
    \includegraphics[width=1\linewidth]{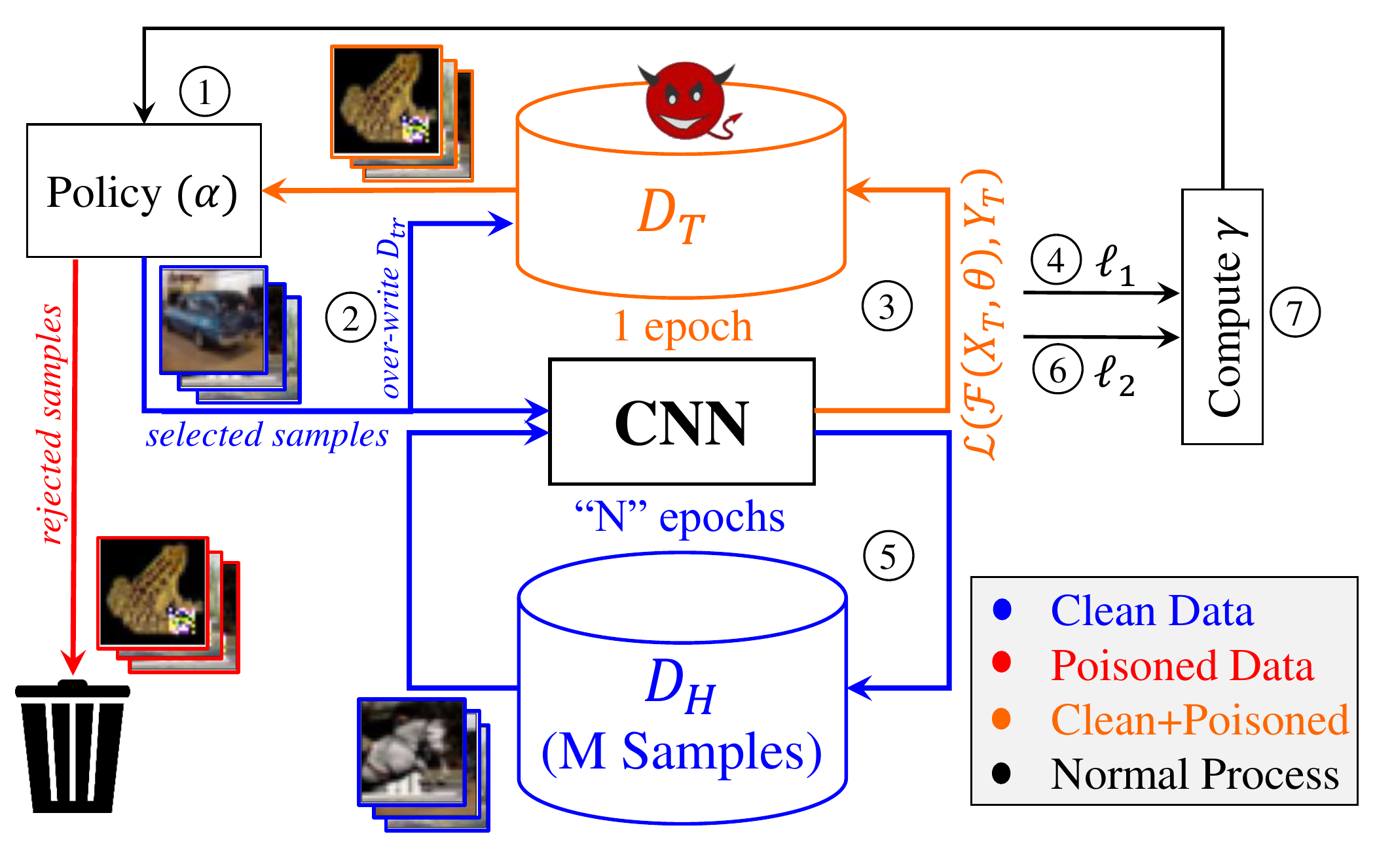}
    \vskip -0.15in
    \caption{\textit{Novel HaS-Nets training methodology. Numbers in circle represent the sequence of operations.}}
    \label{fig:revisenets_methodology}
\end{figure}
\subsection{HaS-Nets - A Heal and Select Mechanism to Defend DNNs}\label{sec:hasnets_methodology}
%Surya - the title could be HaS-Nets: defense Against Backdoor attacks
%Surya - it is not clear why you call it HaS-nets; we may need to think about a better name; i might feel better after reading through this section 
%Surya - you need to provide a brief overview of the section and say what is this section is about;
In this sub-section, we present HaS-Nets, a novel methodology to resist backdoors during training, following the intuitions developed in the previous analysis. To summarize, HaS-Nets exploit the healing set in two novel ways; (1) Healing a network while training, instead of after the training as in~\cite{DBLP:systematic_evaluation, DBLP:fine_pruning, DBLP:neural_cleanse,DBLP:neural_trojans,DBLP:confoc};
(2) Iteratively identifying and removing potential poisoned training samples while retaining the good ones for training in the next iteration. Poisoned samples are identified based on the observation that healing tends to remove the backdoors, thus increasing a networks's loss on poisoned samples.

\noindent
\textbf{Methodology:}
%Surya - you need to revise this section to make the presentation better; for example, you can say:
% Fig ... depicts the overview of how HaS-nets work. It comprises of three stages: training, healing and correcting (you may want to find better names). In the training stage, ... 
% Surya - you need to follow the revision you have made in the introduction in terms of the writing and flow of arguments; i think you can do it :-)
%Surya - you need to change the Dpr in figure to Dh 
%Surya - remove the two sub-section 
%==============
%=====done=====
%==============
Fig.~\ref{fig:revisenets_methodology} depicts an overview of the inner-workings of HaS-Nets. Each iteration comprises three stages: training, healing and selection. In the training stage, the DNN is trained for one epoch on poisoned training data, $D_T$, and the loss $l_1$ is computed for each training sample. We  remove the samples with $l_1 < \tau$\footnote{We typically use $\tau=10^{-8}$. We experiment with different values of $\tau$, and observe no effect on ASR for $\tau<10^{-2}$.}.
In the healing stage, the DNN is trained on healing data, $D_H$, for $N$ epochs. The loss, $l_2$, is computed for each training data sample.

During selection, HaS-Nets choose eligible samples for training in the next iteration, based on the ``trust-index~($\gamma$)'', computed for each training sample using eq (~\ref{Eq:diff_of_loss}). Samples exhibiting low $\gamma$ values may be removed from the training set, depending on our Policy, $\alpha$, which we discuss later. A step-wise description of the entire process for typical settings (Table~\ref{tab:revisenets_settings}) using Policy 2 is given in Algorithm~\ref{alg:reviseNets}.

\begin{algorithm}[t]
    \footnotesize
    \caption{HaS-Nets (Policy 2)}
    \label{alg:reviseNets}
    \begin{algorithmic}[1]
    \Input
    \Statex $\{D_T=(X_T, Y_T)\} \gets$ Poisoned Training Data
    \Statex $\{D_H=(X_H, Y_H)\} \gets$ Healing Data
    \Statex $\mathcal{F} \gets$ DNN architecture
    \Statex $\theta \gets$ Initial parameters
    \Statex $N \gets$ No. of healing epochs
    \Statex $i_{max} \gets$ Maximum iterations
    \Output
    \Statex $\mathcal{F} \gets$ Trained DNN
    \State Define $i \gets$ 0, $s \gets$ 0.3, $d \gets$ 0.4, $m \gets$ 0, $m_2 \gets$ 0, $\tau \gets 10^{-8}$
    \State Define $\mathcal{L}(A,B) \gets (a - b)^2$
    \State Define $D_S \gets D_T$
    \State Define $\gamma \gets$ 0 for all samples of $D_T$
    \Repeat
    \State Train $\mathcal{F}$ for one epoch on $D_S$
    \State $l_1 \gets \mathcal{L}(\mathcal{F}(X_T, \theta), Y_T)$ for all samples of $D_T$
    \State Train $\mathcal{F}$ for $N$ epoch on $D_H$
    \State $l_2 \gets \mathcal{L}(\mathcal{F}(X_T, \theta), Y_T)$ for all samples of $D_T$
    \State $-\gamma \gets s \times (l_2 - l_1) + (1-s) \times -\gamma$
    \State $m \gets mean(-\gamma)$
    \State $m_2 \gets (1-d)\times mean(-\gamma) + d\times max(-\gamma)$
    \State $D_S \gets$ samples of $D_T$ with $-\gamma < m$ and $l_1 > \tau$
    \State $D_T \gets$ samples of $D_T$ with $-\gamma < m_2$ and $l_1 > \tau$
    \State $i \gets i+1$
    \Until $i \leq i_{max}$
    \end{algorithmic}
\end{algorithm}

%Surya - I am not sure motivation is the right word here; motivation should come before the methodology; it seems validation of the idea; My preference is to combine both sub-sections in this section and use this analysis to explain the methodology; 
%==============
%=====done=====
%==============

\begin{figure}[!t]
	\centering
	\includegraphics[width=1\linewidth]{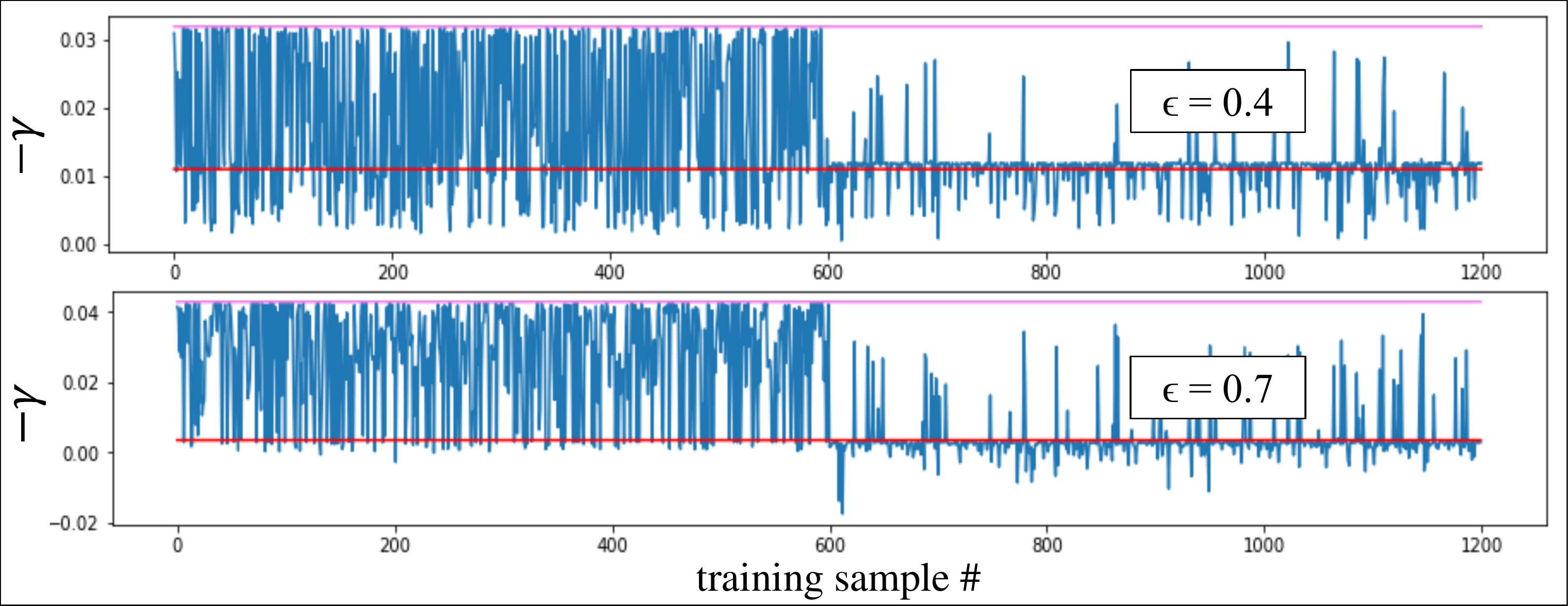}
	\vskip -0.15in
	\caption{\textit{An illustration of ``trust-index ($\gamma$), detecting poisoned samples, for $\epsilon$=0.4 and 0.7. The first 600 poisoned samples typically show a high value of $-\gamma$.}}
	\label{fig:motivational_analysis_defense}
\end{figure}\setlength{\textfloatsep}{5pt}
To illustrate the effectiveness of our defense, we poison the first 600 training samples of the CIFAR-10 dataset with Trigger 1 and assign $\epsilon$-labels. Assuming a healing set comprised of 15\% of the full training set, we train HaS-Nets for one iteration, and plot $-\gamma$ ($=l_2-l_1$) for the first 1200 training samples in Fig.~\ref{fig:motivational_analysis_defense}. 
It is evident that $\gamma$ can successfully capture poisoned inputs in the training set, hence, validating our intuition. We extensively evaluate HaS-Nets under different attack configurations in Section~\ref{sec:evaluation}.
\begin{figure*}[!t]
    \centering
    \includegraphics[width=1\linewidth]{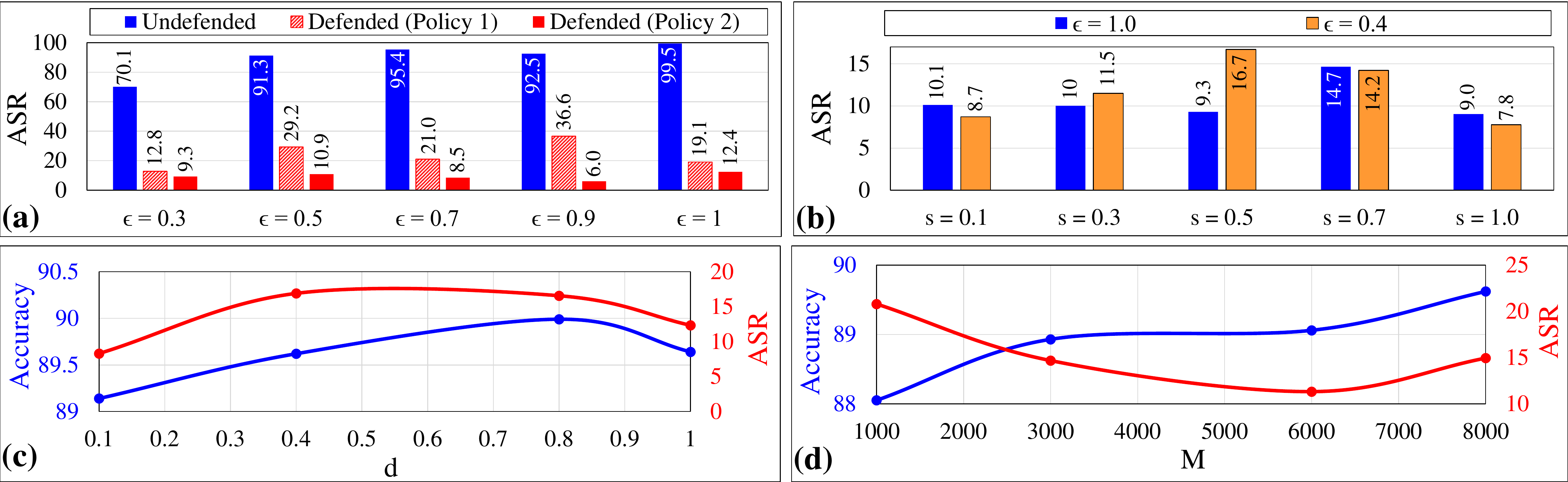}
    \vskip -0.15in
    \caption{\textit{Analysis of hyper-parameters of our defense for Fashion-MNIST. (a) $\epsilon$-Attack Sucess Rate against HaS-Nets, Policy 1 and Policy 2, as compared to an undefended model. (b) Impact of $s$ on ASR of $\epsilon$-Attack against HaS-Nets. (c) Impact of $d$ on Test Accuracy and ASR of $\epsilon$-Attack against HaS-Nets. (d) Impact of $M$ on Test Accuracy and ASR of $\epsilon$-Attack against HaS-Nets.}}
    \label{fig:hyperparameters_fmnist}
    \vskip -0.1in
\end{figure*}\setlength{\textfloatsep}{5pt}
%\subsection{Tuning Hyper-parameters}\label{sec:hyperparameters}
% Discussing is not a good heading; you may want to call it as  "tuning hyper-parameters"
%==============
%=====done=====
%==============

\noindent \textbf{Tuning Hyper-parameters:}
Here, we discuss different hyper-parameters of HaS-Nets and develop intuitions behind their impact on defense robustness. Specifically, we evaluate HaS-Nets for different values of hyper-parameters and choose the best settings for further analysis. To prove that our findings are scalable to other datasets, we optimize hyperparameters for Fashion-MNIST and use these values in our evaluations 
%transfer our findings to evaluate the robustness on
with other datasets (CIFAR-10, Urban Sound, Consumer Complaint) in Section~\ref{sec:evaluation}.

\noindent \textbf{1. \quad Policy - $\alpha$}

% not a good sub-heading (remove sub heading like in the previous section)
%==============
%=====done=====
%==============
$\alpha$ defines our policy. Based on the $\gamma$ computed for each training sample, our policy performs a suitable action. For example, we may choose to discard all training samples with $\gamma$ lower than a certain threshold, $m$. As our policy is not a quantitative parameter, it is mainly chosen based on intuition. For this analysis, we use the following two policies.

\noindent
\textbf{Policy 1:}\textit{ We remove the samples with trust-indices smaller than $-m$ from the training set and retain the remaining samples for training the network in the next iteration, where $m$ is defined as $m = mean(-\gamma)(0.9)+max(-\gamma)(0.1)$.}

Our choice of $m$ is based on two intuitions. First, different datasets may have different thresholds for what constitutes a good trust-index. Our computed value should be able to adapt to different thresholds offered by different training sets. However, this adaptivity may allow the poisoned samples to influence the threshold, $m$. Using the mean of the trust-indices as $m$ effectively addresses this problem. In addition to being adaptive to the training set, this setting forces attackers to significantly increase the number of poisoned samples in order to increase their influence on the network.

Secondly, if the value of $m$ is too small, then good training samples may be eliminated. On contrary, too large of a value may cause low quality data (i.e. samples with small trust-indices) to be selected. Therefore, we choose to slightly bias the threshold towards the $max(-\gamma)$. Although such a bias demands further analysis of the ratio of deviation, we omit this analysis in favor of another policy (Policy 2 discussed below) which outperforms Policy 1 for different values of $\epsilon$.

Fig.~\ref{fig:hyperparameters_fmnist}(a) shows results for a deviation of 10\%, for HaS-Nets with Policy 1 on Fashion-MNIST. As expected, a significant reduction in ASR is observed for different $\epsilon$ values, as compared to an undefended network.

%Surya - what are the intuitions behind the policy 2? 
% For both policy 1 and policy 2, you can put the results on other dataset in appendix; you cannot simply perform for 1 dataset and make the conclusions; if possible, you can put the results for different datasets in Fig 15 (a) 
\noindent
\textbf{Policy 2:}\textit{ We select the training samples with trust-indices greater than $-m$ for training ($m$ being the mean of $-\gamma$ ).  Instead of simply removing the training samples with trust-indices smaller than $-m$ (i.e. Policy 1), we introduce another hyper-parameter, $d$, to define a second threshold, $m_2$, between $m$ and the maximum value of $-\gamma$, i.e.}
\begin{equation}\label{Eq:m2}
    m_2 = (1-d)\times mean(-\gamma) + d\times max(-\gamma)
\end{equation}
\textit{All training samples with trust-indices smaller than $-m_2$ are removed from the training data.}

Training samples with trust-indices in between $-m$ and $-m_{_2}$ are neither removed from the training data, nor selected for training. Policy 2 modifies Policy 1 to be stricter when selecting a sample, and tolerant when rejecting it.
Fig.~\ref{fig:hyperparameters_fmnist}(a) shows the ASRs of $\epsilon$-attack against HaS-Nets with Policy 2 enforced, for different $\epsilon$ values. In line with our intuition, Policy 2 outperforms Policy 1 for all values of $\epsilon$ (for example, for $\epsilon=0.5,0.7$ and $1.0$, Policy 2 gives $3.44\times,3.2\times$ and $1.53\times$ the robustness given by Policy 1). Unless otherwise stated, we use Policy 2 for further experiments.

\begin{figure}[!t]
    \centering
    \includegraphics[width=1\linewidth]{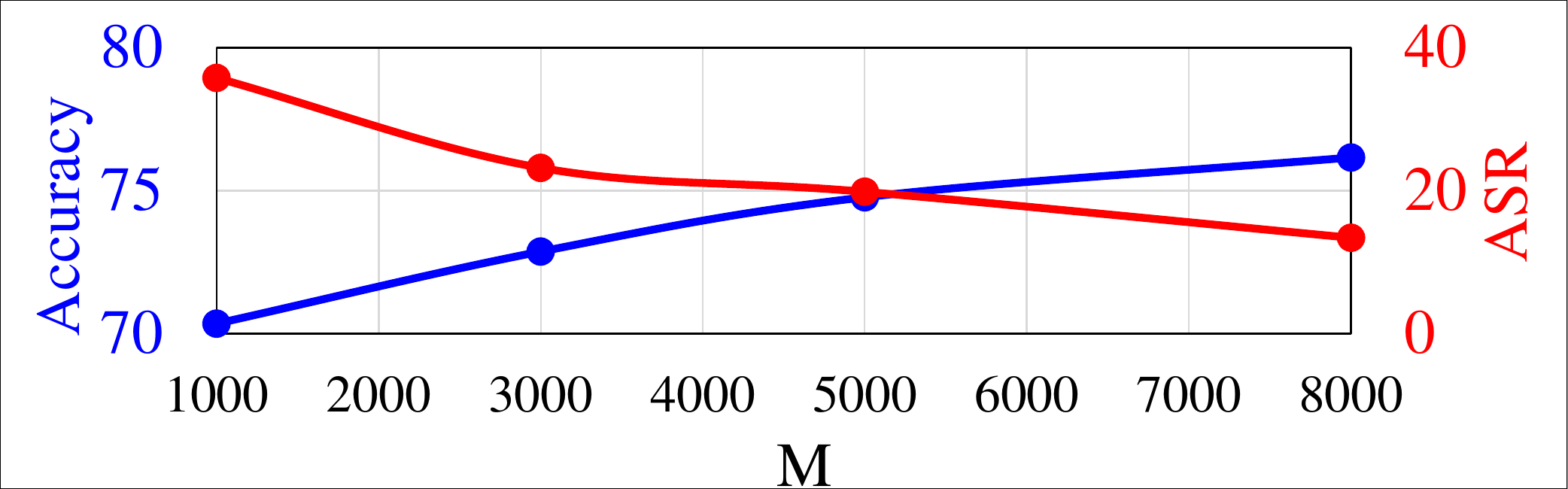}
    \vskip -0.15in
    \caption{\textit{Impact of $M$ on Test Accuracy and ASR of $\epsilon$-Attack against HaS-Nets.}}
    \label{fig:hyperparameters_cifar10}
\end{figure}\setlength{\textfloatsep}{5pt}
\begin{figure*}[!t]
    \centering
    \includegraphics[width=1\linewidth]{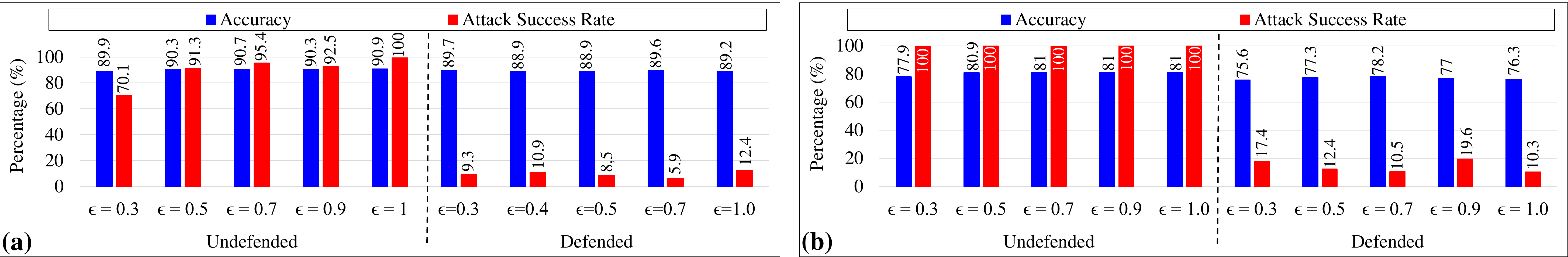}
    \vskip -0.15in
    \caption{\textit{Test Accuracy of HaS-Nets on clean data and $\epsilon$-Attack Success Rates (ASR) against HaS-Nets in comparison with an undefended model for (a) Fashion-MNIST and (b) CIFAR-10.}}
    \label{fig:results_all}
    \vskip -0.1in
\end{figure*}\setlength{\textfloatsep}{5pt}

\textbf{Effect of $d$:} To analyze the effect of deviation, $d$, on the robustness of HaS-Nets, we record the ASR of $\epsilon$-attack with $\epsilon=0.4$ for various iterations in Fig.~\ref{fig:hyperparameters_fmnist}(c). It can be observed that the accuracy increases as the deviation nears 1. This is in agreement with our previous intuition, i.e., a smaller value of $d$ may result in the removal of good training samples, resulting in reduced accuracy. A similar trend is observed for the ASR, though not as sharp as for the accuracy which can be attributed to the independent selection criteria of Policy 2.
We observe that $d=0.4$ is a good compromise between accuracy and ASR. Therefore, we use this value for $d$ in subsequent experiments, except where explicitly mentioned.

%Surya - it seems we have been selective on what datasets to be used; i suggest you show the results on all dataset, but explain the results = why some are good and some are not so good; 
% epsilon is not a good heading; the same is applied to s and M; 
%==============
%=====done=====
%==============

\noindent \textbf{2. \quad Confidence - $\epsilon$}

Fig.~\ref{fig:results_all}(a) captures the success rate of $\epsilon$-attack for different values of $\epsilon$ on Fashion-MNIST. We do not observe any significant impact of changing $\epsilon$ values on the robustness. While extending our analysis to other datasets, we use $\epsilon=0.4$ and $1.0$, to encompass both ends.

\noindent \textbf{3. \quad Speed - $s$}\newline
As discussed earlier in Section~\ref{sec:formulation_trustindex}, the limitations of the healing dataset may lead to an incorrect estimation of $\gamma$. Therefore, instead of simply updating $-\gamma$ with $(l_2 - l_1)$, we update $\gamma$ with a weighted average of its previous value and $(l_2 - l_1)$. Mathematically,
\begin{equation}
    -\gamma = -\gamma(1-s) + (l_2-l_1)(s)
    \label{Eq:hyperparameter_s}
\end{equation}
where $s$ is the hyperparameter. Note that for a special case of $s=1$, eq (~\ref{Eq:hyperparameter_s}) becomes $-\gamma=l_2-l_1$.

To analyze the impact of $s$ on HaS-Nets, we choose $\epsilon=0.4$ and $1.0$. Fig.~\ref{fig:hyperparameters_fmnist}(b) shows the ASR of $\epsilon$-Attack for different values of $s$ on Fashion-MNIST with Policy 2 enforced. We do not find any significant impact of changing $s$ on ASR (note that $s=0.1$ and $s=1.0$ exhibit similar performance) and attribute it to the adaptiveness of our policy. We use $s=0.3$ for further experiments due to its average performance, unless otherwise stated.

\noindent \textbf{4. \quad The Size of Healing Set - $M$}\newline
Intuitively, the \textit{size of the healing set}, $M$, can significantly affect the accuracy of a DNN when tested on the clean test data while estimating $\gamma$. The Fashion-MNIST dataset is known to be easier to learn and thus may not capture the broader impact of $M$ on accuracy and ASR. We therefore analyze this parameter for both Fashion-MNIST and CIFAR-10.

Fig.~\ref{fig:hyperparameters_fmnist}(d) and Fig.~\ref{fig:hyperparameters_cifar10} record the impact of the size of healing data on the performance of Has-Nets for Fashion-MNIST and CIFAR-10, respectively. As expected, an increase in the size of the healing data increases the network performance on test data and reduces the ASR. This is because too small of a healing set, may not sufficiently represent a comprehensive real distribution. The corresponding $\gamma$ values would result in an unfair selection/rejection during training.

For further experiments we choose $M$ to be $\leq$15\% of the full training data to ensure the effectiveness of HaS-Nets for practical scenarios.

\section{Evaluations}\label{sec:evaluation}
\begin{figure}[!t]
	\centering
	\includegraphics[width=1\linewidth]{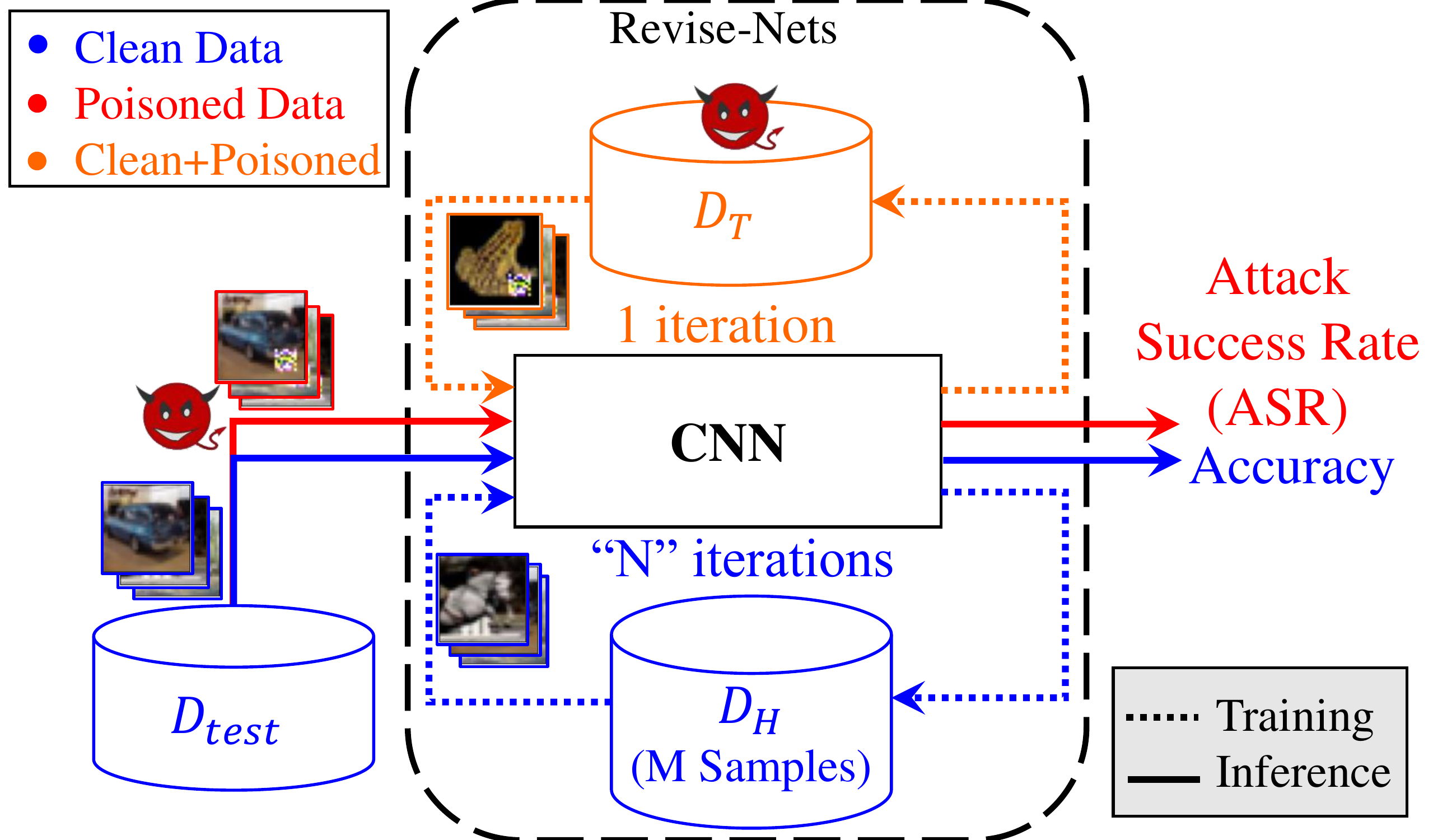}
	\vskip -0.15in
	\caption{\textit{Experimental Setup for evaluating HaS-Nets. Figure also illustrates our Threat Model, where an adversary can poison training data, $D_T$, and inference-time inputs.}}
	\label{fig:revisenets_setup}
\end{figure}\setlength{\textfloatsep}{5pt}
\renewcommand{\tabcolsep}{3pt}
\begin{table}[tb]
\vskip -0.2in
\centering
\caption{Typical settings for evaluating \textit{HaS-Nets}} % title name of the table
\resizebox{\columnwidth}{!}{
\begin{tabular}{|l|c|c|c|c|c|c|c|c|}
\hline\hline
\multirow{2}{1em}{Dataset} & \multirow{2}{2em}{Type} & \multirow{2}{3em}{Classes} & Training & Test & Poisoned & \multicolumn{3}{c|}{Hyperparameters} \\
\cline{7-9}
 & & & Data & Data & Samples & M & s & d \\
\hline
Fashion-MNIST & Image & 10 & 60000 & 10000 & 600 & 13\% & 0.3 & 0.4 \\
CIFAR-10 & Image & 10 & 50000 & 10000 & 600 & 16\% & 0.3 & 0.4 \\
IMDB & Text & 2 & 25000 & 25000 & 600 & 8\% & 0.3 & 0.4 \\
Consumer Complaint & Text & 11 & 50104 & 16702 & 600 & 9\% & 0.3 & 0.4 \\
Urban Sound & Audio & 10 & 6549 & 2183 & 80 & 8\% & 0.3 & 0.4 \\
\hline
\end{tabular}
}
\label{tab:revisenets_settings}
\end{table}
In this section, we evaluate our defense against different variants and configurations of backdoor attacks for Fashion-MNIST, CIFAR-10, mini-Consumer Complaint and Urban Sound datasets. Our experiments suggest that HaS-Nets can effectively resist backdoor insertion under several attack settings, irrespective of the dataset and the network architecture.
%Surya what about the architecture? Are't we testing against different architecture as well. 
% You also need to state what are the main aims of the evaluations.
% As before remove too many subsections 
% You need describe the dataset in a paragraph 
% I would rather say this as dataset and explain in a bit more details. 
\subsection{Experimental Setup}
The experimental setup is shown in Fig.~\ref{fig:revisenets_setup}. We evaluate HaS-Nets for different datasets, choosing each for its wide use in the literature. The typical settings used are given in Table~\ref{tab:revisenets_settings}.

\noindent \textbf{Datasets and Architectures:}
For evaluation on vision tasks, we train 2D-CNNs on Fashion-MNIST and CIFAR-10 datasets. Both datasets contain 10 classes, where Fashion-MNIST is a 28x28 grey-scale image dataset, while CIFAR-10 contains color images 32x32 in size. For text and audio applications, we use an MLP model for IMDB, a 1D-CNN for mini-Consumer Complaint\footnote{https://www.kaggle.com/cfpb/us-consumer-finance-complaints.} and a 2D-CNN for Urban Sound\footnote{https://www.kaggle.com/chrisfilo/urbansound8k.} dataset. A more detailed description of datasets and respective architectures is provided in Appendix~\ref{app:defense}.

\subsection{Evaluation on $\epsilon$-Attack}
% It seems parameter setting and attacks does not have much difference; I suggest you look the recent paper and see how you can describe such things; it is a bit confusing as you are doing the attacks while setting the parameters; 
% I am not sure what is the best way to present;
%==============
%=====done=====
%==============
\noindent \textbf{Fashion-MNIST:}
We evaluate the robustness of HaS-Nets by performing $\epsilon$-Attack on Fashion-MNIST. Results are shown in Fig.~\ref{fig:results_all}(a). We notice a slight decrease in accuracy of the network on test data as compared to an undefended model. This may be attributed to the robustness offered by HaS-Nets.

\noindent \textbf{CIFAR-10:}
The results of our evaluation on CIFAR-10 are shown in Fig.~\ref{fig:results_all}(b). One can observe a significant increase in robustness, for different values of $\epsilon$. Here we again observe the cost of improved robustness of HaS-Nets, in the form of a slight reduction in accuracy on clean test data.

\begin{figure}[!t]
    \centering
    \includegraphics[width=1\linewidth]{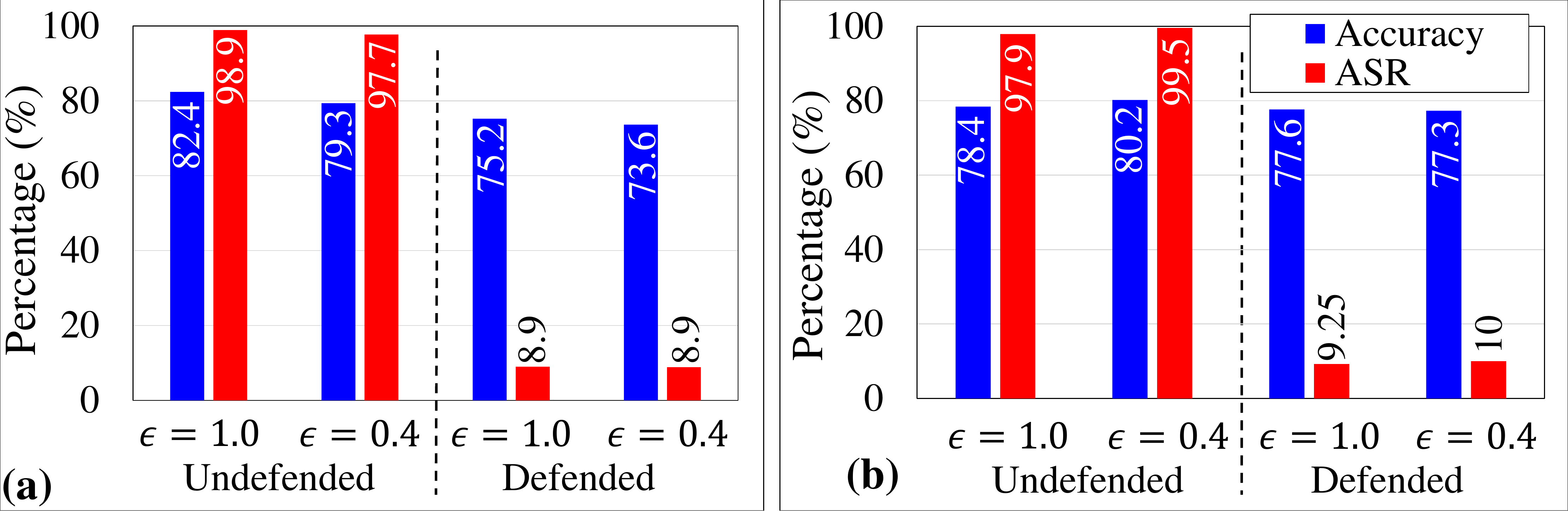}
    \vskip -0.15in
    \caption{\textit{Test accuracy of HaS-Nets and $\epsilon$-Attack ASR against HaS-Nets for (a) Consumer Complaint and (b) Urban Sound dataset.}}
    \label{fig:results_all_ex2}
\end{figure}

\noindent \textbf{Consumer Complaint:}
The results of our evaluation on mini-Consumer Complaint dataset are shown in Fig.~\ref{fig:results_all_ex2}(a). We use 1D-CNN to perform the classification task. The figure shows that HaS-Nets are able to effectively resist $\epsilon$-Attack for different $\epsilon$ values. We again observe a reduction in the accuracy of HaS-Nets on clean test data as compared to an undefended model. This shows the scalability of HaS-Nets to other datasets and architectures, even at the behavioral level.

\noindent \textbf{Urban Sound:}
The results of our evaluation on Urban Sound dataset are shown in Fig.~\ref{fig:results_all_ex2}(b). We make similar observations as in the previous experiments, i.e. a significant improvement in robustness and a slight degradation in test accuracy.

% Surya - provide some more results in the appendix 
\subsection{Evaluation on All-Trojan Attack}\label{sec:evaluation_allTrojan}
\begin{figure}[!t]
    \centering
    \includegraphics[width=1\linewidth]{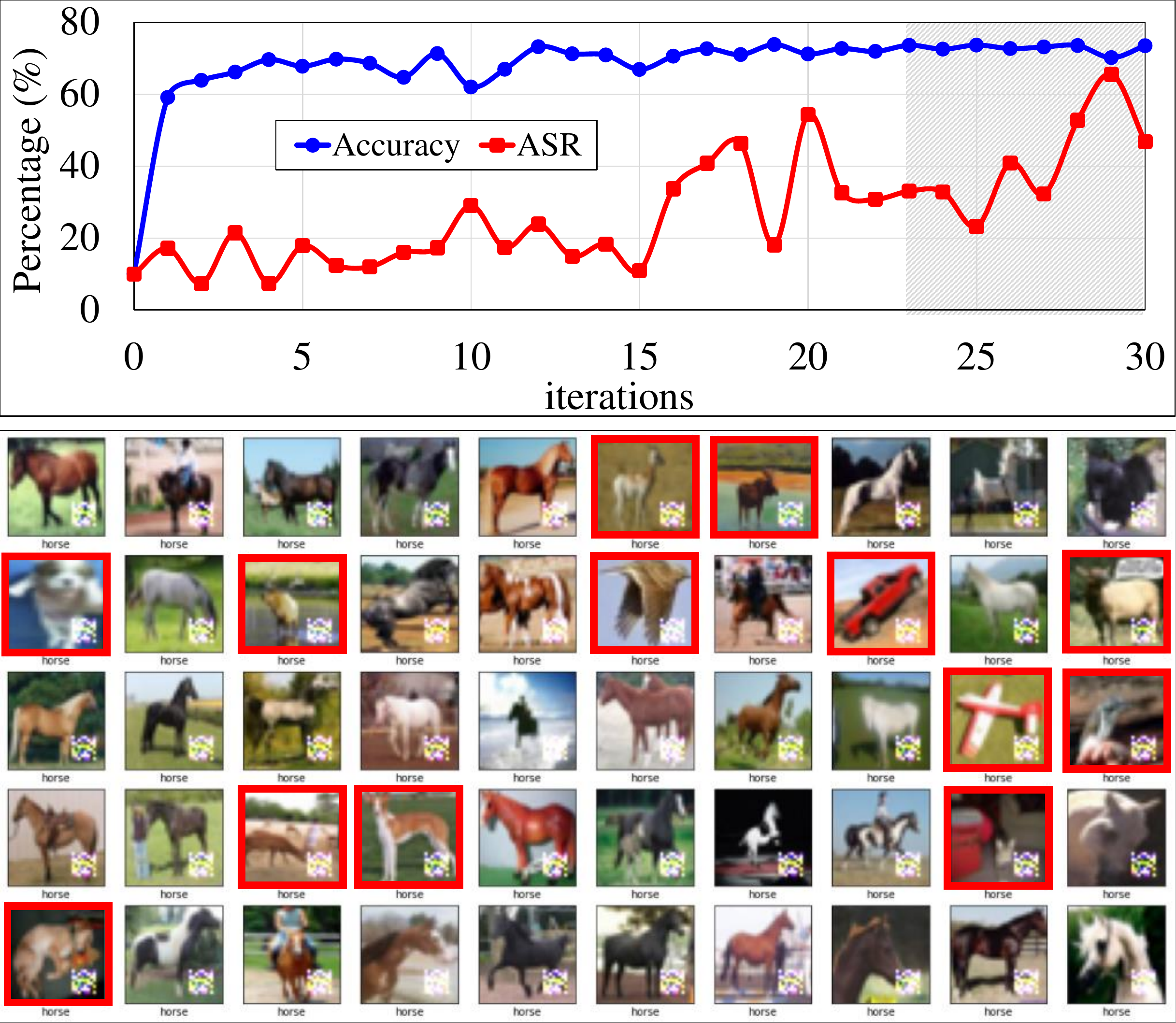}
    \vskip -0.15in
    \caption{\textit{All-trojan attack on HaS-Nets (target class being ``horse'') for CIFAR-10. \textbf{Top:} Test accuracy and ASR against HaS-Nets for $\epsilon=1.0$. \textbf{Bottom:} First 50 training samples to have passed the selection criteria of HaS-Nets after 30 iterations. Images which are not horses are marked with a red-box around them.}}
    \label{fig:evaluation_allTrojanCifar10}
\end{figure}

\begin{figure}[!t]
    \centering
    \includegraphics[width=1\linewidth]{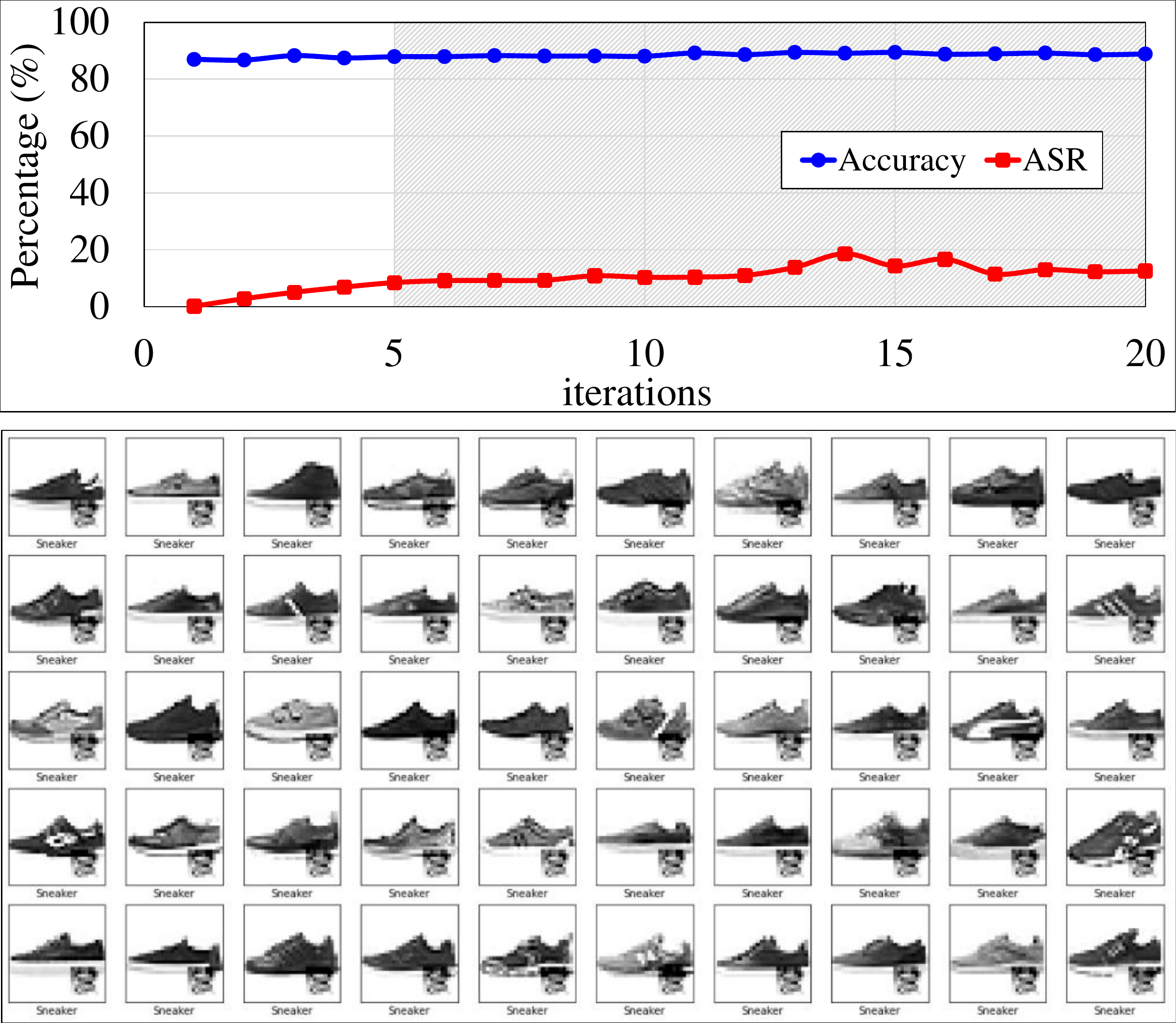}
    \vskip -0.15in
    \caption{\textit{All-trojan attack on HaS-Nets (target class being ``sneakers'') for Fashion-MNIST. \textbf{Top:} Test accuracy and ASR against HaS-Nets for $\epsilon=1.0$. \textbf{Bottom:} First 50 training samples to have passed the selection criteria of HaS-Nets after 20 iterations.}}
    \label{fig:evaluation_allTrojanFMNIST}
\end{figure}

One limitation of our defense is the non-availability of healing dataset. To uncover further vulnerabilities, we observe how HaS-Nets respond to an all-trojan attack (where all the images of the training data are poisoned images) for CIFAR-10 and Fashion-MNIST datasets. Specifically, we imprint a trigger on all the images of training data and relabel them as ``horses'' for CIFAR-10 and ``sneakers'' for Fashion-MNIST, using $\epsilon=1.0$. Then, we train HaS-Nets on the poisoned training data for a few iterations. The results are shown in Fig.~\ref{fig:evaluation_allTrojanCifar10} and Fig.~\ref{fig:evaluation_allTrojanFMNIST}.
Surprisingly, \textit{HaS-Nets can resist such a large-scale attack}. We attribute this to HaS-Net removing all the training samples which appear to be poisoning a DNN i.e. HaS-Nets only use the training samples which are consistent with the healing set.

\setlength{\textfloatsep}{5pt}
\setlength{\textfloatsep}{5pt}

After several iterations, CIFAR-10 and Fashion-MNIST training sets only contain 4876 and 3169 samples, respectively.
Fig.~\ref{fig:evaluation_allTrojanCifar10} and Fig.~\ref{fig:evaluation_allTrojanFMNIST} show the first 50 training samples for CIFAR-10 and Fashion-MNIST, respectively. Interestingly, though unsurprisingly, we observe that most of these images are horses and sneakers (i.e. belonging to the target class to which we initially labelled the poisoned samples). In other words, the network appears to have successfully removed most of the incorrectly labelled images due to their poor consistency. To verify this, we input the training sets into a model with no backdoor inserted (a clean model). Interestingly, 91.899\% of CIFAR-10 remaining samples were classified as ``horse'' by the clean model. A similar experiment for Fashion-MNIST yields 99.9\% of the remaining training samples to be ``sneakers''.

% Can you put a marker in the figures to explain the graphs (like flatten place etc.)
% It is not clear to me wht is the lesson learned from the all trojan attacks; our defense with healing dataset does not work? I think we need clarity on the messages
%==============
%=====done=====
%==============
\subsection{Evaluation on Invisible $\epsilon$-Attack}
Invisible backdoor attacks are specially effective against defenses based on the statistical filtering of inputs. Such attacks use a small magnitude noise, instead of a visible patch, as the trigger. Consequently, a human observer cannot distinguish between a clean and a poisoned input.

\begin{figure}[!t]
    \centering
    \includegraphics[width=1\linewidth]{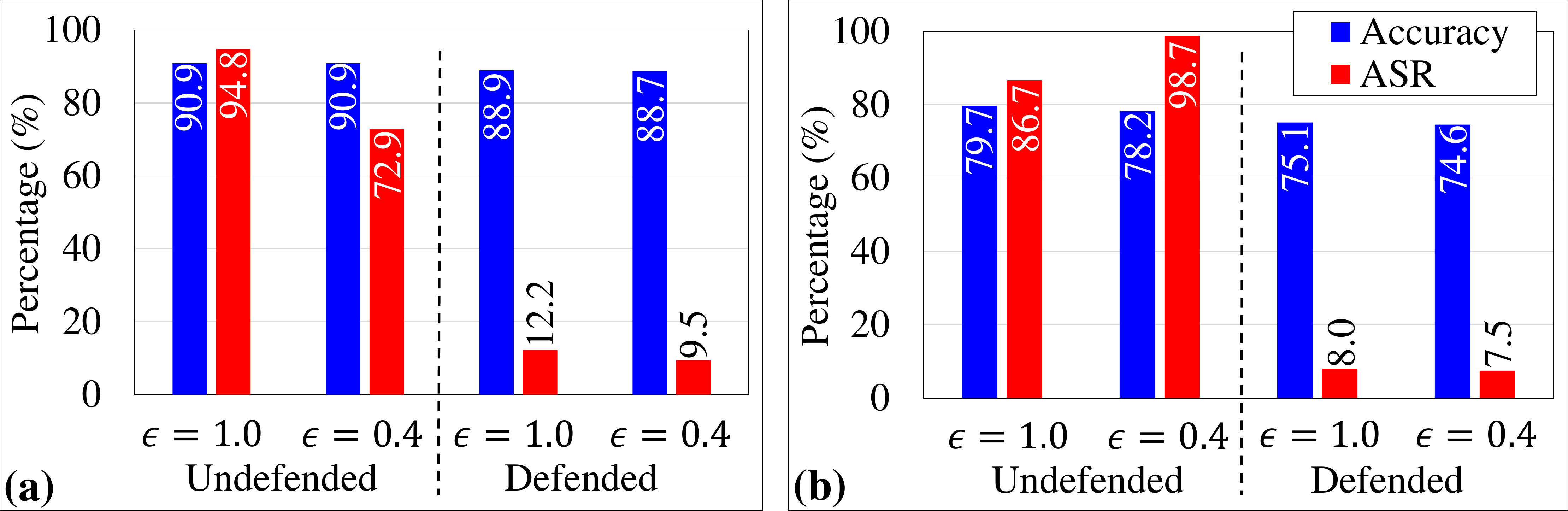}
    \vskip -0.15in
    \caption{\textit{Test accuracy of HaS-Nets and invisible $\epsilon$-Attack ASR against HaS-Nets for (a) Fashion-MNIST and (b) CIFAR-10 datasets.}}
    \label{fig:evaluation_invisible_attack}
\end{figure}

We expose HaS-Net to invisible-backdoor attack (See Appendix~\ref{app:details} for details) and study the test accuracy and ASR as compared to an undefended model in Fig.~\ref{fig:evaluation_invisible_attack}(a) and (b) for Fashion-MNIST and CIFAR-10, respectively. A reduction in the ASR from above 90\% for an undefended model to below 12\% for HaS-Nets indicates that HaS-Nets can effectively counter such backdoor attack variants.

\subsection{Evaluation on Label-Consistent Attack}
Label-consistent backdoor attacks~\cite{DBLP:label_consistent} exploit a GAN to poison an image without changing its label. Specifically, they interpolate latent representations of a few target class images to the latent representations of other classes. These poisoned images have highly inconsistent latent representations, but appear benign to human observers. The poisoned images, stamped with a trigger are harder for a DNN to learn. Consequently, the DNN preferably learns the backdoor.

\setlength{\textfloatsep}{5pt}
\begin{figure}[!t]
    \centering
    \includegraphics[width=1\linewidth]{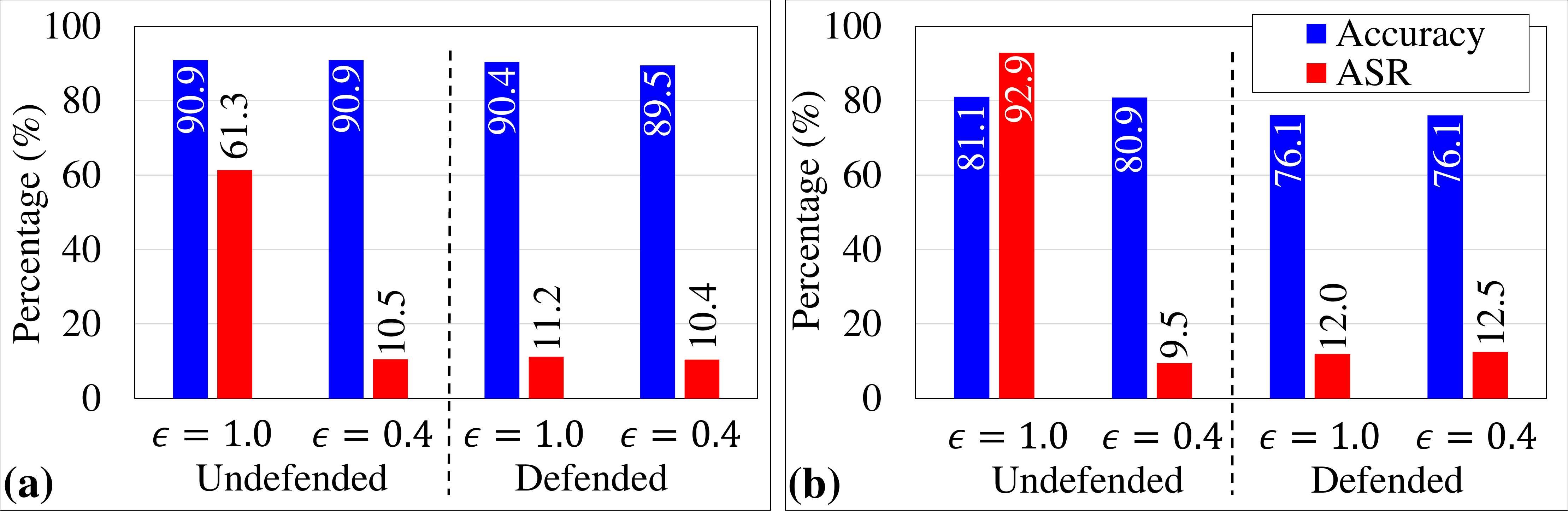}
    \vskip -0.15in
    \caption{\textit{Test accuracy of HaS-Nets and label-Consistent $\epsilon$-Attack ASR against HaS-Nets for (a) Fashion-MNIST and (b) CIFAR-10 datasets.}}
    \label{fig:evaluation_labelConsistent_attack}
\end{figure}\setlength{\textfloatsep}{5pt}
We evaluate HaS-Nets against Label-Consistent Attacks (See Appendix~\ref{app:details} for implementation details) for Fashion-MNIST and CIFAR-10, by poisoning 25\% of target class samples in the training set. Results are shown in Fig.~\ref{fig:evaluation_labelConsistent_attack}(a) and (b) for Fashion-MNIST and CIFAR-10, respectively. In case of $\epsilon=1.0$, HaS-Nets can successfully resist the backdoor insertion. Specifically, we observe a reduction in ASR from 61\% and 93\% to 11\% and 12\%, for Fashion-MNIST and CIFAR-10, respectively.
For $\epsilon=0.4$, the attacker is unable to insert a backdoor in the network for both datasets. This shows that GAN-based label consistent backdoor attacks are relatively weaker than conventional targeted backdoor attacks - a cost of the increased inconspicuousness of the attack~\cite{DBLP:label_consistent}.
\section{Conclusion}\label{sec:conclusion}
In this work, we challenged the robustness of recently proposed defenses by proposing a low-confidence backdoor attack, and its two variants, $\epsilon$-Attack and $\epsilon^2$-Attack. Low-confidence backdoor attacks utilize low confidence labels to hide their presence from the defender. By carefully analyzing the behaviour of poisoned samples, we developed useful insights for a generic defense and propose a novel strategy, ``\textit{HaS-Nets}'', for defending DNNs against backdoor attacks by assuming the presence of a small healing dataset available to the defender. To prove the scalability of our approach, we analyzed \textit{HaS-Nets} for different hyper-parameter values on Fashion-MNIST and extended the best settings to evaluate the robustness on other datasets i.e. CIFAR-10, IMDB, Consumer Complaint and Urban Sound datasets.

HaS-Nets were shown to resist many variants of backdoor attacks (e.g. $\epsilon$-attack, invisible backdoor attack, all-Trojan attack, Label-consistent attack) under diverse settings and outperform state-of-the-art defenses (i.e. Gradient-shaping, ULP-defense, Februus and STRIP). Moreover, \textit{HaS-Nets} were shown to be agnostic to dataset type and network architecture.

Our work is the first to evaluate a defense on an all-trojan backdoor attack by poisoning 100\% of the training data. We demonstrated the effectiveness of \textit{HaS-Nets} under such extreme settings.
%-------------------------------------------------------------------------------
\bibliographystyle{plain}
\bibliography{main.bib}

\begin{thebibliography}{10}

\bibitem{DBLP:differential_privacy}
Mart{\'{\i}}n Abadi, Andy Chu, Ian~J. Goodfellow, H.~Brendan McMahan, Ilya
  Mironov, Kunal Talwar, and Li~Zhang.
\newblock Deep learning with differential privacy.
\newblock In Edgar~R. Weippl, Stefan Katzenbeisser, Christopher Kruegel,
  Andrew~C. Myers, and Shai Halevi, editors, {\em Proceedings of the 2016 {ACM}
  {SIGSAC} Conference on Computer and Communications Security, Vienna, Austria,
  October 24-28, 2016}, pages 308--318. {ACM}, 2016.

\bibitem{DBLP:security_of_machine_learning}
Marco Barreno, Blaine Nelson, Anthony~D. Joseph, and J.~D. Tygar.
\newblock The security of machine learning.
\newblock {\em Mach. Learn.}, 81(2):121--148, 2010.

\bibitem{DBLP:activation_clustering}
Bryant Chen, Wilka Carvalho, Nathalie Baracaldo, Heiko Ludwig, Benjamin
  Edwards, Taesung Lee, Ian Molloy, and Biplav Srivastava.
\newblock Detecting backdoor attacks on deep neural networks by activation
  clustering.
\newblock In Hu{\'{a}}scar Espinoza, Se{\'{a}}n~{\'{O}} h{\'{E}}igeartaigh,
  Xiaowei Huang, Jos{\'{e}} Hern{\'{a}}ndez{-}Orallo, and Mauricio
  Castillo{-}Effen, editors, {\em Workshop on Artificial Intelligence Safety
  2019 co-located with the Thirty-Third {AAAI} Conference on Artificial
  Intelligence 2019 (AAAI-19), Honolulu, Hawaii, January 27, 2019}, volume 2301
  of {\em {CEUR} Workshop Proceedings}. CEUR-WS.org, 2019.

\bibitem{DBLP:chen_targeted_invisible}
Xinyun Chen, Chang Liu, Bo~Li, Kimberly Lu, and Dawn Song.
\newblock Targeted backdoor attacks on deep learning systems using data
  poisoning.
\newblock {\em CoRR}, abs/1712.05526, 2017.

\bibitem{DBLP:sentinet}
Edward Chou, Florian Tram{\`{e}}r, Giancarlo Pellegrino, and Dan Boneh.
\newblock Sentinet: Detecting physical attacks against deep learning systems.
\newblock {\em CoRR}, abs/1812.00292, 2018.

\bibitem{DBLP:sever}
Ilias Diakonikolas, Gautam Kamath, Daniel Kane, Jerry Li, Jacob Steinhardt, and
  Alistair Stewart.
\newblock Sever: {A} robust meta-algorithm for stochastic optimization.
\newblock In Kamalika Chaudhuri and Ruslan Salakhutdinov, editors, {\em
  Proceedings of the 36th International Conference on Machine Learning, {ICML}
  2019, 9-15 June 2019, Long Beach, California, {USA}}, volume~97 of {\em
  Proceedings of Machine Learning Research}, pages 1596--1606. {PMLR}, 2019.

\bibitem{Februus}
Bao~Gia Doan, Ehsan Abbasnejad, and Damith~C. Ranasinghe.
\newblock Februus: Input purification defense against trojan attacks on deep
  neural network systems.
\newblock In {\em Proceedings of the 36th Annual Computer Security Applications
  Conference (ACSAC)}, ACSAC 2020, 2020.

\bibitem{DBLP:human_DL}
Samuel~F. Dodge and Lina~J. Karam.
\newblock A study and comparison of human and deep learning recognition
  performance under visual distortions.
\newblock In {\em 26th International Conference on Computer Communication and
  Networks, {ICCCN} 2017, Vancouver, BC, Canada, July 31 - Aug. 3, 2017}, pages
  1--7. {IEEE}, 2017.

\bibitem{DBLP:robust_anomaly_diff_privacy}
Min Du, Ruoxi Jia, and Dawn Song.
\newblock Robust anomaly detection and backdoor attack detection via
  differential privacy.
\newblock In {\em 8th International Conference on Learning Representations,
  {ICLR} 2020, Addis Ababa, Ethiopia, April 26-30, 2020}. OpenReview.net, 2020.

\bibitem{DBLP:backdoor_attacks_and_countermeasures}
Yansong Gao, Bao~Gia Doan, Zhi Zhang, Siqi Ma, Jiliang Zhang, Anmin Fu, Surya
  Nepal, and Hyoungshick Kim.
\newblock Backdoor attacks and countermeasures on deep learning: {A}
  comprehensive review.
\newblock {\em CoRR}, abs/2007.10760, 2020.

\bibitem{DBLP:strip2}
Yansong Gao, Yeonjae Kim, Bao~Gia Doan, Zhi Zhang, Gongxuan Zhang, Surya Nepal,
  Damith~Chinthana Ranasinghe, and Hyoungshick Kim.
\newblock Design and evaluation of a multi-domain trojan detection method on
  deep neural networks.
\newblock {\em CoRR}, abs/1911.10312, 2019.

\bibitem{DBLP:strip}
Yansong Gao, Change Xu, Derui Wang, Shiping Chen, Damith~Chinthana Ranasinghe,
  and Surya Nepal.
\newblock {STRIP:} a defence against trojan attacks on deep neural networks.
\newblock In David Balenson, editor, {\em Proceedings of the 35th Annual
  Computer Security Applications Conference, {ACSAC} 2019, San Juan, PR, USA,
  December 09-13, 2019}, pages 113--125. {ACM}, 2019.

\bibitem{DBLP:gtsrb}
{\'{A}}lvaro~Arcos Garc{\'{\i}}a, Juan~Antonio {\'{A}}lvarez, and Luis~Miguel
  Soria{-}Morillo.
\newblock Deep neural network for traffic sign recognition systems: An analysis
  of spatial transformers and stochastic optimisation methods.
\newblock {\em Neural Networks}, 99:158--165, 2018.

\bibitem{GAN}
Ian Goodfellow, Jean Pouget-Abadie, Mehdi Mirza, Bing Xu, David Warde-Farley,
  Sherjil Ozair, Aaron Courville, and Yoshua Bengio.
\newblock Generative adversarial nets.
\newblock pages 2672--2680, 2014.

\bibitem{DBLP:badnets_2019}
Tianyu Gu, Kang Liu, Brendan Dolan{-}Gavitt, and Siddharth Garg.
\newblock Badnets: Evaluating backdooring attacks on deep neural networks.
\newblock {\em {IEEE} Access}, 7:47230--47244, 2019.

\bibitem{DBLP:gradient_shaping}
Sanghyun Hong, Varun Chandrasekaran, Yigitcan Kaya, Tudor Dumitras, and Nicolas
  Papernot.
\newblock On the effectiveness of mitigating data poisoning attacks with
  gradient shaping.
\newblock {\em CoRR}, abs/2002.11497, 2020.

\bibitem{DBLP:manipulating_ml}
Matthew Jagielski, Alina Oprea, Battista Biggio, Chang Liu, Cristina
  Nita{-}Rotaru, and Bo~Li.
\newblock Manipulating machine learning: Poisoning attacks and countermeasures
  for regression learning.
\newblock In {\em 2018 {IEEE} Symposium on Security and Privacy, {SP} 2018,
  Proceedings, 21-23 May 2018, San Francisco, California, {USA}}, pages 19--35.
  {IEEE} Computer Society, 2018.

\bibitem{DBLP:red_attack}
Faiq Khalid, Hassan Ali, Muhammad~Abdullah Hanif, Semeen Rehman, Rehan Ahmed,
  and Muhammad Shafique.
\newblock Fadec: {A} fast decision-based attack for adversarial machine
  learning.
\newblock In {\em 2020 International Joint Conference on Neural Networks,
  {IJCNN} 2020, Glasgow, United Kingdom, July 19-24, 2020}, pages 1--8. {IEEE},
  2020.

\bibitem{DBLP:ulps}
Soheil Kolouri, Aniruddha Saha, Hamed Pirsiavash, and Heiko Hoffmann.
\newblock Universal litmus patterns: Revealing backdoor attacks in cnns.
\newblock In {\em 2020 {IEEE/CVF} Conference on Computer Vision and Pattern
  Recognition, {CVPR} 2020, Seattle, WA, USA, June 13-19, 2020}, pages
  298--307. {IEEE}, 2020.

\bibitem{DBLP:deep_learning_backdoors}
Shaofeng Li, Shiqing Ma, Minhui Xue, and Benjamin Zi~Hao Zhao.
\newblock Deep learning backdoors.
\newblock {\em CoRR}, abs/2007.08273, 2020.

\bibitem{DBLP:backdoor_learning_a_survey}
Yiming Li, Baoyuan Wu, Yong Jiang, Zhifeng Li, and Shu{-}Tao Xia.
\newblock Backdoor learning: {A} survey.
\newblock {\em CoRR}, abs/2007.08745, 2020.

\bibitem{DBLP:rethinking_trigger}
Yiming Li, Tongqing Zhai, Baoyuan Wu, Yong Jiang, Zhifeng Li, and Shutao Xia.
\newblock Rethinking the trigger of backdoor attack.
\newblock {\em CoRR}, abs/2004.04692, 2020.

\bibitem{DBLP:fine_pruning}
Kang Liu, Brendan Dolan{-}Gavitt, and Siddharth Garg.
\newblock Fine-pruning: Defending against backdooring attacks on deep neural
  networks.
\newblock In Michael Bailey, Thorsten Holz, Manolis Stamatogiannakis, and
  Sotiris Ioannidis, editors, {\em Research in Attacks, Intrusions, and
  Defenses - 21st International Symposium, {RAID} 2018, Heraklion, Crete,
  Greece, September 10-12, 2018, Proceedings}, volume 11050 of {\em Lecture
  Notes in Computer Science}, pages 273--294. Springer, 2018.

\bibitem{DBLP:abs}
Yingqi Liu, Wen{-}Chuan Lee, Guanhong Tao, Shiqing Ma, Yousra Aafer, and
  Xiangyu Zhang.
\newblock {ABS:} scanning neural networks for back-doors by artificial brain
  stimulation.
\newblock In Lorenzo Cavallaro, Johannes Kinder, XiaoFeng Wang, and Jonathan
  Katz, editors, {\em Proceedings of the 2019 {ACM} {SIGSAC} Conference on
  Computer and Communications Security, {CCS} 2019, London, UK, November 11-15,
  2019}, pages 1265--1282. {ACM}, 2019.

\bibitem{DBLP:neural_trojans}
Yuntao Liu, Yang Xie, and Ankur Srivastava.
\newblock Neural trojans.
\newblock In {\em 2017 {IEEE} International Conference on Computer Design,
  {ICCD} 2017, Boston, MA, USA, November 5-8, 2017}, pages 45--48. {IEEE}
  Computer Society, 2017.

\bibitem{DBLP:biometrics_survey}
Shervin Minaee, AmirAli Abdolrashidi, Hang Su, Mohammed Bennamoun, and David
  Zhang.
\newblock Biometric recognition using deep learning: {A} survey.
\newblock {\em CoRR}, abs/1912.00271, 2019.

\bibitem{DBLP:unv_adv_pers}
Seyed{-}Mohsen Moosavi{-}Dezfooli, Alhussein Fawzi, Omar Fawzi, and Pascal
  Frossard.
\newblock Universal adversarial perturbations.
\newblock In {\em 2017 {IEEE} Conference on Computer Vision and Pattern
  Recognition, {CVPR} 2017, Honolulu, HI, USA, July 21-26, 2017}, pages 86--94.
  {IEEE} Computer Society, 2017.

\bibitem{DBLP:breastcancer}
Wajahat Nawaz, Sagheer Ahmed, Ali Tahir, and Hassan~Aqeel Khan.
\newblock Classification of breast cancer histology images using {ALEXNET}.
\newblock In Aur{\'{e}}lio Campilho, Fakhri Karray, and Bart~M. ter
  Haar~Romeny, editors, {\em Image Analysis and Recognition - 15th
  International Conference, {ICIAR} 2018, P{\'{o}}voa de Varzim, Portugal, June
  27-29, 2018, Proceedings}, volume 10882 of {\em Lecture Notes in Computer
  Science}, pages 869--876. Springer, 2018.

\bibitem{DBLP:gradcam}
Ramprasaath~R. Selvaraju, Michael Cogswell, Abhishek Das, Ramakrishna Vedantam,
  Devi Parikh, and Dhruv Batra.
\newblock Grad-cam: Visual explanations from deep networks via gradient-based
  localization.
\newblock {\em Int. J. Comput. Vis.}, 128(2):336--359, 2020.

\bibitem{DBLP:poison_frogs}
Ali Shafahi, W.~Ronny Huang, Mahyar Najibi, Octavian Suciu, Christoph Studer,
  Tudor Dumitras, and Tom Goldstein.
\newblock Poison frogs! targeted clean-label poisoning attacks on neural
  networks.
\newblock In Samy Bengio, Hanna~M. Wallach, Hugo Larochelle, Kristen Grauman,
  Nicol{\`{o}} Cesa{-}Bianchi, and Roman Garnett, editors, {\em Advances in
  Neural Information Processing Systems 31: Annual Conference on Neural
  Information Processing Systems 2018, NeurIPS 2018, 3-8 December 2018,
  Montr{\'{e}}al, Canada}, pages 6106--6116, 2018.

\bibitem{DBLP:demon}
Di~Tang, Xiaofeng Wang, Haixu Tang, and Kehuan Zhang.
\newblock Demon in the variant: Statistical analysis of dnns for robust
  backdoor contamination detection.
\newblock {\em CoRR}, abs/1908.00686, 2019.

\bibitem{DBLP:spectral_signatures}
Brandon Tran, Jerry Li, and Aleksander Madry.
\newblock Spectral signatures in backdoor attacks.
\newblock In Samy Bengio, Hanna~M. Wallach, Hugo Larochelle, Kristen Grauman,
  Nicol{\`{o}} Cesa{-}Bianchi, and Roman Garnett, editors, {\em Advances in
  Neural Information Processing Systems 31: Annual Conference on Neural
  Information Processing Systems 2018, NeurIPS 2018, 3-8 December 2018,
  Montr{\'{e}}al, Canada}, pages 8011--8021, 2018.

\bibitem{DBLP:systematic_evaluation}
Loc Truong, Chace Jones, Brian Hutchinson, Andrew August, Brenda Praggastis,
  Robert Jasper, Nicole Nichols, and Aaron Tuor.
\newblock Systematic evaluation of backdoor data poisoning attacks on image
  classifiers.
\newblock In {\em 2020 {IEEE/CVF} Conference on Computer Vision and Pattern
  Recognition, {CVPR} Workshops 2020, Seattle, WA, USA, June 14-19, 2020},
  pages 3422--3431. {IEEE}, 2020.

\bibitem{DBLP:label_consistent}
Alexander Turner, Dimitris Tsipras, and Aleksander Madry.
\newblock Label-consistent backdoor attacks.
\newblock {\em CoRR}, abs/1912.02771, 2019.

\bibitem{DBLP:confoc}
Miguel Villarreal{-}Vasquez and Bharat~K. Bhargava.
\newblock Confoc: Content-focus protection against trojan attacks on neural
  networks.
\newblock {\em CoRR}, abs/2007.00711, 2020.

\bibitem{DBLP:neural_cleanse}
Bolun Wang, Yuanshun Yao, Shawn Shan, Huiying Li, Bimal Viswanath, Haitao
  Zheng, and Ben~Y. Zhao.
\newblock Neural cleanse: Identifying and mitigating backdoor attacks in neural
  networks.
\newblock In {\em 2019 {IEEE} Symposium on Security and Privacy, {SP} 2019, San
  Francisco, CA, USA, May 19-23, 2019}, pages 707--723. {IEEE}, 2019.

\bibitem{DBLP:malware_detection}
Qinglong Wang, Wenbo Guo, Kaixuan Zhang, Alexander G.~Ororbia II, Xinyu Xing,
  Xue Liu, and C.~Lee Giles.
\newblock Adversary resistant deep neural networks with an application to
  malware detection.
\newblock In {\em Proceedings of the 23rd {ACM} {SIGKDD} International
  Conference on Knowledge Discovery and Data Mining, Halifax, NS, Canada,
  August 13 - 17, 2017}, pages 1145--1153. {ACM}, 2017.

\bibitem{DBLP:latent_backdoors}
Yuanshun Yao, Huiying Li, Haitao Zheng, and Ben~Y. Zhao.
\newblock Latent backdoor attacks on deep neural networks.
\newblock In Lorenzo Cavallaro, Johannes Kinder, XiaoFeng Wang, and Jonathan
  Katz, editors, {\em Proceedings of the 2019 {ACM} {SIGSAC} Conference on
  Computer and Communications Security, {CCS} 2019, London, UK, November 11-15,
  2019}, pages 2041--2055. {ACM}, 2019.

\bibitem{DBLP:UAP_invisible_zhong}
Haoti Zhong, Cong Liao, Anna~Cinzia Squicciarini, Sencun Zhu, and David~J.
  Miller.
\newblock Backdoor embedding in convolutional neural network models via
  invisible perturbation.
\newblock In Vassil Roussev, Bhavani~M. Thuraisingham, Barbara Carminati, and
  Murat Kantarcioglu, editors, {\em {CODASPY} '20: Tenth {ACM} Conference on
  Data and Application Security and Privacy, New Orleans, LA, USA, March 16-18,
  2020}, pages 97--108. {ACM}, 2020.

\end{thebibliography}
% \bibliography{main.bbl}
\clearpage
\begin{appendices}

\section{Attacking State-of-the-Art defenses}\label{app:attack}
\begin{figure}[h]
    \centering
    \includegraphics[width=1\linewidth]{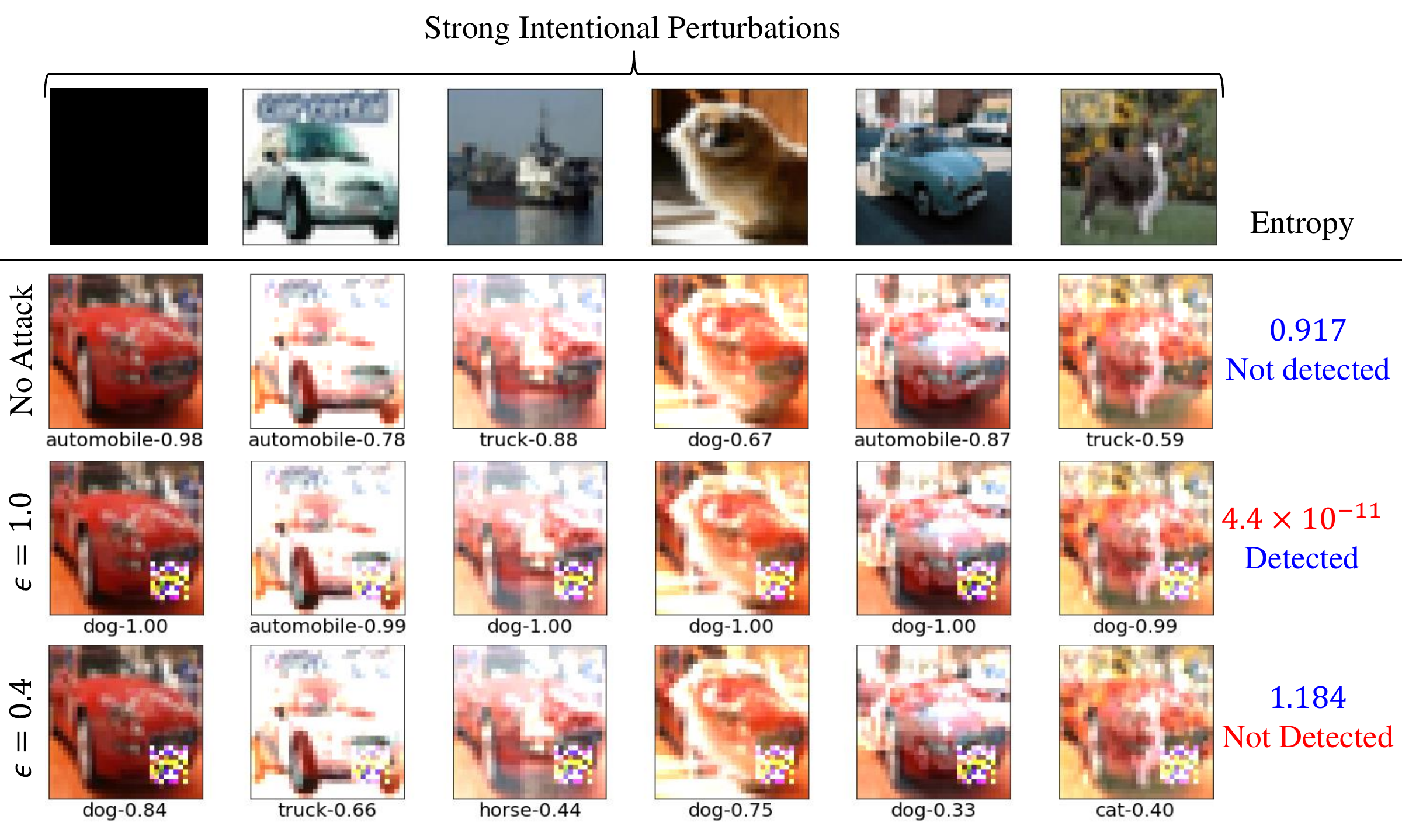}
    \vskip -0.1in
    \caption{\textit{An illustration of STRIP defense working and $\epsilon=0.4$ attack evading STRIP. STRIP uses an auto-computed threshold of $t$=0.699 (in this case), to differentiate clean and poisoned inputs.}}
    \label{figA:strip}
\end{figure}

\paragraph{Data Inspection: STRIP-ViTA~\cite{DBLP:strip}} Fig.~\ref{figA:strip} illustrates $\epsilon$-Attack (target class, ``dog'') on STRIP for a typical example of CIFAR-10 and 5 intentional perturbations per sample (Note that for evaluation in Section~\ref{sec:evaluationAttacks_strip}, we use 2000 perturbations per sample in coherence with the authors~\cite{DBLP:strip}). The perturbed inputs are presented to a DNN for classification and the entropy of decision is recorded in the last column of Fig.~\ref{figA:strip}. For $\epsilon=1.0$-Attack, a DNN classifies most of the perturbed poisoned inputs as ``dog''. This static behavior results in a small output entropy, making it suspicious to STRIP based on an auto-computed threshold ($t=0.699$ in our case). On contrary, $\epsilon=0.4$-Attack causes a large output entropy for perturbed inputs, and hence, successfully evades the detection. A no-Attack case is also presented for comparison.

\begin{figure}[h]
    \centering
    \includegraphics[width=1\linewidth]{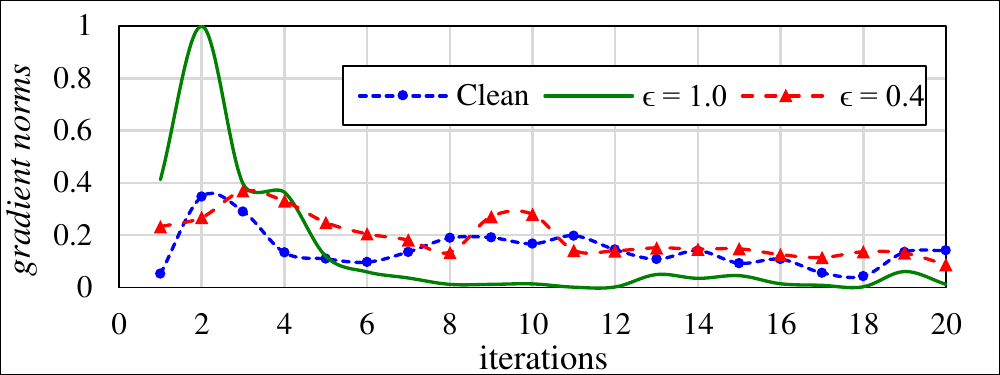}
    \vskip -0.1in
    \caption{\textit{Normalized average gradient norms of poisoned training samples with $\epsilon=0.4$ and $\epsilon=1.0$ labels. Gradient norm for clean samples is given for comparison.}}
    \label{figA:gsLoss}
\end{figure}

\paragraph{Poison Suppression: Gradient-Shaping~\cite{DBLP:gradient_shaping}}
Fig.~\ref{figA:gsLoss} gives the gradient norms of poisoned training samples for different iterations as compared to those of clean samples, with gradient-shaping applied. We use a clipping norm of 4.0 and a noise multiplier of 0.01. We observe that the gradient norms for $\epsilon=0.4$ more closely follow the gradient norms for clean samples. A peak in the gradients for $\epsilon=1.0$-Attack results in stronger updates, causing backdoors to be learned more quickly. This further validates our hypothesis regarding the early learning of backdoors in Section~\ref{alg:reviseNets}.

\begin{figure}[h]
    \centering
    \includegraphics[width=1\linewidth]{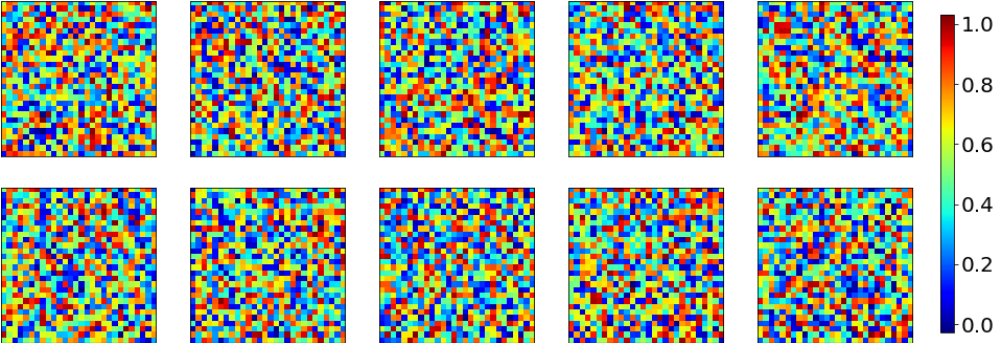}
    \vskip -0.1in
    \caption{\textit{Optimized litmus Patterns for ULP-10 defense.}}
    \label{figA:ulp10}
\end{figure}

\paragraph{Model Inspection: Universal Litmus Patterns~\cite{DBLP:ulps}}
Fig.~\ref{figA:ulp10} shows optimized so-called litmus patterns for the case of $M=10$. These litmus-patterns are given as inputs to a DNN under inspection and the output of the DNN is observed by a detector (trained together with the litmus patterns) for suspicious behavior.

% \begin{figure}[h]
%     \centering
%     \includegraphics[width=0.3\linewidth]{Figs_Appendix/ulp_patterns1.pdf}
%     \vskip -0.1in
%     \caption{\textit{Caption to be added. M=1}}
%     \label{figA:ulp1}
% \end{figure}
% \begin{figure}[h]
%     \centering
%     \includegraphics[width=1\linewidth]{Figs_Appendix/ulp_patterns.pdf}
%     \vskip -0.1in
%     \caption{\textit{Caption to be added. M=10 and M=1}}
%     \label{figA:ulp10_1}
% \end{figure}

\begin{figure*}[!t]
    \centering
    \includegraphics[width=1\linewidth]{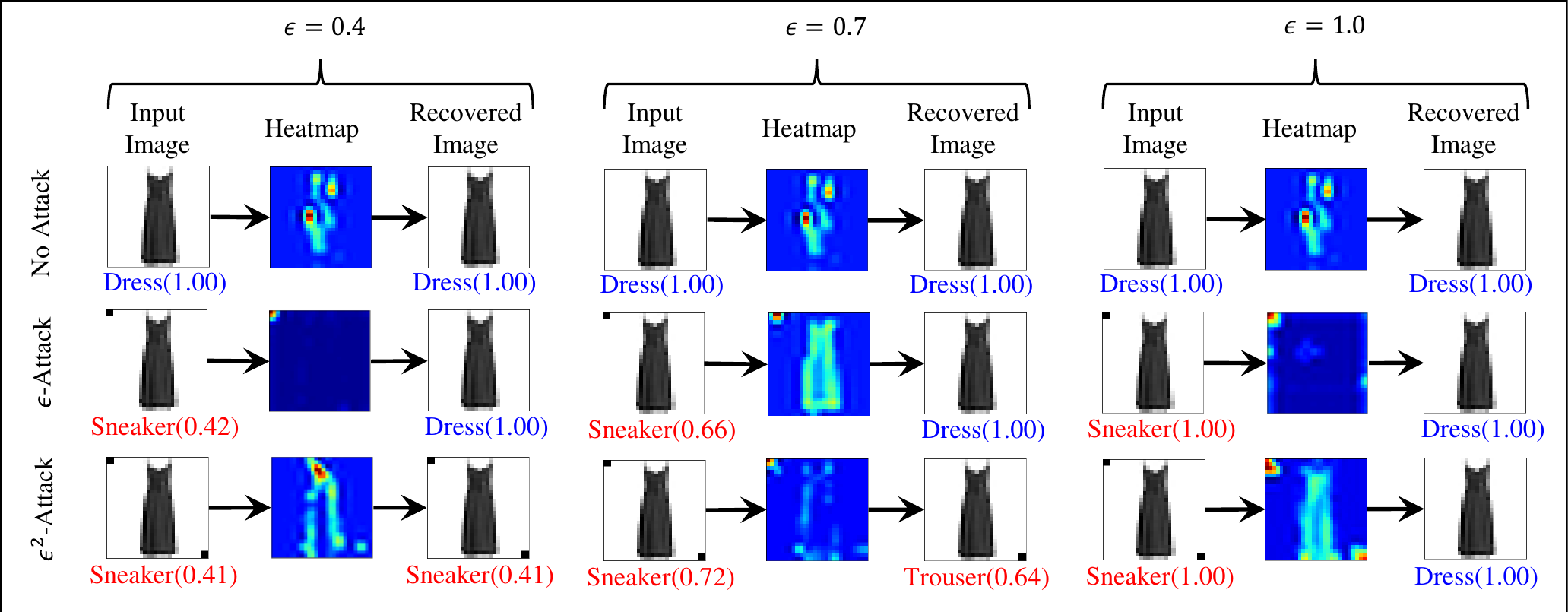}
    \vskip -0.1in
    \caption{\textit{A typical example, illustrating $\epsilon^2$-Attack evading Februus\cite{Februus}. Februus searches for potential triggers in an input image using heatmaps, and mask those triggers. The masked image is fed to a GAN, which recovers the input image trigger-free. The input image, heatmaps and the recovered images are given for different scenarios and settings.}}
    \label{fig:attack_on_defense_februus_b}
    \vskip -0.1in
\end{figure*}

\paragraph{Blind defense: Februus~\cite{Februus}}
Fig.~\ref{fig:attack_on_defense_februus_b} illustrates a typical example of our proposed attacks on Fashion-MNIST for different $\epsilon$ values. For the simple $\epsilon$-Attack, a trigger influencing the decision is identified and removed by Februus. Februus does the same for $\epsilon^2$-Attacks - removes $Z_1 \cap Z_2$ - unaware that doing this will cause $Z_2$ to get activated as shown in the figure.

\section{Evaluating HaS-Nets}\label{app:defense}

\noindent \textit{\textbf{1) Fashion-MNIST: }}Fashion-MNIST consists of 70000 28x28 grey-scale images belonging to 10 different classes. The training set contains 60000 images while the test set has 10000 images. When conducting attacks, we poison the first 600 images of the training set.

\begin{figure}[h]
    \centering
    \includegraphics[width=1\linewidth]{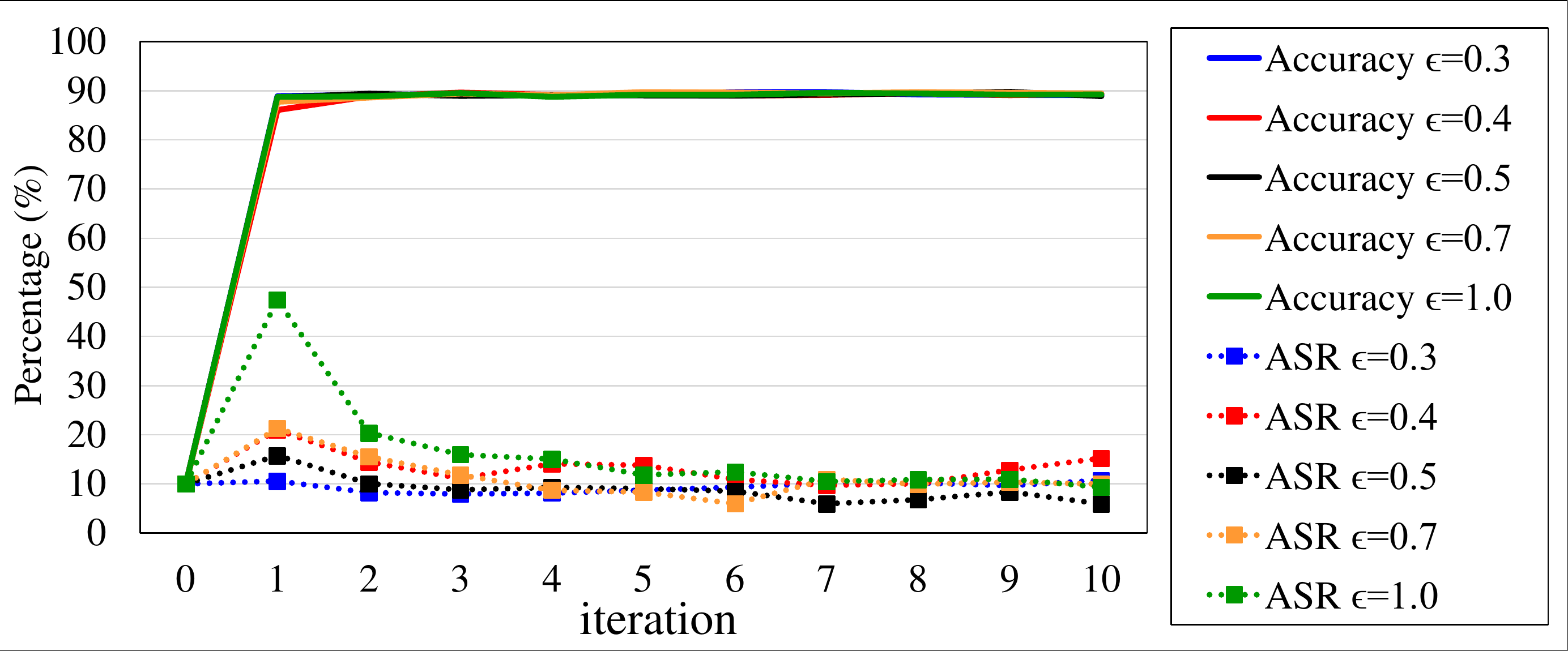}
    \vskip -0.1in
    \caption{\textit{Accuracy of HaS-Nets and Attack Success Rates(ASRs) of $\epsilon$-Attack against HaS-Nets on Fashion-MNIST for different iterations.}}
    \label{figA:results_fmnist}
\end{figure}

The architecture used for evaluating HaS-Nets on Fashion-MNIST is given in Table~\ref{tab:architectures}. Results are reported in Fig.\ref{figA:results_fmnist} for different iterations. We observe that HaS-Nets are effective irrespective of the number of iterations. We also note relatively larger ASRs for the first iteration. This can be attributed to HaS-Nets allowing the DNN to train on full training data for the first iteration.

\noindent \textit{\textbf{2) CIFAR-10: }}CIFAR-10 contains 60000 32x32x3 images belonging to 10 different classes, divided as 50000 training images and 10000 test images. When conducting attacks, we poison the first 600 images of the training set.

\begin{figure}[h]
    \centering
    \includegraphics[width=1\linewidth]{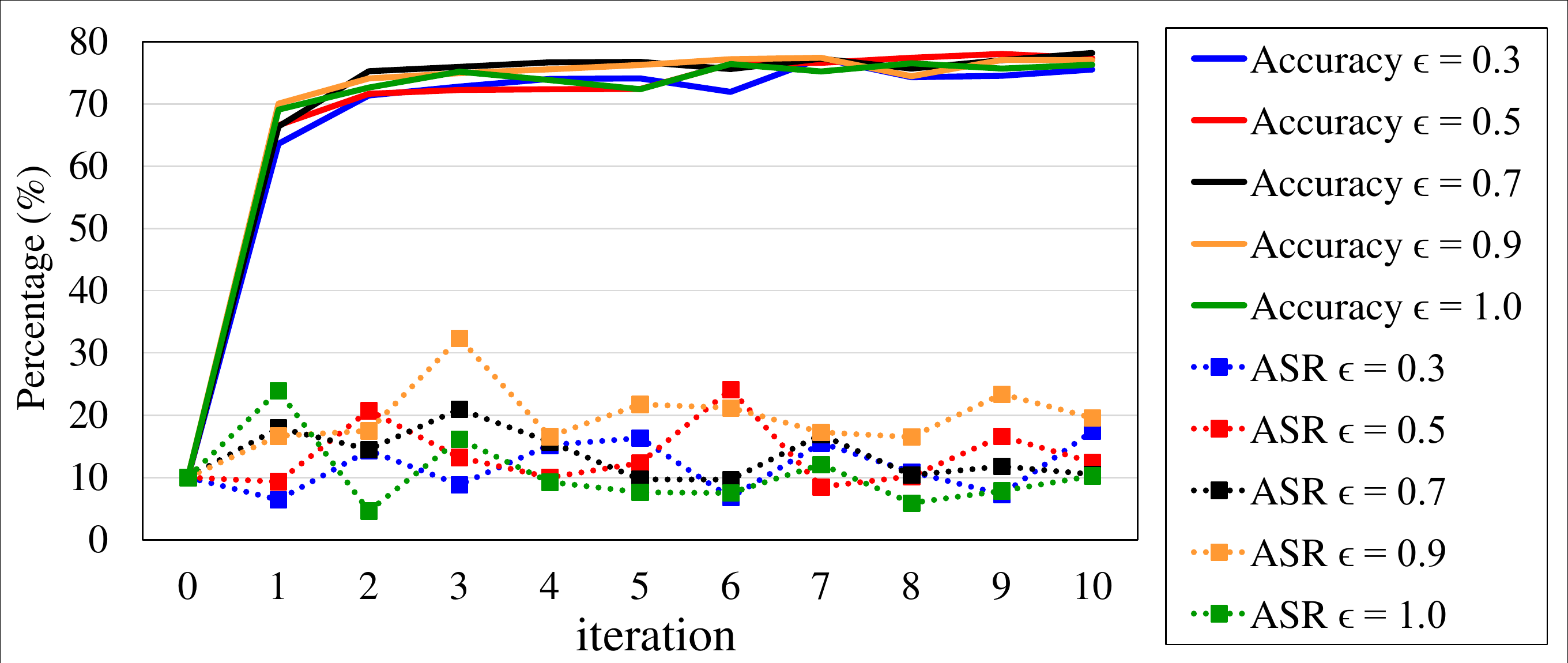}
    \vskip -0.1in
    \caption{\textit{Accuracy of HaS-Nets and ASRs of $\epsilon$-Attack against HaS-Nets on CIFAR-10 for different iterations.}}
    \label{figA:results_cifar}
\end{figure}

The architecture used for the analysis is given in Table~\ref{tab:architectures}. Results are reported in Fig.\ref{figA:results_cifar} for different iterations. Again, we observe the effectiveness of HaS-Nets for different iterations.

\noindent \textit{\textbf{3) IMDB:}} The IMDB dataset contains 50000 movie reviews categorized into two classes: positive and negative. There are 25000 samples in the training set and the same in the test set. When conducting attacks, we poison the first 600 images of the training set.

\begin{figure}[h]
    \centering
    \includegraphics[width=1\linewidth]{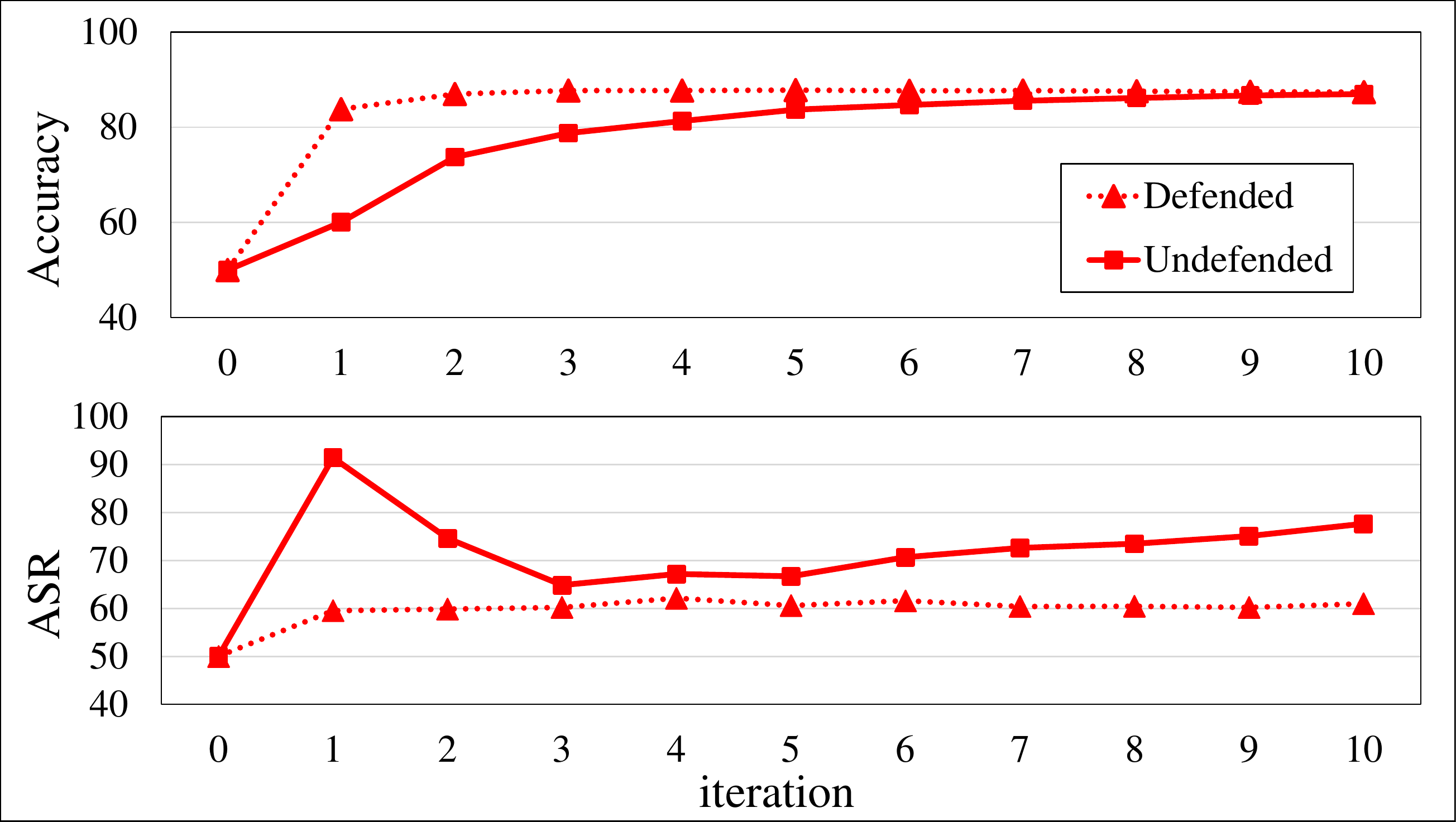}
    \vskip -0.1in
    \caption{\textit{Accuracy of HaS-Nets and ASRs of $\epsilon$-Attack against HaS-Nets on IMDB dataset for different iterations.}}
    \label{figA:results_imdb}
\end{figure}

The architecture used for this analysis is given in Table~\ref{tab:architectures} and results are shown in Fig.~\ref{figA:results_imdb}. We observe a reduction in ASR from about 80\% to 60\%. Although, this is not a very appreciable decrease, recall that the IMDB dataset only has two classes, and even if the attack were to fail, it would still have a success rate of about 50\%.

\begin{figure}[h]
    \centering
    \includegraphics[width=1\linewidth]{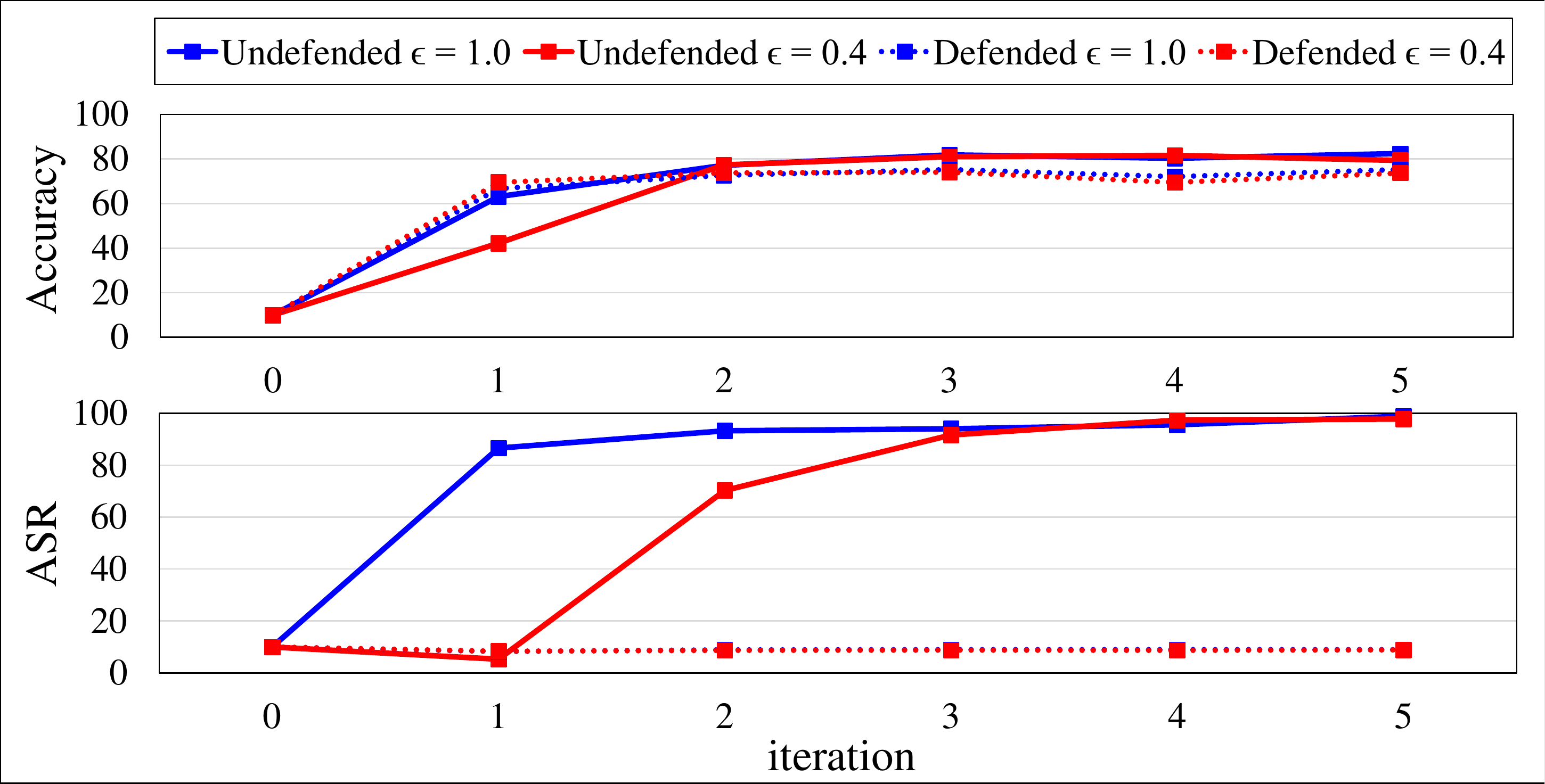}
    \vskip -0.1in
    \caption{\textit{Accuracy of HaS-Nets and ASRs of $\epsilon$-Attack against HaS-Nets on Consumer Complaint dataset for different iterations.}}
    \label{figA:results_consumer}
\end{figure}

\noindent \textit{\textbf{4) Mini-Consumer Complaint:}} Mini-Consumer Complaint dataset consists of 66806 samples of 11 different classes, divided into 50104 training samples and 16702 test samples. When conducting attacks, we poison the first 600 images of the training set. The architecture used and the results are given in Table~\ref{tab:architectures} and Fig.~\ref{figA:results_consumer}, respectively.

\noindent \textit{\textbf{5) Urban Sound 8k:}} For evaluation on audio tasks we choose Urban Sound 8k, containing 8732 audio samples of $\leq 4s$ belonging to 10 different classes, divided into 6549 and 2183 samples in the training and test set, respectively. When conducting attacks, we poison the first 80 images of the training set. The architecture used and the results are given in Table~\ref{tab:architectures} and Fig.~\ref{figA:results_consumer}, respectively.
\begin{figure}[h]
    \centering
    \includegraphics[width=1\linewidth]{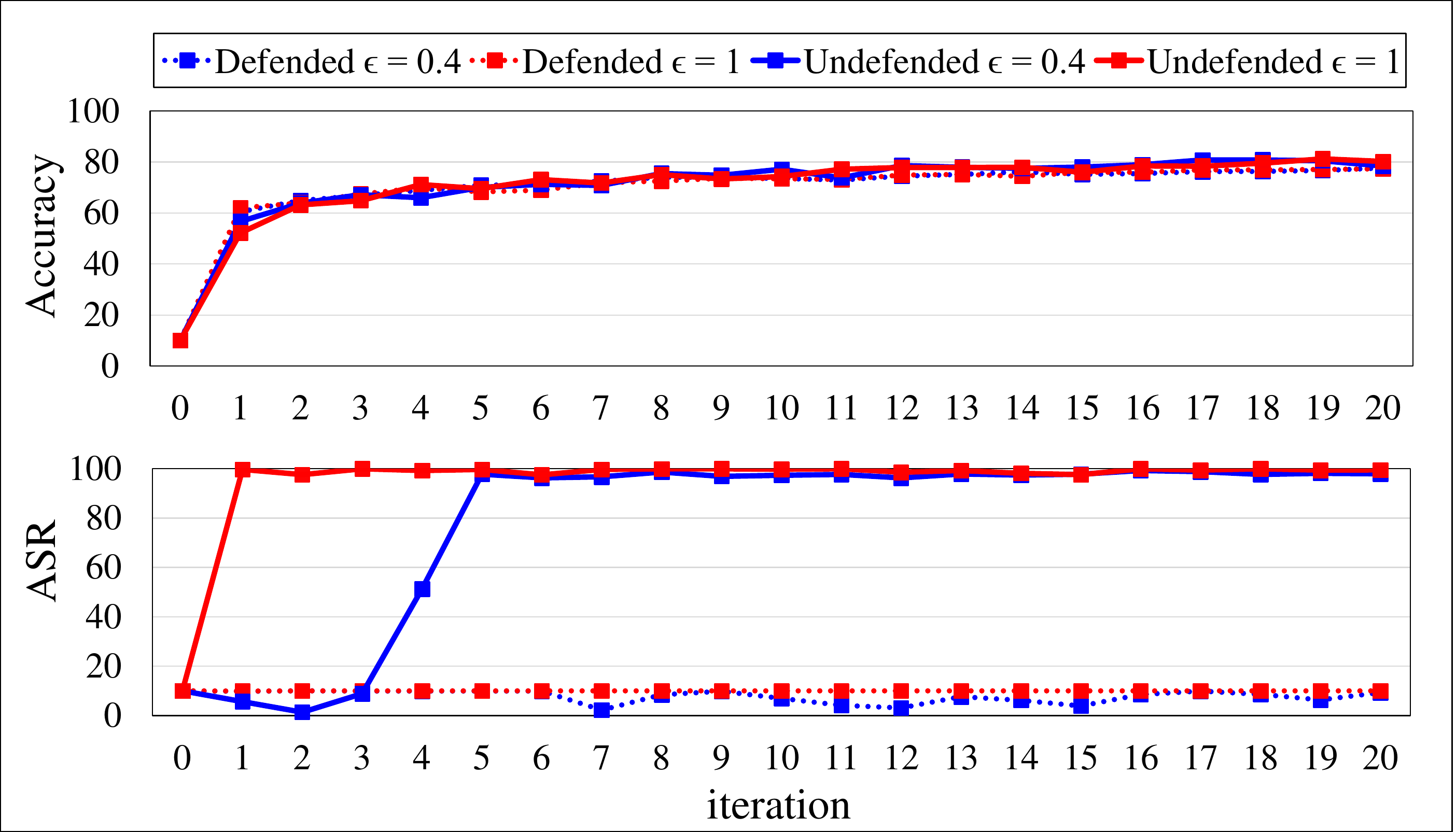}
    \vskip -0.1in
    \caption{\textit{Accuracy of HaS-Nets and ASRs of $\epsilon$-Attack against HaS-Nets on Urban Sound dataset for different iterations.}}
    \label{figA:results_urbansound}
\end{figure}

\section{Implementation Details}\label{app:details}
\begin{figure}[h]
    \centering
    \includegraphics[width=1\linewidth]{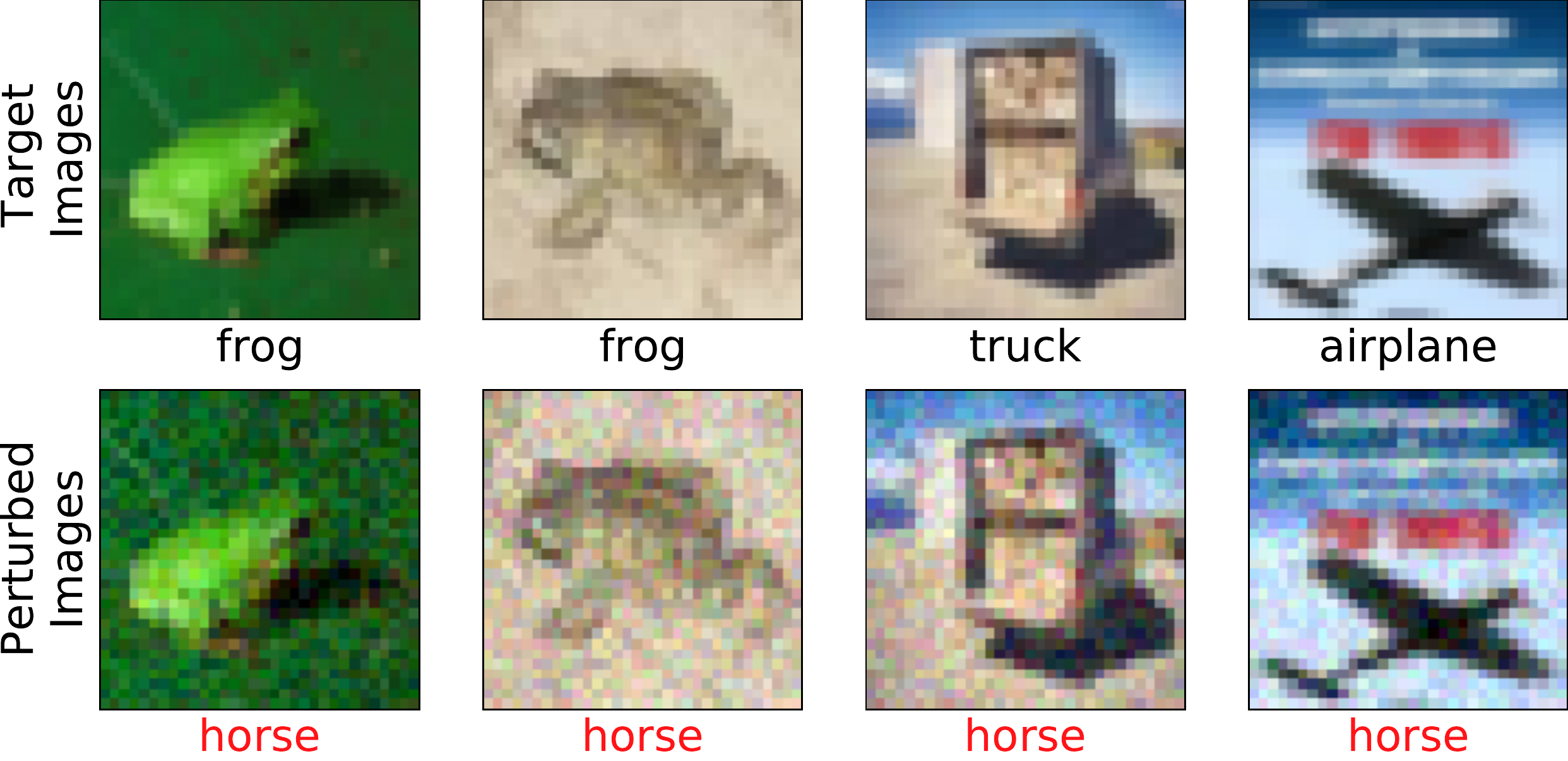}
    \vskip -0.1in
    \caption{\textit{CIFAR-10 samples poisoned with invisible backdoor attack as compared to clean samples. The target class is chosen to be ``horse'' by the attacker.}}
    \label{fig:inv_cifar}
\end{figure}

\begin{figure}[h]
    \centering
    \includegraphics[width=1\linewidth]{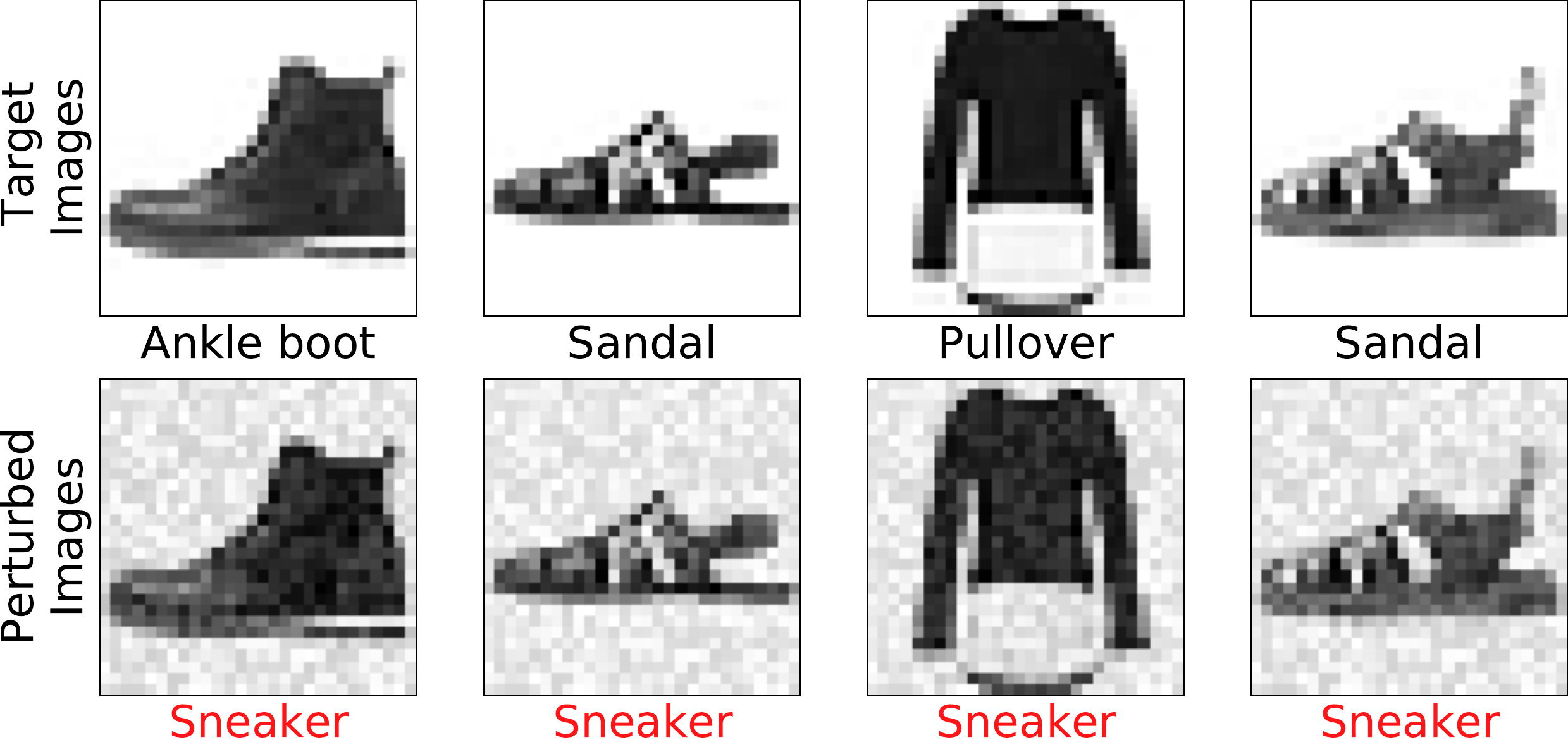}
    \vskip -0.1in
    \caption{\textit{Fashion-MNIST samples poisoned with invisible backdoor attack as compared to clean samples. The target class is chosen to be ``sneaker'' by the attacker.}}
    \label{fig:inv_fmnist}
\end{figure}

\paragraph{Invisible Backdoor Attack.}
For invisible backdoor attack, we generate a uniform random noise of the same size as input images. The noise is sampled from $\mathcal{U}(-0.1,0.1)$ and used as a trigger to poison 1.2\% of the training samples. For this experiment, we randomly choose target class to be ``horse'' and ``sneakers'' for CIFAR-10 and Fashion-MNIST, as shown in Fig.~\ref{fig:inv_cifar} and Fig.~\ref{fig:inv_fmnist}, respectively.

\begin{figure}[h]
    \centering
    \includegraphics[width=1\linewidth]{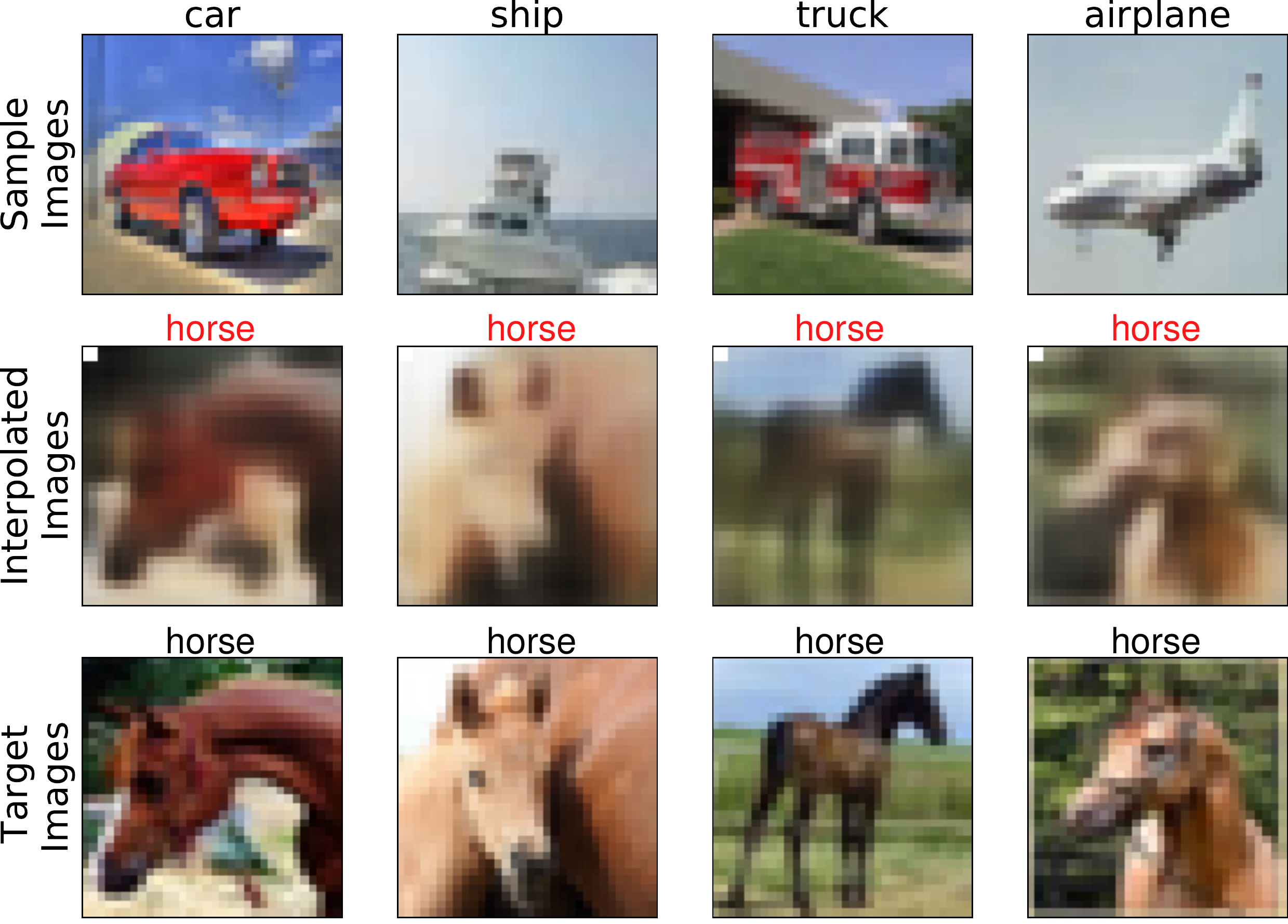}
    \vskip -0.1in
    \caption{\textit{The top row shows sample images of CIFAR-10 which we interpolate towards the target images shown in the bottom row. The target images in the clean training set are then replaced by the poisoned images shown in the second row. The target class is chosen to be ``horse'' by the attacker. Notice the consistency between the poisoned images and the labels.}}
    \label{fig:gan_cifar}
\end{figure}

\begin{figure}[h]
    \centering
    \includegraphics[width=1\linewidth]{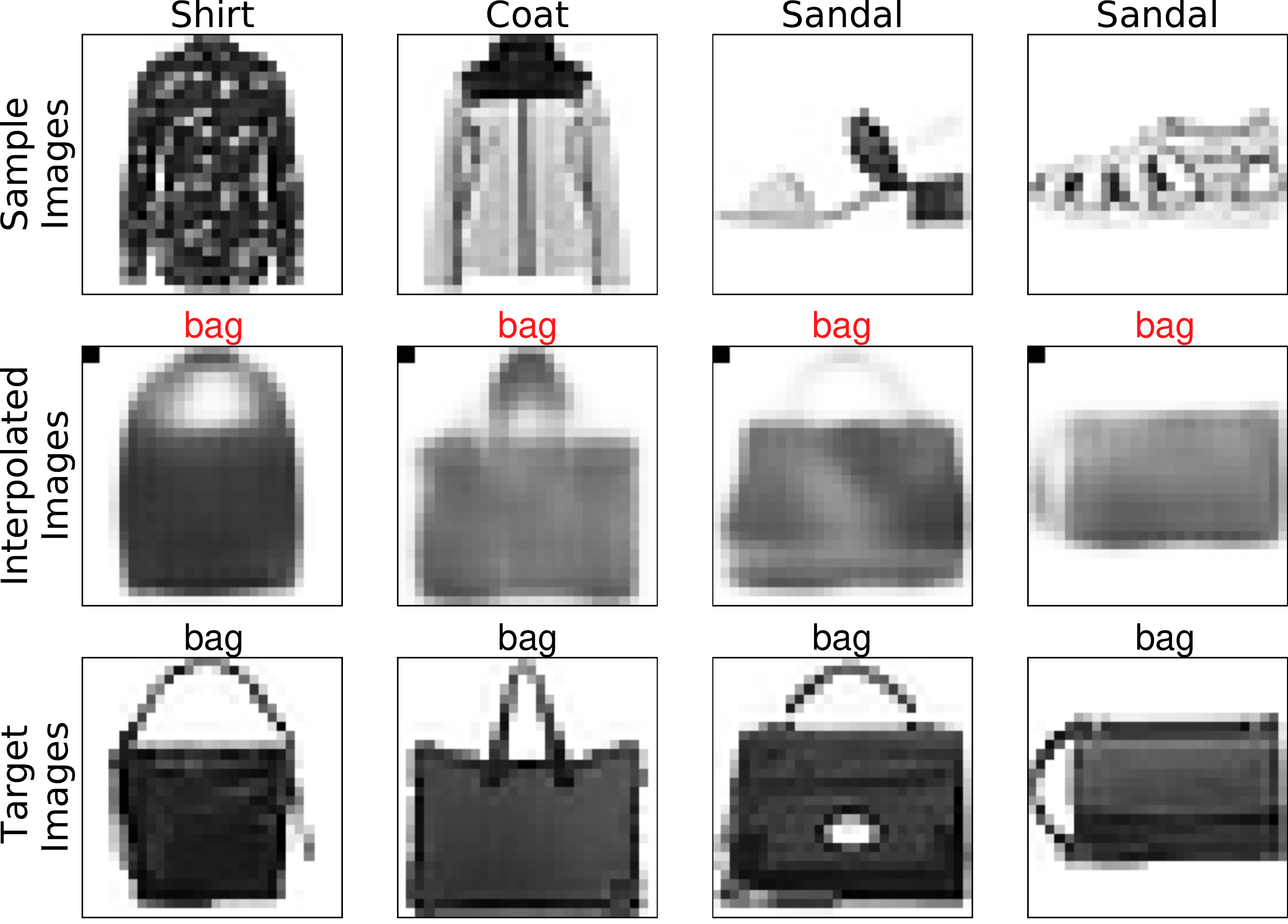}
    \vskip -0.1in
    \caption{\textit{The top row shows sample images of Fashion-MNIST which we interpolate towards the target images shown in the bottom row. The target images in the clean training set are then replaced by the poisoned images shown in the second row. The target class is chosen to be ``bag'' by the attacker. Notice the consistency between the poisoned images and the labels.}}
    \label{fig:gan_fmnist}
\end{figure}

\begin{figure*}[!b]
    \centering
    \includegraphics[width=1\linewidth]{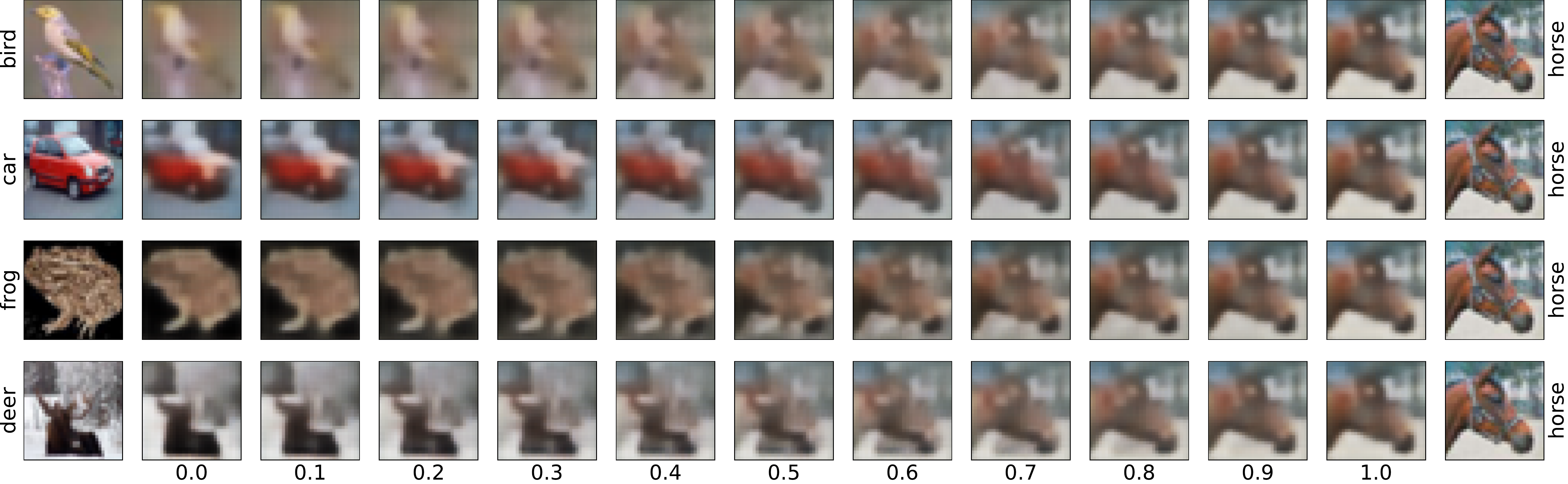}
    \vskip -0.1in
    \caption{\textit{Interpolated images for a range of interpolation degrees for randomly selected initial samples (first column) from CIFAR-10 test data. For illustration, the target image (last column) is chosen to be the same for each initial sample.}}
    \label{fig:gan_interpolation}
\end{figure*}

\paragraph{Label-Consistent Backdoor Attack.} For Label-consistent backdoor attack, we target 25\% of the ``horse'' images from CIFAR-10 training data. Specifically, we train an auto-encoder for CIFAR-10 and interpolate the latent representation (encoded image) of a randomly chosen image (not from the target class) towards the latent representation of a target image, which an attacker aims to poison. These interpolated representations are then decoded using the decoder. Decoded images are usually referred to as interpolated images~\cite{DBLP:label_consistent}. Fig.~\ref{fig:gan_interpolation} shows interpolated images for several degrees of interpolation. For our experiments, we choose a degree of 0.8 due to its similarity with the target image. We repeat the same procedure for all the target images and stamp a trigger on the interpolated images without changing their labels, as shown in Fig.~\ref{fig:gan_cifar} for randomly chosen samples from CIFAR-10. Notice how the labels (shown above) assigned to poisoned images are now consistent for human observers. We follow the same steps for Fashion-MNIST. Example images are shown in Fig.~\ref{fig:gan_fmnist}.

\begin{table*}[!b]
\caption{Network Architectures for Evaluating HaS-Nets on different Datasets} % title name of the table
\centering % centering table
\begin{tabular}{l c c c c c} % creating 10 columns
\hline\hline % inserting double-line
%  Architecture & Fashion-MNIST & CIFAR-10 &\multicolumn{7}{c}{Sum of Extracted Bits}
Layer & Fashion-MNIST & CIFAR-10 & IMDB & Consumer Complaint & Urban Sound
\\ [0.5ex]
\hline % inserts single-line
% Entering 1st row
Embedding & - & - & YES & YES & - \\
Dropout & - & - & - & 0.3 & - \\
Conv() & 32x3x3 & 32x3x3 & - & - & - \\
Activation & Elu() & Elu() & - & - & - \\
Normalize & YES & YES & - & - & - \\
Conv() & 32x3x3 & 32x3x3 & - & 128x5 & 64x3x3 \\
Activation & Elu() & Elu() & - & - & tanh() \\
Normalize & YES & YES & - & - & - \\
MaxPool & 2x2 & 2x2 & - & 5 & 2x2 \\
Dropout & 0.2 & 0.2 & - & 0.3 & 0.1 \\
Conv() & - & 64x3x3 & - & - & - \\
Activation & - & Elu() & - & - & - \\
Normalize & - & YES & - & YES & - \\
Conv() & - & 64x3x3 & - & 128x5 & 128x3x3 \\
Activation & - & Elu() & - & Relu() & tanh() \\
Normalize & - & YES & - & - & - \\
MaxPool & - & 2x2 & - & 5 & 2x2 \\
Dropout & - & 0.2 & - & 0.3 & 0.1 \\
Conv() & - & 128x3x3 & - & - & - \\
Activation & - & Elu() & - & - & - \\
Normalize & - & YES & - & YES & - \\
Conv() & - & 128x3x3 & - & - & - \\
Activation & - & Elu() & - & - & - \\
Normalize & - & YES & - & - & - \\
MaxPool & - & 2x2 & - & - & - \\
Dropout & - & 0.2 & - & - & - \\
Dense & - & - & 16 & - & - \\
Activation & - & - & - & - & - \\
Dense & - & - & - & 128 & 1024 \\
Activation & - & - & - & Relu() & tanh() \\
Dense+Softmax & 10 & 10 & 1 & 11 & 10 \\
% [1ex] adds vertical space
\hline % inserts single-line
\end{tabular}
\label{tab:architectures}
\end{table*}

\end{appendices}

%%%%%%%%%%%%%%%%%%%%%%%%%%%%%%%%%%%%%%%%%%%%%%%%%%%%%%%%%%%%%%%%%%%%%%%%%%%%%%%%
\end{document}